\documentclass{article}

\PassOptionsToPackage{numbers,compress}{natbib}
\usepackage[preprint]{neurips_2026}

\usepackage{microtype}
\usepackage{graphicx}
\graphicspath{{./}}
\usepackage{subcaption}
\usepackage{adjustbox}
\usepackage{placeins} 
\usepackage{booktabs}

\usepackage[dvipsnames]{xcolor}
\definecolor{linkpink}{RGB}{255,105,180}
\usepackage[colorlinks=true,
            citecolor=MidnightBlue,
            linkcolor=MidnightBlue,
            urlcolor=blue]{hyperref}

\usepackage{algorithm}
\usepackage{algpseudocode}

\usepackage{amsmath}
\usepackage{amssymb}
\usepackage{mathtools}
\usepackage{amsthm}
\usepackage{tikz}
\usetikzlibrary{arrows.meta,positioning,calc}

\usepackage{amsmath,amsfonts,bm}

\def\eqref#1{equation~\ref{#1}}

\def\1{\bm{1}}

\DeclareMathAlphabet{\mathsfit}{\encodingdefault}{\sfdefault}{m}{sl}
\SetMathAlphabet{\mathsfit}{bold}{\encodingdefault}{\sfdefault}{bx}{n}

\def\gP{{\mathcal{P}}}

\def\gS{{\mathcal{S}}}

\newcommand{\E}{\mathbb{E}}

\newcommand{\rev}[1]{\textcolor{black}{#1}}
\newcommand{\revcolor}{\color{black}}
\newcommand{\revr}[1]{\textcolor{black}{#1}}

\usepackage[capitalize,noabbrev]{cleveref}
\usepackage{aliascnt}

\theoremstyle{plain}
\newtheorem{theorem}{Theorem}[section]
\crefname{theorem}{Theorem}{Theorems}
\Crefname{theorem}{Theorem}{Theorems}

\newaliascnt{proposition}{theorem}
\newtheorem{proposition}[proposition]{Proposition}
\aliascntresetthe{proposition}
\crefname{proposition}{Proposition}{Propositions}
\Crefname{proposition}{Proposition}{Propositions}

\newaliascnt{lemma}{theorem}
\newtheorem{lemma}[lemma]{Lemma}
\aliascntresetthe{lemma}
\crefname{lemma}{Lemma}{Lemmas}
\Crefname{lemma}{Lemma}{Lemmas}

\newaliascnt{corollary}{theorem}
\newtheorem{corollary}[corollary]{Corollary}
\aliascntresetthe{corollary}
\crefname{corollary}{Corollary}{Corollaries}
\Crefname{corollary}{Corollary}{Corollaries}

\theoremstyle{definition}
\newaliascnt{definition}{theorem}
\newtheorem{definition}[definition]{Definition}
\aliascntresetthe{definition}
\crefname{definition}{Definition}{Definitions}
\Crefname{definition}{Definition}{Definitions}

\newaliascnt{assumption}{theorem}
\newtheorem{assumption}[assumption]{Assumption}
\aliascntresetthe{assumption}
\crefname{assumption}{Assumption}{Assumptions}
\Crefname{assumption}{Assumption}{Assumptions}

\theoremstyle{remark}
\newaliascnt{remark}{theorem}
\newtheorem{remark}[remark]{Remark}
\aliascntresetthe{remark}
\crefname{remark}{Remark}{Remarks}
\Crefname{remark}{Remark}{Remarks}
\crefname{appendix}{Appendix}{Appendices}
\Crefname{appendix}{Appendix}{Appendices}

\usepackage{enumitem}
\usepackage{wrapfig}

\title{MINT: Meeting-time Indicators for Truncation in Multi-Step Off-Policy RL}

\newcommand{\authsep}{,\hspace{0.8em}}

\author{%
  Seungyub Han$^1$\authsep
  Taehyun Cho$^1$\authsep
  Dohyeong Kim$^1$\authsep
  Kyungjae Lee$^2$\thanks{Corresponding authors.}\kern0.45em\authsep
  Jungwoo Lee$^1$\footnotemark[1]\\
  $^1$Seoul National University, $^2$Korea University\\
  {\small\texttt{\{seungyubhan,talium,rlaehgudhoho,junglee\}@snu.ac.kr}\quad
  \texttt{kyungjae\_lee@korea.ac.kr}}\\
}

\begin{document}

\maketitle

\begin{abstract}
Many multi-step off-policy temporal-difference methods handle policy mismatch through products of per-step trace coefficients, leading to a trade-off between variance control and long-horizon signal.
The maximum trace length consequently becomes a sensitive hyperparameter.
We propose \textbf{\texttt{MINT}} (Meeting-time INdicators for Truncation), which replaces these products with a binary stopping indicator: one until coupled behavior and target trajectories first meet, zero thereafter.
At the true value function, the indicator's idempotence yields a variance bound governed by meeting-time survival.
For practical implementation, a single sampled action provides a model-free surrogate for the stopping rule, requiring no transition model or mixing-time estimate.
We derive separate sufficient conditions for contraction of the state-coupled operator and its practical surrogate under fixed-policy evaluation.
On MuJoCo and DeepMind Control tasks, \texttt{MINT} outperforms the reported multi-step baselines; the main MuJoCo comparison uses a common cap without per-task cap-length tuning.
We also demonstrate the effectiveness of \texttt{MINT} on Atari~100k, RWRL, and OGBench experiments; \texttt{MINT} adapts the backup horizon to reduce target variability relative to long fixed-horizon backups across tasks and dynamics.\\
Code: \href{https://github.com/seungyubhan/mint}{\textcolor{linkpink}{\nolinkurl{https://github.com/seungyubhan/mint}}}
\end{abstract}

\section{Introduction}
\label{sec:introduction}

Multi-step temporal-difference (TD) learning propagates rewards over longer horizons, making it a central tool for policy evaluation and credit assignment in reinforcement learning \citep{mahmood2017multi, DBLP:conf/uai/AsisS18, harutyunyan2018beyond, sutton1998intra, precup2001off}. In off-policy learning, however, longer backups also accumulate the effects of mismatch between the behavior and target policies. Many methods handle this mismatch through products of per-step trace coefficients. Unclipped importance-sampling ratios can amplify variance, while clipping controls the weights but can weaken long-horizon learning signals \citep{munos2016safe, rowland2020adaptive}. Methods such as Retrace and Peng's $Q(\lambda)$ make different compromises between signal retention, fixed-point bias, and convergence conditions \citep{munos2016safe, kozuno2021revisiting, geist2014off}. These trade-offs make the maximum trace length, or \emph{cap length}, a consequential choice and motivate adapting the backup horizon to the encountered policy mismatch.
Prior on-policy analysis offers a complementary perspective by relating sufficient backup length to temporal dependence \citep{duan2022policy}. Estimating the relevant mixing properties can be difficult, and the corresponding guarantees do not directly extend to off-policy sampling. This motivates a sample-based stopping rule that adapts the backup horizon without estimating mixing time:
\begin{center}
\textit{Can idempotent coupling indicators enable adaptive backups that better balance long-horizon learning signals and target variability?}
\end{center}

\begin{figure}[t]
    \centering
    \begin{subfigure}[t]{0.60\linewidth}
        \centering
        \vspace{0pt}
        \includegraphics[width=\linewidth]{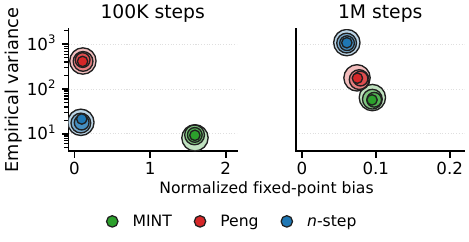}
        \caption{Bias--variance trade-off.}
        \label{fig:bv_sac}
    \end{subfigure}
    \hfill
    \begin{subfigure}[t]{0.38\linewidth}
        \centering
        \vspace{0pt}
        \includegraphics[width=\linewidth]{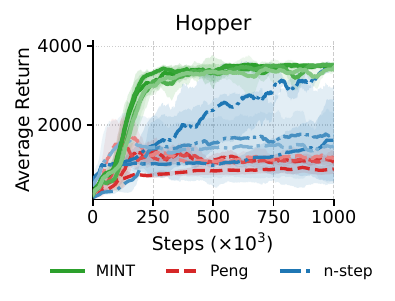}
        \caption{Learning curves.}
        \label{fig:hopper-cap}
    \end{subfigure}
    \caption{\textbf{Cap-length sensitivity on MuJoCo \texttt{Hopper} (SAC).}
    (a) Bias and variance at $100$K and $1$M steps for caps $\{3,5,10,20\}$ (darker shades indicate shorter caps). \texttt{MINT} exhibits a favorable empirical bias--variance balance across the tested caps; several baseline configurations have higher variance at longer caps.
    (b) Learning curves for the same caps. \texttt{MINT} attains high returns across the tested values, whereas baseline performance varies more with the cap.}
    \label{fig:hopper-headline}
\end{figure}

We introduce \textbf{\texttt{MINT}} (Meeting-time \textbf{IN}dicators for Truncation). In its theoretical construction, behavior and target state processes are coupled, and a binary mask retains the trace until their first agreement. The cumulative mask remains zero thereafter, turning the backup into a randomly truncated sum of TD errors. For model-free implementation, an action-level surrogate uses the policy ratio at a single sampled action to decide whether to continue. This surrogate defines its own stopping process and requires neither transition-kernel access nor a mixing-time estimate.
Figure~\ref{fig:hopper-headline} illustrates the empirical motivation on MuJoCo \texttt{Hopper}. Across the tested caps, \texttt{MINT} exhibits a favorable empirical bias--variance balance and maintains high returns. This motivates examining how stochastic truncation changes both learning performance and the resulting target estimates.

\paragraph{Contributions.}
We provide (i) a coupling-based formulation of stochastic truncation and a model-free action-level implementation (\cref{sec:t4}); (ii) a survival-based variance bound at $Q^\pi$ for the state-coupled estimator and separate sufficient contraction conditions for the state-coupled and practical surrogate operators under fixed-policy evaluation (\cref{lem:variance,thm:contraction,thm:surrogate-contraction}); and (iii) comparative evaluations demonstrating gains over multi-step baselines on MuJoCo and DMC, using a common cap without per-task cap tuning in the main MuJoCo comparison, alongside task-dependent benefits on Atari~100k, RWRL, and OGBench (\cref{sec:exp}).

\section{Preliminaries}\label{sec:prelim}

We consider a Markov decision process (MDP)
$(\mathcal{S}, \mathcal{A}, \mathcal{P}, \mathcal{P}_0, r, \gamma)$ with finite
state and action spaces, transition kernel
$\mathcal{P} : \mathcal{S} \times \mathcal{A} \to \Delta(\mathcal{S})$, initial
distribution $\mathcal{P}_0$, bounded reward
$r : \mathcal{S} \times \mathcal{A} \to [-r_{\max}, r_{\max}]$, and discount
$\gamma \in [0, 1)$ \citep{sutton2018reinforcement}. For a policy $\pi$, the
Bellman operator is $\mathcal{T}^\pi Q := r + \gamma \mathcal{P}^\pi Q$, where
$(\mathcal{P}^\pi Q)(s, a) := \mathbb{E}_{s' \sim \mathcal{P}(\cdot \mid s,a),\, a' \sim \pi(\cdot \mid s')}[Q(s', a')]$.
Trajectories $(s_t, a_t, r_t)_{t \ge 0}$ are generated by a behavior policy
$\beta$, and our goal is to estimate the action-value function $Q^\pi$ of a
target policy $\pi \neq \beta$ from these off-policy samples
\citep{precup2001off, sutton1998intra}.

\subsection{The Retrace Operator and the Variance Explosion Problem}\label{sec:prelim-retrace}
A central challenge in multi-step off-policy evaluation is correcting the
distributional mismatch between $\beta$ and $\pi$ without inflating the
variance of the multi-step return. The \emph{general Retrace operator}
\citep{munos2016safe} formalizes this correction through a sequence of per-step
\emph{trace coefficients} $c_i := c(s_i, a_i)$:
\begin{equation}\label{eq:retrace-general}
\mathcal{R} Q
\;=\; Q
\;+\; \mathbb{E}_{\beta}\!\left[
\sum_{t=0}^{\infty} (\gamma\lambda)^t
\Bigl(\textstyle\prod_{i=1}^{t} c_i\Bigr)\,
\delta_t^{Q}
\right],
\qquad
\delta_t^{Q} := r_t + \gamma\,\mathbb{E}_{\pi}\!\left[Q(s_{t+1}, \cdot)\right] - Q(s_t, a_t),
\end{equation}
with the convention $\prod_{i=1}^{0} c_i = 1$ and $\lambda \in [0,1]$
controlling the geometric decay of the trace
\citep{bertsekas1996temporal}. Existing multi-step off-policy methods
instantiate \cref{eq:retrace-general} by a choice of $c_i$:
Retrace \citep{munos2016safe} uses $c_i=\min(1,\pi(a_i\mid s_i)/\beta(a_i\mid s_i))$;
Harutyunyan's $Q^\pi(\lambda)$ uses $c_i=1$ \citep{harutyunyan2016q}.
A sufficient conservative constraint is $0\le c_i\le\pi(a_i\mid s_i)/\beta(a_i\mid s_i)$,
which guarantees contraction of the expected evaluation operator toward $Q^\pi$
for fixed policies \citep{munos2016safe}.

\paragraph{The variance explosion problem.}
For a finite cap $L$, write $Z_t:=(\gamma\lambda)^t(\prod_{i=1}^tc_i)\delta_t^Q$.
The estimator's second moment has the exact expansion
\begin{equation}\label{eq:variance-product}
\E\!\left[\left(\sum_{t=0}^{L-1}Z_t\right)^2\right]
=\underbrace{\sum_{t=0}^{L-1}(\gamma\lambda)^{2t}\E\!\left[
\Bigl(\prod_{i=1}^tc_i^2\Bigr)(\delta_t^Q)^2\right]}_{\text{diagonal terms }\E[Z_t^2]}
+\underbrace{2\sum_{0\le u<t<L}\E[Z_uZ_t]}_{\text{cross terms}}.
\end{equation}
Trace weights and TD errors can be dependent, and cross terms need not
vanish at arbitrary $Q$. Unclipped importance-sampling ratios can amplify
second moments over long traces; clipped coefficients need not exhibit
this growth. \Cref{tab:operator-comparison} summarizes representative
operators using the classification of \citet{kozuno2021revisiting}.
Conservative methods such as Retrace bound coefficient products but may
weaken long-horizon signals. Non-conservative methods can retain more
signal but may introduce off-policy bias or require additional stability
conditions \citep{kozuno2021revisiting, rowland2020adaptive}.
These trade-offs can make performance sensitive to cap length,
motivating task-specific tuning \citep{hessel2018rainbow}.

\subsection{Mixing Time and Effective Horizon}\label{sec:prelim-mixing}

Mixing time describes how quickly a Markov chain forgets its initial
state. For an ergodic behavior chain with transition kernel
$\mathcal P^\beta$ and stationary distribution $\mu_\beta$,
the total-variation (TV) mixing time is
\[
\tau_{\mathrm{mix}}^\beta
:=\min\left\{t\ge0:\sup_s\left\|
(\mathcal P^\beta)^t(s,\cdot)-\mu_\beta
\right\|_{\mathrm{TV}}\le\tfrac14\right\}.
\]
It measures convergence to stationarity uniformly over initial states
\citep{levin2017markov}, but can be difficult to estimate directly
from trajectories \citep{wolfer2019estimating}.
For on-policy evaluation ($\beta=\pi$) with bounded rewards,
\citet[Section~3.3.2, Eq.~(29b)]{duan2022policy} obtain finite-sample
bounds for multi-step kernel least-squares TD under their assumptions
and the look-ahead condition
\begin{equation}\label{eq:mixing-bound}
K\;\gtrsim\;\min\!\left\{H,\tau_*^\beta\right\},
\qquad H:=(1-\gamma)^{-1}.
\end{equation}
Here $\tau_*^\beta$ denotes the mixing parameter used in their analysis,
whose definition differs from the TV mixing time
$\tau_{\mathrm{mix}}^\beta$ above; see \cref{app:mixing-background}
for its definition and the detailed assumptions.
This condition links the backup horizon to chain-level temporal
dependence, but applying it requires knowledge of mixing properties.
Moreover, the guarantees of \citet{duan2022policy} rely on on-policy
sampling and do not directly extend to $\beta\ne\pi$.
These limitations motivate our sample-based stopping rule for
off-policy multi-step evaluation (\cref{sec:t4}).

\section{The \texttt{MINT} Operator}\label{sec:t4}

We introduce \texttt{MINT}, which replaces products of real-valued trace coefficients with a binary stopping rule. We first define the state-coupled operator, then establish a variance bound at $Q^\pi$, and finally describe its model-free action-level surrogate.
Let $S_t^\beta,S_t^\pi$ start from a shared state. At policy-generated steps, maximally couple their policy-induced next-state kernels. Write $\mathcal H_t$ for the joint pre-action history and $\mathcal G_t=\mathcal H_t\vee\sigma(a_t)$. At such steps, the behavior action has conditional law $\beta(\cdot\mid S_t^\beta)$ given $\mathcal H_t$. The conditional environment law of $(r_t,S_{t+1}^\beta)$ given $\mathcal G_t$ must equal its law given $(S_t^\beta,a_t)$. The filtration and initial-action conditioning are specified in \cref{app:variance-lemma}.

\begin{definition}[State-coupled stopping rule]\label{def:meeting-time}
For $i\ge1$ and $t\ge0$, we define
\[
A_i:=\mathbf1\{S_i^\beta\ne S_i^\pi\},\quad
T_{\beta,\pi}:=\inf\{i\ge1:A_i=0\},\quad
W_t:=\prod\nolimits_{i=1}^tA_i=\mathbf1\{T_{\beta,\pi}>t\},
\]
with $W_0=1$ and $\inf\varnothing=\infty$.
\end{definition}
The cumulative mask remains zero after first agreement, even if the states later separate. It is $\mathcal H_t$-measurable, hence $\mathcal G_t$-measurable, and cannot depend on the next reward or transition.
We write $s_t=S_t^\beta$ for brevity, and let $\E_x,\Pr_x$ refer to the joint trajectory law conditioned on the starting pair $x=(s_0,a_0)$. Replacing $\prod_{i=1}^tc_i$ in \cref{eq:retrace-general} by $W_t$ yields
\begin{equation}\label{eq:t4-operator}
\begin{aligned}
(\mathcal R_{p,\lambda}Q)(x)
&=Q(x)+\E_x\!\left[\sum_{t\ge0}(\gamma\lambda)^tW_t\delta_t^Q\right]=Q(x)+\E_x\!\left[\sum_{t=0}^{\textcolor{linkpink}{\bm{T}_{\beta,\pi}-1}}(\gamma\lambda)^t\delta_t^Q\right].
\end{aligned}
\end{equation}
The expectation averages environment and coupling randomness. Subscript $p$ identifies the disagreement process; indicators can be dependent. The operator fixes $Q^\pi$; uniqueness and convergence require the separate contraction condition in \cref{sec:theory}.

\begin{table}[t]
\centering\small
\caption{\textbf{Multi-step off-policy operators for behavior and target policies.} $W_t$ excludes $(\gamma\lambda)^t$; $\E[W_t^2]$ describes squared weights, not the full estimator variance.}
\label{tab:operator-comparison}
\setlength{\tabcolsep}{4pt}
\begin{tabular}{@{}lcccc@{}}
\toprule
Operator & Trace / return rule & $\E[W_t^2]$ & \shortstack{Conservative?} & Fixed point \\
\midrule
Retrace \citep{munos2016safe} & $c_i=\min(1,\pi/\beta)$ & $\E[\prod_{i=1}^tc_i^2]$ & Yes & $Q^\pi$ \\
Peng's $Q(\lambda)$ \citep{kozuno2021revisiting} & $c_i=1^{\dagger}$ & $1$ & No & $Q^{\pi_\lambda}$ \\
Uncorrected $n$-step & $\mathcal N_n^{\ddagger}$ & $\mathbf1\{t<n\}$ & No & $\beta$-dependent \\
\textbf{\texttt{MINT}} & $c_i=A_i\in\{0,1\}$ & $\Pr(T>t)$ & No & $Q^\pi$ \\
\bottomrule
\end{tabular}
\par\smallskip
\begin{minipage}{\linewidth}\footnotesize
$\dagger$~Peng's return mixture has this TD-error representation with target $\pi_\lambda:=\lambda\beta+(1-\lambda)\pi$, rather than $\pi$.
$\ddagger$~$\mathcal N_n:=(\mathcal T^\beta)^{n-1}\mathcal T^\pi$ is the ordinary fixed-horizon return ($\lambda=1$); its fixed point is generally behavior-dependent for $n>1$.
\end{minipage}
\end{table}

\subsection{Variance at the Target Value Function}\label{sec:t4-variance}

Two properties enter the variance analysis: conditional mean-zero TD errors eliminate cross terms, while idempotence turns squared binary masks into survival indicators.

\begin{definition}[Discounted survival and policy mismatch]\label{def:M}
For starting pair $x=(s_0,a_0)$, write $T=T_{\beta,\pi}$ and $D_s:=\mathrm{TV}(\beta(\cdot\mid s),\pi(\cdot\mid s))$. For $\gamma\in(0,1)$ and $\lambda\in[0,1]$, we define
\begin{equation}\label{eq:M-def}
\begin{aligned}
M_x(\gamma,\lambda)&:=\sum_{t\ge0}(\gamma\lambda)^t\Pr_x(T>t),\qquad N_x(\gamma,\lambda)&:=\sum_{t\ge1}(\gamma\lambda)^t\E_x[\mathbf1\{T>t\}D_{s_t}].
\end{aligned}
\end{equation}
$M_x$ measures discounted trace survival; $N_x$ weights policy mismatch along the surviving trace.
\end{definition}

\begin{lemma}[Martingale property, fixed point, and variance]\label{lem:variance}
Under the joint sampling conditions above, let $\delta_t^{Q^\pi}$ be the TD error of \cref{eq:retrace-general} at $Q^\pi$. Let
$C:=\|r\|_\infty+(1+\gamma)\|Q^\pi\|_\infty<\infty$.
Then, for every starting pair $x$:

\noindent\textbf{(i) Martingale property.}
$\E_x[\delta_t^{Q^\pi}\mid\mathcal G_t]=0$ and $W_t$ is $\mathcal G_t$-measurable.

\noindent\textbf{(ii) Fixed point.}
$\mathcal R_{p,\lambda}Q^\pi=Q^\pi$.

\noindent\textbf{(iii) Variance.} For
$\hat\Delta:=\sum_{t\ge0}(\gamma\lambda)^tW_t\delta_t^{Q^\pi}$,
\begin{equation}\label{eq:variance-bound}
\begin{aligned}
\operatorname{Var}_x(\hat\Delta)=\E_x[\hat\Delta^2]
&=\sum_{t\ge0}(\gamma\lambda)^{2t}\E_x[W_t(\delta_t^{Q^\pi})^2] \le C^2M_x(\gamma^2,\lambda^2).
\end{aligned}
\end{equation}
\end{lemma}

For $u<t$, the earlier TD error and both masks are measurable at $\mathcal G_t$, so the tower property removes the cross term. Idempotence then gives $W_t^2=W_t$, yielding the diagonal expression in \cref{eq:variance-bound}. The full proof is in \cref{app:variance-lemma}. The cross-term argument also holds for other suitably adapted weights; binary masks additionally express diagonal weights through survival probabilities, giving
\[
\operatorname{Var}_x(\hat\Delta)\le C^2\min\!\left\{\E_x[T_{\beta,\pi}],\frac1{1-\gamma^2\lambda^2}\right\}.
\]
This bound concerns evaluation at $Q^\pi$; it does not establish variance dominance over Retrace or describe the full error of a learned critic. It offers a survival-based perspective alongside the mixing-based analysis in \cref{sec:prelim-mixing}. Contraction requires the additional bounds developed in \cref{sec:theory}.

\subsection{Practical Implementation}\label{sec:t4-practical}

Constructing the state coupling requires transition-kernel information. The implementation instead uses the action-level keep probability
\begin{equation}\label{eq:practical-disagreement}
\hat p_i:=1-\min\!\left(1,\frac{\pi(a_i\mid s_i)}{\beta(a_i\mid s_i)}\right).
\end{equation}
Here $\beta$ is the policy that generated the stored transition: its log-probability is recorded at collection time and compared with the current $\pi$ at replay.
Using randomness independent of subsequent rewards and transitions, draw $\hat A_i\sim\operatorname{Bernoulli}(\hat p_i)$ and define
\begin{equation}\label{eq:effective-K}
\hat T_{\beta,\pi}:=\inf\{i\ge1:\hat A_i=0\},\qquad
\hat K:=\min(K_{\max},\hat T_{\beta,\pi}),
\end{equation}
with $\inf\varnothing=\infty$. The implementation fixes a hard cap $K_{\max}$; the sampled stopping time selects the realized prefix within it. The state-coupled operator above is uncapped, whereas the surrogate evaluation theorem uses an explicit finite cap.
The rule retains the trace on disagreement: when $\pi=\beta$, $\hat p_i=0$ and the backup is one-step. A larger $\hat p_i$ makes continuation more likely at that step. Unlike Retrace, the keep rule uses the complement of the clipped policy ratio. The expected effective horizon $K_{\mathrm{eff}}:=\E[\hat K]$ depends on the full sequence of keep probabilities.

\paragraph{Relation to the state coupling.}
The surrogate's conditional mean is the action-level policy TV,
\[
\E_{a\sim\beta(\cdot\mid s)}[\hat p(s,a)]=D_s
\ge d(s):=\mathrm{TV}(\mathcal P^\beta(s,\cdot),\mathcal P^\pi(s,\cdot)),
\]
by the data-processing inequality (\cref{prop:dpi-stochastic}). This compares one-step kernels from a shared state; it does not identify its multi-step survival law with that of the state coupling. The expected surrogate evaluation has its own finite-cap contraction condition (\cref{thm:surrogate-contraction,app:surrogate-contraction}).

\begin{figure}[t]
    \centering
    \begingroup
    \sbox0{\includegraphics[width=0.46\textwidth]{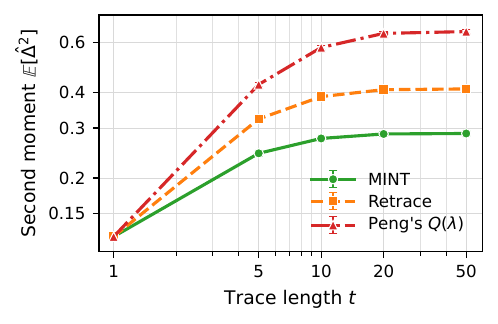}}
    \begin{subfigure}[t]{0.52\textwidth}
        \centering
        \vspace{0pt}
        \begin{minipage}[c][\ht0][c]{\linewidth}
        \centering
        \begin{tikzpicture}[
            >={Stealth[length=1.8mm]},
            state/.style={circle, draw=black!75, line width=0.65pt,
                minimum size=7.5mm, inner sep=1pt, font=\small},
            trap/.style={state, fill=red!12, line width=0.9pt},
            arr/.style={->, draw=black!80, line width=0.7pt, shorten >=1pt, shorten <=1pt},
            lbl/.style={font=\footnotesize, inner sep=2pt},
        ]
            \node[state] (s0) at (0, 0)   {$0$};
            \node[state] (s1) at (1.12, 0) {$1$};
            \node       (sd) at (2.24, 0) {$\cdots$};
            \node[state] (sN1) at (3.36, 0) {$N{-}1$};
            \node[trap]  (sN)  at (4.48, 0) {$N$};
            \draw[arr] (s0) to[bend left=32] node[lbl, above=1pt] {$+1$} (s1);
            \draw[arr] (s1) to[bend left=32] (sd);
            \draw[arr] (sd) to[bend left=32] (sN1);
            \draw[arr] (sN1) to[bend left=32] (sN);
            \draw[arr] (s1) to[bend left=32] node[lbl, below=1pt] {$-1$} (s0);
            \draw[arr] (sd) to[bend left=32] (s1);
            \draw[arr] (sN1) to[bend left=32] (sd);
            \draw[arr] (sN) to[bend left=32] (sN1);
            \draw[arr] (s0) to[out=150, in=210, looseness=6] (s0);
            \draw[arr] (sN) to[out=-30, in=30, looseness=6] (sN);
            \node[lbl, anchor=south] at ([yshift=2pt]sN.north) {$r{=}{+}1$};
        \end{tikzpicture}
        \par\vspace{5pt}
        {\footnotesize
            $\beta(\pm1\mid s)=0.5$;\quad $\pi(\pm1\mid s\ne N)=0.5$\par
            \vspace{2pt}
            $\pi(+1\mid N)=1$;\quad slip $\xi=0.1$\par
        }
        \end{minipage}
        \caption{Single-Trap chain MDP.}
        \label{fig:chain-diagram}
    \end{subfigure}
    \hfill
    \begin{subfigure}[t]{0.46\textwidth}
        \centering
        \vspace{0pt}
        \usebox0
        \caption{Empirical second moment of $\hat\Delta$.}
        \label{fig:variance-empirical-curve}
    \end{subfigure}
    \endgroup
      \caption{\textbf{Single-Trap chain: empirical second moment.} (a) Chain MDP. (b) Second moment of the multi-step Bellman residual $\hat\Delta$ over trace lengths at $\gamma=0.9$, $\lambda=1$ (log--log axes). Error bars ($\pm1$ standard deviation, 30 replicates) are smaller than the markers. At $t=50$, Retrace and Peng's $Q(\lambda)$ are $1.43\times$ and $2.28\times$ \texttt{MINT}, whose curve levels off over the tested lengths. Setup: \cref{app:single-trap-moment}.}
    \label{fig:variance-empirical}
\end{figure}

\section{Theoretical Analysis}\label{sec:theory}\label{sec:theory-toy}

The variance lemma (\cref{lem:variance}) controls the estimator at $Q^\pi$. We next give a sufficient contraction condition for the state-coupled operator under an arbitrary behavior policy.
\begingroup\color{black}
Use $M_x=M_x(\gamma,\lambda)$ and $N_x=N_x(\gamma,\lambda)$ from \cref{def:M}, retaining their starting-pair dependence.

\begin{theorem}[Contraction of $\mathcal{R}_{p,\lambda}$]\label{thm:contraction}
Under the joint sampling conditions of \cref{sec:t4}, fix policies $\beta,\pi$, $\gamma\in(0,1)$, and $\lambda\in[0,1]$. We define
\begin{equation}\label{eq:contraction-condition}
\kappa:=\sup_x\bigl\{1-(1-\gamma)M_x+2N_x\bigr\}.
\end{equation}
Then, for all bounded $Q,Q'$,
\begin{equation}\label{eq:contraction-modulus}
\|\mathcal R_{p,\lambda}Q-\mathcal R_{p,\lambda}Q'\|_\infty
\le\kappa\|Q-Q'\|_\infty.
\end{equation}
If $\kappa<1$, $\mathcal R_{p,\lambda}$ is a $\kappa$-contraction with unique fixed point $Q^\pi$.
\end{theorem}
\endgroup

\begin{algorithm}[t]
\caption{\texttt{MINT} target (summary). Full procedure: \cref{alg:t4-full}.}
\label{alg:t4}
\begin{algorithmic}[1]
\Require Trajectory with stored behavior log-probabilities; $\pi,Q_{\bar\theta},\gamma,\lambda$; $K_{\max}\ge1$
\State For $1\le i<K_{\max}$, compute $\hat p_i$ by \cref{eq:practical-disagreement} and independently draw $\hat A_i\sim\operatorname{Bernoulli}(\hat p_i)$
\State $\hat K\gets\min(K_{\max},\hat T_{\beta,\pi})$
\State \Return $Q_{\bar\theta}(s_0,a_0)+\sum_{t=0}^{\hat K-1}(\gamma\lambda)^t\delta_t^{Q_{\bar\theta}}$
\end{algorithmic}
\end{algorithm}

For each starting pair, $1-(1-\gamma)M_x$ is the total update-weight contribution and $2N_x$ bounds the negative-weight correction. In the on-policy case $M_x=1$ and $N_x=0$ for all $x$, recovering $\kappa=\gamma$; the full proof is in \Cref{proof:contraction}.
The first agreement time here differs from the behavior chain's mixing time (\cref{app:two-couplings}). The action-dependent surrogate has its own finite-cap contraction theorem (\cref{thm:surrogate-contraction,tab:assumption-usage}).
A separate exact-control result for Markov trace operators, under additional overlap and policy-improvement hypotheses, is given in \cref{thm:online_t4}.
We can also normalize the surviving mismatch by setting $D_{\mathrm{wgt}}^{\,x}:=N_x/(M_x-1)$ for $M_x>1$ and $D_{\mathrm{wgt}}^{\,x}:=0$ otherwise; let $\bar D_{\mathrm{wgt}}:=\sup_xD_{\mathrm{wgt}}^{\,x}$.

\begin{wrapfigure}{r}{0.45\linewidth}
\vspace{-10pt}
\centering
\captionsetup{font=small,labelfont=bf,justification=raggedright,singlelinecheck=false,skip=3pt}
\includegraphics[width=0.98\linewidth]{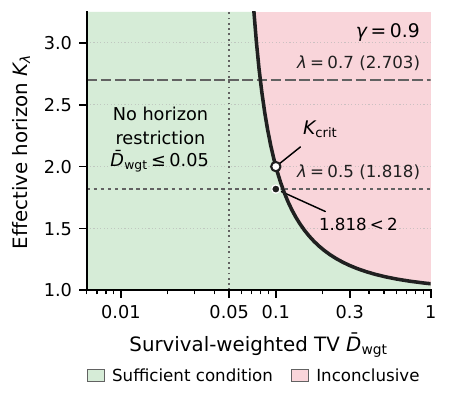}
\caption{\textbf{State-coupled horizon budget.}
At $\gamma=0.9$, green satisfies the sufficient condition in
\cref{cor:practical-worst}; pink is inconclusive.
Guides show discount-only bounds for $\lambda\in\{0.7,0.5\}$.}
\label{fig:phase-diagram-main}
\vspace{-10pt}
\end{wrapfigure}

\Cref{thm:contraction} relates contraction to policy mismatch
along surviving traces. To express this relationship as a
horizon budget, let
$K_\lambda:=\sup_x\min\{\E_x[T],\allowbreak
(1-\gamma\lambda)^{-1}\}$, so $M_x\le K_\lambda$.
This survival bound is distinct from the hard cap
$K_{\max}$; \cref{cor:practical-worst} gives two sufficient regimes.
When $\bar D_{\mathrm{wgt}}\le(1-\gamma)/2$, no additional
meeting-time restriction is needed. Otherwise, contraction is
ensured by
$K_\lambda<2\bar D_{\mathrm{wgt}}/
[2\bar D_{\mathrm{wgt}}-(1-\gamma)]$.
Thus, the admissible horizon depends jointly on survival
and the policy mismatch encountered before stopping.
At $\gamma=0.9$, mismatch up to $0.05$ requires no additional
horizon restriction. At mismatch $0.1$, the requirement
$K_\lambda<2$ is met by $\lambda=0.5$, since
$K_\lambda\le1.818$.
\Cref{fig:phase-diagram-main,tab:contraction-gamma09}
illustrate these budgets.
Under additional overlap assumptions, a geometric specialization is given in \cref{cor:geometric,fig:phase-diagram}.
The action-level surrogate instead has its
finite-cap contraction conditions
(\cref{thm:surrogate-contraction,cor:surrogate-overlap}).

\section{Experiments}\label{sec:exp}

We evaluate data-efficient learning, adaptive trace lengths, task-dependent backup trade-offs, and behavior under stochastic dynamics:
\begin{itemize}
    \item[\textbf{RQ1}] How does \texttt{MINT} compare with multi-step baselines?
    \item[\textbf{RQ2}] Does stochastic truncation produce a per-task adaptive trace length without estimating $\tau_{\mathrm{mix}}$?
    \item[\textbf{RQ3}] Under which task conditions does \texttt{MINT} improve the backup trade-off, and how do policy mismatch and target spread characterize these gains?
    \item[\textbf{RQ4}] How do target estimation and learned-critic behavior change under stochastic dynamics?
\end{itemize}

\subsection{Experimental Setup}\label{sec:exp-setup}

We evaluate MuJoCo \citep{todorov2012mujoco} with SAC \citep{haarnoja2018soft} and TD3 \citep{pmlr-v80-fujimoto18a}, eleven tasks from the DeepMind Control Suite (DMC) \citep{tassa2018deepmindcontrolsuite} with SAC (task selection: \cref{app:dmc-task-selection}), and Atari~100k \citep{kaiser2019model} with a DrQ($\varepsilon$)-based variant \citep{yarats2021image}. Additional RWRL \citep{dulacarnold2021challenges} and OGBench \citep{park2025ogbench} experiments diagnose cap sensitivity and fixed-policy estimation. The five-task MuJoCo comparison uses a common cap of $5$, $\lambda=0.7$ for \texttt{MINT}, Peng's $Q(\lambda)$, and Retrace, fixed SAC $\alpha=0.2$, and five seeds per method/task for 1M steps.
The 11-task DMC comparison uses five seeds per method/task, 500K steps, $\gamma=0.99$, automatic entropy tuning, and cap length $10$; \texttt{MINT} uses $\lambda=0.9$, while its $n$-step SAC baseline uses a pure uncorrected return with $n=10$. IQM uses $95\%$ stratified-bootstrap intervals \citep{agarwal2021deep}. Settings are matched within each comparison; protocols and normalization are not pooled across suites. Atari settings are in \cref{app:atari100k}. The supplementary five-task DMC evaluation uses a separate 1M-step protocol; \cref{tab:dmc-protocols} distinguishes its seeds, targets, and score windows from the primary comparison.

\begin{figure}[t]
\centering
\includegraphics[width=5.5in]{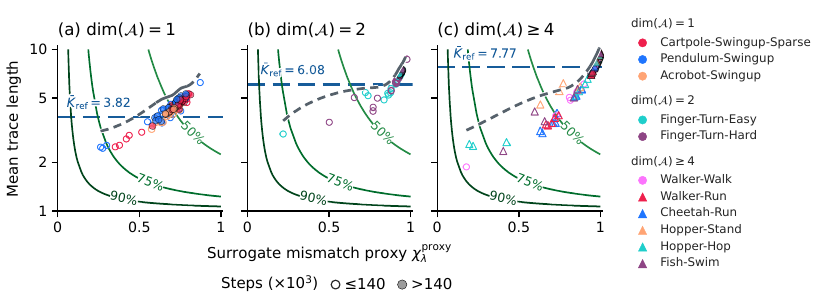}
\caption{\textbf{Adaptive trace lengths and mismatch visualization on 11-task DMC.} Points show mean trace lengths vs the action-based mismatch proxy $\chi_\lambda^{\rm proxy}$. Hollow/filled markers denote measurements at $\le140$K/$>140$K steps. \textcolor{ForestGreen}{\textbf{Green contours}} show the fraction of discount-grid values satisfying an approximate contraction inequality; \textcolor{RoyalBlue}{\textbf{Blue lines}} mark calibrated transition-based horizon references (group-mean $K_{\rm ref}$); \textcolor{Gray}{\textbf{Gray curves}} show empirical 90th-quantile references. Trace lengths vary across tasks and steps. More details are in \cref{app:dmc11-proxy}.}
\label{fig:dmc11-trace-tradeoff}
\end{figure}

\subsection{Benchmark Performance (RQ1)}\label{sec:exp-main}

\begin{figure}[t]
    \centering
    \includegraphics[width=\textwidth]{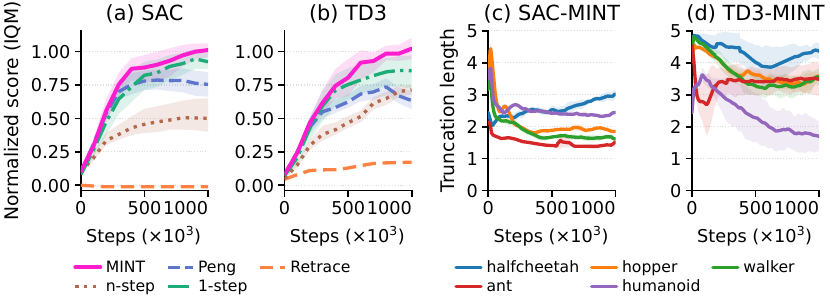}
    \caption{\textbf{Performance and adaptive truncation behavior on MuJoCo.} (a--b) Aggregate normalized scores (IQM) of SAC (fixed-$\alpha$) and TD3 variants over training steps; shaded regions show $95\%$ stratified bootstrap confidence intervals across 5 seeds per task. (c--d) Mean truncation lengths by \texttt{MINT} across five MuJoCo environments, with $\pm1$ standard deviation across seeds. SAC horizons shorten over training, while TD3 horizons remain closer to the cap.}
    \label{fig:iqm}
\end{figure}

\paragraph{MuJoCo.}
\Cref{fig:iqm}(a--b) compares aggregate IQM over five tasks for SAC (fixed-$\alpha$) and TD3 based methods. SAC-\texttt{MINT} has the best aggregate IQM among all variants. SAC-Retrace matches \texttt{MINT} on \texttt{humanoid-v2} and achieves lower returns on the remaining four tasks; Peng's $Q(\lambda)$ and uncorrected $n$-step often underperform one-step SAC. Per-task SAC and TD3 curves are in \cref{fig:mujoco_sac,fig:mujoco_td3}.

\begin{figure}[t]
\centering
\includegraphics[width=\textwidth]{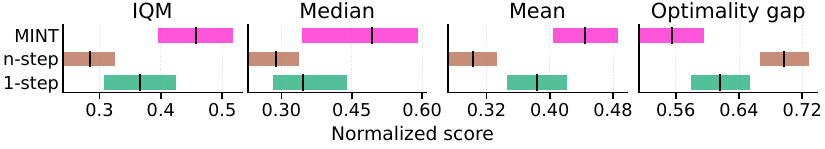}
\caption{\textbf{11-task DMC aggregate performance at 500K steps (5 seeds).} Scores are normalized by $1000$. Rectangles show $95\%$ bootstrap intervals; black ticks mark estimates.}
\label{fig:dmc11-main}
\end{figure}

\paragraph{11-task DMC}
Motivated by \texttt{MINT}'s fixed-policy contraction conditions and survival-based variance analysis, we examine whether adaptive truncation provides a useful balance between value-error propagation and target-sampling variability in actor--critic learning. We evaluate sample efficiency on eleven tasks selected from the DMC Easy \& Medium set used by SimBa \citep{lee2025simba} (\cref{app:dmc-task-selection}), using a finite budget of 500K steps. \texttt{MINT} attains IQM $0.457$ versus $0.366$ for one-step SAC and $0.284$ for $n$-step SAC (\cref{fig:dmc11-main}): relative point-estimate gains of $24.7\%$ and $60.8\%$. Mean returns improve on $10/11$ and $9/11$ tasks, respectively (\cref{tab:dmc11-tasks,fig:dmc11-tasks}). Learning curves and profiles support the aggregate comparison (\cref{fig:dmc11-learning,fig:dmc11-profile}); task-group differences are examined in \cref{sec:exp-task-conditions}.

\paragraph{Atari~100k.}
The DrQ-based \texttt{MINT} variant achieves higher mean HNS ($0.512$ vs.\ $0.465$) and a slightly lower optimality gap ($0.627$ vs.\ $0.631$) than published DrQ($\varepsilon$) (\cref{fig:atari-v3-aggregate,tab:atari-context}). The clearest task-group gains occur in the easy/human-optimal group (\cref{sec:exp-task-conditions}), while full-suite median ($0.152/0.313$) and IQM ($0.251/0.280$) remain lower. A direct extension of DrQ with the TD-error target learned poorly; we therefore modify the implementation as detailed in \cref{app:atari100k}.

\subsection{Adaptive Horizons and Cap Sensitivity (RQ2)}\label{sec:exp-adaptive}

\phantomsection\label{sec:rq2-answer}
\texttt{MINT} selects $\hat K=\min(K_{\max},\hat T_{\beta,\pi})$ without estimating $\tau_{\mathrm{mix}}$. MuJoCo lengths vary across tasks and training, with SAC generally shorter than TD3 (\cref{fig:iqm}(c--d)). On \texttt{Hopper}, returns vary little across caps $\{3,5,10,20\}$, unlike Peng's $Q(\lambda)$ and uncorrected $n$-step (\cref{fig:hopper-cap,app:mujoco-cap}). Atari mean $K$ falls from $6.20$ to $5.05$ as the fraction of matching behavior and current greedy actions rises from $14.1\%$ to $28.1\%$ (\cref{fig:atari-v3-horizons}). DMC temperature effects are task-dependent (\cref{fig:alpha-trunc-diag,fig:dmc11-temperature}).

\begin{wraptable}{r}{0.45\linewidth}
\vspace{-15pt}
\centering\small
\captionsetup{font=small,skip=4pt}
\caption{\textbf{RWRL cap ablation.} Mean $\pm$ std. across three seeds; actuator-noise levels pooled.}
\label{tab:rwrl-fixed-alpha}
\setlength{\tabcolsep}{4pt}
\begin{tabular}{@{}rcc@{}}
\toprule
$K_{\max}$ & $K_{\mathrm{eff}}$ & \shortstack{Stops before cap (\%)} \\
\midrule
5 & $3.9\pm0.1$ & $36.7\pm3.3$ \\
50 & $4.8\pm0.8$ & $99.5\pm0.7$ \\
\bottomrule
\end{tabular}
\vspace{-10pt}
\end{wraptable}

RWRL illustrates the value of leaving room for stopping. At fixed $\alpha=0.05$ and 200K steps, increasing the cap tenfold changes mean length only from $3.9$ to $4.8$, with $99.5\%$ stopping before cap $50$ (\cref{tab:rwrl-fixed-alpha}).
A separate auto-$\alpha$ diagnostic fixes $\pi$ at 1M and varies $\beta$ over 900K, 500K, and 100K.

At cap $50$, mean lengths are $7.06\to14.21\to45.96$ for $\sigma=0.1$ and $10.52\to18.10\to44.61$ for $\sigma=0.3$. For recent $\beta$, cap $50$ cuts only $0.007\%/0.583\%$ of traces; for the oldest $\beta$, these fractions rise to $79.73\%/75.70\%$ (\cref{fig:rwrl-age-survival}). Thus the cap margin needed for stopping to determine most lengths depends on the behavior distribution (\cref{app:rwrl-stopping-hist}).

\WFclear
\subsection{Task-Dependent Benefits (RQ3)}\label{sec:exp-task-conditions}

We group DMC tasks by the fraction of uniform-random episodes containing positive reward, as measured in \cref{app:dmc11-performance}: rare ($<5\%$), occasional ($10$--$50\%$), and frequent ($100\%$). \texttt{MINT} improves mean returns over one-step SAC on all six DMC tasks with rare or occasional positive rewards, and over ten-step SAC on five of the six (\cref{tab:dmc11-tasks}). The largest aggregate gains occur on the occasional-reward \texttt{Finger}/\texttt{Hopper} tasks: IQM $0.453$ versus $0.322/0.157$ for one-step/ten-step SAC (\cref{fig:dmc11-reward}). Both strictly rare tasks improve over one-step, while \texttt{Pendulum} favors ten-step. Frequent-reward tasks retain near-one-step IQM while improving over ten-step. Task-level exceptions and supplementary task and $\lambda$ sweeps are detailed in \cref{sec:additional_exp}.
Under Bellemare's Atari exploration taxonomy \citep{bellemare2016unifying}, \texttt{MINT}'s strongest task-group gains occur in the easy/human-optimal group: mean HNS is $0.689$ versus $0.463$ for published DrQ, the median of game means is $0.669$ versus $0.429$, and game means improve on $7/11$ games (\cref{fig:atari-group-metrics,tab:atari-group-metrics}). The largest gains occur on \texttt{DemonAttack}, \texttt{Boxing}, \texttt{Assault}, and \texttt{CrazyClimber}; their positive contributions to the suite mean are partly offset by lower hard/dense scores and mixed score-exploit results. More details are reported in \cref{app:atari100k}.

\begin{figure}[t]
\centering
\includegraphics[width=\textwidth]{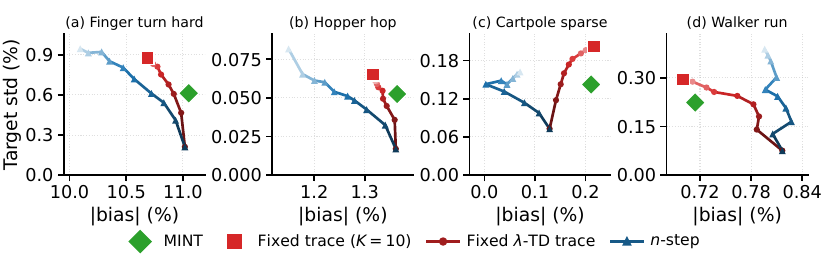}
\caption{\textbf{DMC target diagnostics at 200K (RQ3).} Green: \texttt{MINT}; red: fixed $\lambda$-TD traces (square: $K=10$); blue: $n$-step. Dark-to-light: length 1--10. Five training seeds per task. Axes: absolute aggregated signed bias and target std., relative to $100$; scales differ. Full results: \cref{app:dmc11-bv}.}
\label{fig:dmc11-bv-main}
\end{figure}

\phantomsection\label{sec:rq3-targets}
DMC diagnostics fix the policy and target critic and compare operators on the same behavior-trajectory segments (\cref{app:dmc11-bv}). Pooling current and older behavior checkpoints within each stage and applying stage-specific seed exclusions, \texttt{MINT} has lower target spread than cap-ten $\lambda$-TD in $26/33$ task--checkpoint comparisons across eleven tasks at 200K, 300K, and 500K, and pure ten-step in $27/33$, with median ratios $0.916$ and $0.759$ (\cref{fig:dmc11-spread}). \Cref{fig:dmc11-bv-main} illustrates the 200K diagnostics: spread is lower in all four panels, while absolute aggregated bias is higher in three. A separate five-seed diagnostic reports lower target-variance point estimates than a shuffled-length control in $11/12$ off-policy conditions, preserving the full stopping-length distribution (\cref{app:dmc-reward-retention}). This supports a descriptive benefit from trajectory-dependent stopping, without establishing uniform variance dominance over fixed backups or a causal explanation for control gains. Operator details and counterexamples are provided in \cref{app:dmc11-bv,app:dmc-reward-retention}.

\paragraph{Interpreting adaptive horizons.}
\texttt{MINT} adjusts reward propagation against exposure to off-policy segments and bootstrap estimates. \Cref{fig:dmc11-trace-tradeoff} shows task-dependent lengths across measurements up to and after 140K steps; its conditional references describe this trade-off without certifying an optimal horizon. The shorter early-stage traces in \cref{fig:dmc11-trace-tradeoff} are consistent with greater overlap between the current and collection-time behavior policies, leading to more frequent stopping. As training progresses, policy evolution and replay of transitions collected under earlier policies can increase target--behavior mismatch and produce longer traces. Within the finite 500K-step budget, this evolving policy mixture may help explain the observed temporal differences in trace length.

\begin{figure}[hbtp]
\centering
\includegraphics[width=\textwidth]{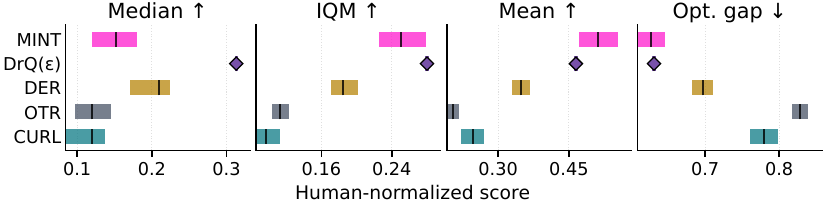}
\caption{\textbf{Atari~100k aggregate performance.} Rectangles: $95\%$ bootstrap intervals; black ticks: estimates. Diamonds: DrQ($\varepsilon$) 100-run estimates \citep[Table~A.1]{schwarzer2023bbf}, without baseline intervals.}
\label{fig:atari-v3-aggregate}
\end{figure}

\subsection{Target and Learned-Critic Evaluation under Stochastic Dynamics (RQ4)}\label{sec:exp-stochastic}

OGBench AntMaze crosses deterministic/stochastic dynamics with \texttt{navigate}/\texttt{stitch} datasets. A fitted Gaussian behavior model defines policy displacements $h$ in standard-deviation units. All operators share a frozen approximate Monte-Carlo critic and finite-rollout reference. At $h=0.5$, \texttt{MINT} has lower RMSE than Retrace on all four datasets (\cref{tab:ogbench-frozen-main}). Its intrinsic variance is lower than a shuffled-$\hat K$ control in seven of eight reported $h\in\{0.5,1\}$ cells (\cref{tab:ogbench-frozen-main,tab:ogbench-frozen}). This control preserves the length distribution but removes trajectory dependence. One-step has the lowest RMSE in most cells; at $h=2$, Retrace also overtakes \texttt{MINT} in the frozen-critic comparison (\cref{app:ogbench-frozen}).

Learned-critic results reveal a propagation trade-off. Retrace has lower RMSE on all evaluated datasets at 100K updates. At 300K, \texttt{MINT} improves on Retrace on the deterministic maze at $h=1$, after slower propagation with $\hat K\approx1.6$; Retrace retains its advantage on the stochastic maze at $h\in\{1,2\}$. The shuffled-$\hat K$ control is at least as accurate and remains stable at $h=2$ on the stochastic maze (\cref{app:ogbench-fqe,tab:ogbench-fqe}).

\begin{table}[H]
\centering\small
\caption{\textbf{OGBench frozen-critic evaluation ($h=0.5$).} Approximate Monte-Carlo critic; 1 paired run per cell over 100 stratified evaluation states. Bold marks the lower value within each metric pair.}
\label{tab:ogbench-frozen-main}
\setlength{\tabcolsep}{4pt}
\begin{tabular}{@{}llcccc@{}}
\toprule
& & \multicolumn{2}{c}{RMSE $\downarrow$} & \multicolumn{2}{c}{Intrinsic variance $\downarrow$} \\
\cmidrule(lr){3-4}\cmidrule(lr){5-6}
Dataset & Dynamics & \texttt{MINT} & Retrace & \texttt{MINT} & Shuffled-$\hat K$ \\
\midrule
Large--navigate & Deterministic & $\mathbf{6.62}$ & $7.76$ & $2.03$ & $\mathbf{1.69}$ \\
Large--stitch & Deterministic & $\mathbf{2.99}$ & $8.85$ & $\mathbf{0.49}$ & $1.01$ \\
Teleport--navigate & Stochastic & $\mathbf{3.82}$ & $6.09$ & $\mathbf{0.33}$ & $1.42$ \\
Teleport--stitch & Stochastic & $\mathbf{2.93}$ & $6.79$ & $\mathbf{1.16}$ & $1.91$ \\
\bottomrule
\end{tabular}
\end{table}

\paragraph{Discussion.}\label{sec:discussion}

The experiments support adaptive backup length as a practical estimator-design choice, with control-performance gains and lower target spread relative to long fixed backups. In RWRL, increasing the cap from 5 to 50 under fixed $\alpha$ changes mean trace length only from 3.9 to 4.8, with $99.5\%$ of traces stopping before cap 50 (\cref{tab:rwrl-fixed-alpha}). In OGBench frozen-critic evaluation, \texttt{MINT} achieves lower RMSE than Retrace on all four datasets at $h=0.5$ and lower intrinsic variance than a shuffled-length control in seven of eight reported $h\in\{0.5,1\}$ conditions (\cref{tab:ogbench-frozen-main,tab:ogbench-frozen}). These findings highlight the practical value of adaptive stopping for horizon selection and target estimation.

\section{Related Work}\label{sec:related}

Multi-step off-policy methods differ in how they handle policy mismatch and preserve the target-policy value function. Retrace constrains trace coefficients to guarantee fixed-policy evaluation under arbitrary behavior policies \citep{munos2016safe}. \citet{harutyunyan2016q} accumulates geometrically weighted TD errors without per-decision corrections, but requires additional policy-mismatch conditions for contraction; our expected operator recovers it when all traces are retained. Peng's method generally converges to a behavior-dependent value function, although control convergence is possible when behavior suitably tracks a greedy policy \citep{kozuno2021revisiting}. \texttt{MINT} preserves the target-policy value function, with separate sufficient contraction conditions for its state-coupled and surrogate operators (\cref{thm:contraction,thm:surrogate-contraction}).
\citet{rowland2020adaptive} analyze trade-offs among update variance, fixed-point bias, and contraction rate; \citet{daley2023trajectory} unify per-decision and trajectory-aware traces under a multistep operator. \texttt{MINT} continues a trace with the complement of the clipped policy ratio used by Retrace, making continuation depend on disagreement and reducing to one-step backups. Its binary masks express the state-coupled estimator's variance bound at the target-policy value function through trace-survival probabilities (\cref{lem:variance}), without establishing variance dominance over Retrace. Importance Resampling resamples replay transitions \citep{schlegel2019importance}, whereas \texttt{MINT} truncates the given trajectory.

Mixing-time analyses offer a complementary perspective on temporal dependence in policy learning \citep{hao2020provably,abbasi2019politex,neu2021online,wei2021learning}, including average-reward methods without oracle mixing-time knowledge \citep{suttle2023beyond}. For on-policy multi-step evaluation, \citet{duan2022policy} relate sufficient look-ahead to temporal dependence and model misspecification (\cref{sec:prelim-mixing}); this does not directly give off-policy guarantees. \texttt{MINT} instead uses policy-mismatch-dependent stopping for discounted TD evaluation, motivated by coupling \citep{meyn2012markov,lindvall1992lectures,johndrow2017coupling}. Behavior-chain mixing uses a same-kernel coalescent coupling; behavior--target first agreement controls state-coupled trace survival (\cref{app:two-couplings}). The model-free surrogate defines a separate stopping process without transition-kernel access or mixing-time estimation.

\section{Conclusion}\label{sec:conclusion}

\texttt{MINT} uses policy mismatch to adapt the backup horizon through binary stopping. Our analysis connects trace survival to variance and gives sufficient conditions for convergence in fixed-policy evaluation, while a model-free stopping rule makes the approach practical. Experiments show that adaptive truncation can improve learning performance and reduce target variability, although the benefits vary across tasks and benchmarks. These results motivate treating backup length as part of estimator design and adapting it to balance long-horizon learning signals with target variability.

\begin{ack}
This work is in part supported by the National Research Foundation of Korea (NRF, RS-2024-00451435(20\%), RS-2024-00413957(10\%), RS-2026-25547694(20\%)), Institute of Information \& communications Technology Planning \& Evaluation (IITP, RS-2025-02305453(10\%), RS-2025-02273157(10\%), RS-2025-25442149(15\%), RS-2021-II211343(15\%)) grant funded by the Ministry of Science and ICT (MSIT), Institute of New Media and Communications (INMAC), and the BK21 FOUR program of the Education, Artificial Intelligence Graduate School Program (Seoul National University), and Research Program for Future ICT Pioneers, Seoul National University in 2026.
\end{ack}

\bibliography{commons/ref_clean}
\bibliographystyle{abbrvnat}

\clearpage
\appendix
\begingroup
\setcounter{tocdepth}{2}
\section*{Contents of the Appendix}
\pdfbookmark[0]{Contents of the Appendix}{mint-appendix-contents}
\makeatletter
\@starttoc{apc}
\makeatother
\endgroup
\clearpage

\makeatletter
\let\mint@appendix@addcontentsline\addcontentsline
\renewcommand{\addcontentsline}[3]{%
  \mint@appendix@addcontentsline{#1}{#2}{#3}%
  \def\mint@contents@extension{#1}%
  \def\mint@contents@toc{toc}%
  \ifx\mint@contents@extension\mint@contents@toc
    \mint@appendix@addcontentsline{apc}{#2}{#3}%
  \fi
}
\makeatother

\crefalias{section}{appendix}
\crefalias{subsection}{appendix}
\begingroup
\raggedbottom
\captionsetup{font=small}
\let\mintAppendixSection\section
\renewcommand{\section}{\FloatBarrier\mintAppendixSection}
\let\mintAppendixSubsection\subsection
\renewcommand{\subsection}{\FloatBarrier\mintAppendixSubsection}
\makeatletter
\c@topnumber=4\relax
\c@bottomnumber=3\relax
\c@totalnumber=7\relax
\renewcommand{\topfraction}{0.95}
\renewcommand{\bottomfraction}{0.85}
\renewcommand{\textfraction}{0.05}
\renewcommand{\floatpagefraction}{0.85}
\setlength{\textfloatsep}{10pt plus 2pt minus 2pt}
\setlength{\floatsep}{8pt plus 2pt minus 2pt}
\setlength{\intextsep}{10pt plus 2pt minus 2pt}
\setlength{\@fptop}{0pt}
\setlength{\@fpsep}{12pt}
\setlength{\@fpbot}{0pt plus 1fil}
\makeatother

\section{Notation and Background}\label{app:notation}

This appendix collects the standard definitions used throughout the paper:
total variation (TV) distance, the coupling lemma, maximal coupling, and
uniform ergodicity. These are the analytic primitives behind the meeting-time
analysis of \cref{thm:nearby} and the contraction analysis of
\cref{thm:contraction}.

\subsection{Total Variation Distance}

For any two probability distributions $\nu_1$ and $\nu_2$ over $\mathcal{S}$, the total variation (TV) distance $\lVert \cdot \rVert_{\mathrm{TV}}$ is
\begin{equation*}
    \left\lVert \nu_1 - \nu_2 \right\rVert_{\mathrm{TV}} \coloneqq \frac{1}{2} \sum_{s \in \mathcal{S}} | \nu_1(s) - \nu_2(s)| = \max_{A \subset \mathcal{S}} |\nu_1(A) - \nu_2(A)|,
\end{equation*}
where the norm $\lVert \cdot \rVert$ corresponds to the $L_1$ metric. By definition, the TV distance takes values in $[0, 1]$.

\subsection{Coupling Lemma and Maximal Coupling}

\begin{lemma}[Coupling Lemma]
\label{lemma:coupling_pre}
Let $\nu_1$ and $\nu_2$ be two probability distributions over a finite space $\mathcal{S}$. For every coupling $(X, Y)$ of $\nu_1$ and $\nu_2$,
\begin{align*}
    \textnormal{P}(X \neq Y) \geq \left\lVert \nu_1 - \nu_2 \right\rVert_{\mathrm{TV}}.
\end{align*}
There exists a coupling attaining equality, called a \textit{maximal coupling}:
\begin{align}\label{eq:maximal_couplinig}
    \textnormal{P}(X \neq Y)  =  \left\lVert \nu_1 - \nu_2 \right\rVert_{\mathrm{TV}}=1- \sum_{s\in\gS}\min(\nu_1(s), \nu_2(s)).
\end{align}
\end{lemma}

Maximal coupling minimizes the probability of disagreement $\textnormal{P}(X \neq Y)$ among all possible couplings of $\nu_1$ and $\nu_2$. Throughout the paper we use the construction \cref{eq:maximal_couplinig} to relate the meeting time of two policy-induced chains to TV distances between their transition kernels.

\subsection{Uniform Ergodicity}

We say the MDP is \textit{uniformly ergodic} with parameter $\alpha_{\mathrm{erg}}$ if there exist a constant $\alpha_{\mathrm{erg}} \in (0, 1)$ and $C_{\mathrm{erg}}>0$ such that
\begin{equation*}
    \max_{s \in \gS}\left\lVert {\gP^\beta}^{(t)}(s, \cdot) - \mu_\beta \right\rVert_{\mathrm{TV}} \leq C_{\mathrm{erg}}(1 - \alpha_{\mathrm{erg}})^t
\end{equation*}
for all $t \in \mathbb{N}$. Uniform ergodicity is a standard mild assumption in the analysis of perturbed Markov chains \citep{mitrophanov2005sensitivity,solan2003perturbed,johndrow2017error}.

\subsection{Data-Processing Inequality for Total Variation}\label{app:data_ineq}

\paragraph{Lemma (TV contraction under Markov kernel).}
Let $P$ and $Q$ be two probability measures over a measurable space $(\mathcal{X}, \mathcal{F})$, and let $f: \mathcal{X} \to \mathcal{Y}$ be a stochastic map, i.e., a Markov kernel such that $f(\cdot \mid x)$ is a probability distribution over $\mathcal{Y}$ for each $x \in \mathcal{X}$. Then the push-forward measures $P_Y, Q_Y$ on $\mathcal{Y}$ defined by
\[
P_Y(B) = \int_{\mathcal{X}} f(B \mid x) \, dP(x), \quad
Q_Y(B) = \int_{\mathcal{X}} f(B \mid x) \, dQ(x)
\]
satisfy
\[
\mathrm{TV}(P_Y, Q_Y) \le \mathrm{TV}(P, Q).
\]

\paragraph{\revr{Proof.}}
\revr{Let $\mu=P-Q$, so $\mu(\mathcal X)=0$. For any measurable $B\subseteq\mathcal Y$, the function $g_B(x)=f(B\mid x)$ lies in $[0,1]$. The variational form $\sup_{0\le g\le1}|\int g\,d\mu|=\mathrm{TV}(P,Q)$ therefore gives}
{\color{black}
\[
\mathrm{TV}(P_Y,Q_Y)
=\sup_B\left|\int g_B\,d(P-Q)\right|
\le\mathrm{TV}(P,Q).
\]
}

\paragraph{Interpretation.}
Applying a stochastic transformation can only reduce, not increase, the distinguishability of two distributions. In RL, this explains why the discrepancy between next-state distributions under two policies is bounded by their action-level difference:
\[
\mathrm{TV}\!\left( \sum_a \pi(a \mid s) \mathcal{P}(\cdot \mid s,a),\; \sum_a \beta(a \mid s) \mathcal{P}(\cdot \mid s,a) \right)
\le
\mathrm{TV}(\pi(\cdot \mid s), \beta(\cdot \mid s)).
\]
This principle underlies the use of action-level disagreement probabilities $\hat p_i$ in \texttt{MINT} as a proxy for policy divergence propagated through the dynamics.

\subsection{Standing Assumptions}\label{app:assumptions}

This subsection states the kernel-level TV bound and cross-Doeblin condition used by the supporting results. The assumption usage table (\cref{tab:assumption-usage}) distinguishes these hypotheses from those of the main contraction theorem and the action-level surrogate theorem.

\begingroup\color{black}
\begin{assumption}[Kernel-level TV bound]\label{assumption:2}
For the fixed policy pair, define the kernel discrepancy by
\[
d:=\sup_{s\in\mathcal S}d(s),\qquad
d(s):=\|\mathcal P^\beta(s,\cdot)-\mathcal P^\pi(s,\cdot)\|_{\mathrm{TV}}.
\]
Thus $d\in[0,1]$ and data processing gives $d\le D:=\sup_s\mathrm{TV}(\beta(\cdot\mid s),\pi(\cdot\mid s))$. This comparison concerns the actual supremum defined here; an arbitrary looser numerical upper bound on $d$ can exceed $D$.
\end{assumption}
\endgroup

\begin{assumption}[Cross-Doeblin]\label{ass:cross-doeblin}\label{assumption:3}
There exists $\rho \in (0, 1-d)$ such that for every pair of states $(s,s') \in \mathcal{S}\times\mathcal{S}$ and every policy pair $(\beta,\pi)$,
\[
\big\lVert \mathcal{P}^\beta(s,\cdot) - \mathcal{P}^\pi(s',\cdot) \big\rVert_{\mathrm{TV}} \le 1 - \rho.
\]
\end{assumption}

The cross-Doeblin condition requires the existence of a uniform overlap $\rho\in(0,1-d)$ such that $\lVert \mathcal{P}^\beta(s,\cdot) - \mathcal{P}^\pi(s',\cdot) \rVert_{\mathrm{TV}}\le 1-\rho$ for all state pairs and all policies. Combined with the maximal coupling \cref{eq:maximal_couplinig}, it ensures that conditional on disagreement at step $t$, the coupled processes meet at step $t+1$ with probability at least $\rho$, which is what produces the geometric meeting-time tail used in \cref{cor:geometric}.

\paragraph{\rev{Where each assumption is used.}}
\revr{\Cref{tab:assumption-usage} distinguishes the state-coupled analysis from the finite-state, fixed-policy surrogate theorem. The latter uses no deterministic-dynamics or cross-Doeblin hypothesis, but requires its own $\chi_\lambda$ sufficient condition. The one-step DPI proposition is not a transfer theorem for contraction.}

\begin{table}[!htbp]
\centering
\small
\caption{\revr{Assumption usage by result. No listed result requires deterministic dynamics. State-coupled and action-level surrogate results have separate sufficient conditions.}}
\label{tab:assumption-usage}
\begingroup\color{black}
\setlength{\tabcolsep}{4pt}
\renewcommand{\arraystretch}{1.2}
\begin{tabular}{@{}p{0.25\textwidth}p{0.30\textwidth}p{\dimexpr0.45\textwidth-4\tabcolsep\relax}@{}}
\toprule
\raggedright\textbf{Result} & \raggedright\textbf{Required assumptions} & \raggedright\textbf{Additional conditions} \tabularnewline
\midrule
\multicolumn{3}{@{}l}{\textbf{State-coupled results and chain bounds}} \tabularnewline
\addlinespace[2pt]
\raggedright\cref{lem:variance}\newline Variance
& \raggedright Adapted joint law
& \raggedright Bounded TD error \tabularnewline
\raggedright\cref{thm:contraction}\newline Contraction
& \raggedright State coupling
& \raggedright $\kappa<1$ in \cref{eq:contraction-condition} \tabularnewline
\raggedright\cref{cor:geometric}\newline Closed form
& \raggedright Cross-Doeblin;\newline kernel TV $d$
& \raggedright --- \tabularnewline
\raggedright\cref{thm:nearby}\newline Disagreement
& \raggedright Cross-Doeblin;\newline kernel TV $d$
& \raggedright --- \tabularnewline
\raggedright\cref{thm:delta_bar}\newline Perturbation
& \raggedright Kernel TV $d$;\newline\cref{ass:beta-doeblin}
& \raggedright Stationary laws; explicit $b_\pi(t)$ \tabularnewline
\raggedright\cref{prop:beta-coalescence}\newline Behavior mixing
& \raggedright Same-kernel\newline coalescent coupling
& \raggedright Stationary law \tabularnewline
\raggedright\cref{thm:truncation_length}\newline State horizon
& \raggedright Cross-Doeblin
& \raggedright Initial-step overlap for the $1/\rho$ bound \tabularnewline
\addlinespace[5pt]
\multicolumn{3}{@{}l}{\textbf{Action-level surrogate and control}} \tabularnewline
\addlinespace[2pt]
\raggedright\cref{thm:surrogate-contraction}\newline Surrogate contraction
& \raggedright Finite state space;\newline fixed policies
& \raggedright Independent masks; $\kappa_{\chi,L}<1$ \tabularnewline
\raggedright\cref{thm:online_t4}\newline Online control
& \raggedright Finite state space;\newline Markov traces
& \raggedright (A1)--(A5) of \cref{proof:online_t4} \tabularnewline
\addlinespace[5pt]
\multicolumn{3}{@{}l}{\textbf{One-step relationships}} \tabularnewline
\addlinespace[2pt]
\raggedright\cref{prop:dpi-stochastic}\newline One-step link
& \raggedright Data processing
& \raggedright --- \tabularnewline
\raggedright\cref{cor:practical-expected}\newline Averaged error
& \raggedright Pointwise bound
& \raggedright Any starting distribution \tabularnewline
\bottomrule
\end{tabular}
\par\smallskip
\parbox{\textwidth}{\footnotesize\raggedright Cross-Doeblin refers to \cref{ass:cross-doeblin}; kernel TV $d$ is defined in \cref{assumption:2}. A dash means no additional condition is listed beyond the stated setup.}
\endgroup
\end{table}

\subsection{Cross-Doeblin and Two TV Quantities}\label{app:two-tv}

\paragraph{Two TV quantities, two roles.}
By the data-processing inequality (\cref{app:data_ineq}), the kernel-level TV bound $d$ from \cref{assumption:2} satisfies $d \le D := \sup_s D_s$. Both quantities measure off-policy discrepancy but at different ``levels'':
\begin{itemize}
\item $d$ enters the cross-Doeblin assumption $\rho \in (0, 1-d)$, controlling the kernel-level meeting rate.
\item $\bar D_{\mathrm{wgt}}$ enters the contraction condition \cref{eq:contraction-M}, controlling the per-step weight deviation in the operator analysis.
\end{itemize}
\revr{$D_s$ is action-level policy mismatch, while $\bar D_{\mathrm{wgt}}$ weights it by the state-coupled survival process. The implemented surrogate has its own survival process; its contraction correction weights $\chi_\lambda(s)$ instead (\cref{thm:surrogate-contraction}). These weighted quantities can differ.}

\paragraph{Justification of the action-level approximation.}
\revr{For transitions from a shared state $s$, define $d(s)=\mathrm{TV}(\mathcal P^\beta(\cdot\mid s),\mathcal P^\pi(\cdot\mid s))$. Data processing bounds $d(s)$ by $D_s$. The implementation samples an action-dependent keep probability whose mean is $D_s$, as follows; this does not identify it with the disagreement probability of a coupled pair at different states.}

\paragraph{(i) Always valid.}
Since $\min\{1,\pi/\beta\}\in[0,1]$, we have $\hat p_i\in[0,1]$ regardless of the relative magnitudes of $\pi(a_i\mid s_i)$ and $\beta(a_i\mid s_i)$, including the unbounded-ratio regime where $\pi/\beta>1$.

\paragraph{(ii) Unbiased estimator of the action-level policy TV.}
Taking the expectation over $a_i\sim\beta(\cdot\mid s_i)$ and using the variational identity for total variation,
\begin{align}
\E_{a\sim\beta}\!\left[\min\!\left\{1,\frac{\pi(a\mid s)}{\beta(a\mid s)}\right\}\right]
&\;=\; \sum_a \min\{\beta(a\mid s),\,\pi(a\mid s)\}
\;=\; 1 - \mathrm{TV}\!\bigl(\beta(\cdot\mid s),\pi(\cdot\mid s)\bigr),\\
\E_{a\sim\beta}[\hat p_i \mid s_i]
&\;=\; \mathrm{TV}\!\bigl(\beta(\cdot\mid s_i),\pi(\cdot\mid s_i)\bigr) \;=\; D_{s_i}.
\end{align}
\revr{Thus, $\hat p_i$ is an unbiased sample-level estimate of $D_{s_i}$. Its conditional mean, rather than each realized sample, upper-bounds $d(s_i)$, the one-step kernel TV from a shared state.}

\paragraph{(iii) Fully model-free.}
$\hat p_i$ depends only on the policy ratio at the sampled action and never queries the environment transition kernel, so it is computable in any actor-critic implementation (cf.\ \cref{alg:t4}).

\begingroup\color{black}
\paragraph{(iv) One-step data processing under arbitrary dynamics.}
\begin{proposition}[One-step surrogate and kernel TV]\label{prop:dpi-stochastic}
For an arbitrary transition kernel, let $d(s):=\mathrm{TV}(\mathcal P^\beta(\cdot\mid s),\mathcal P^\pi(\cdot\mid s))$. Then
\[
\mathbb E_{a\sim\beta(\cdot\mid s)}[\hat p(s,a)]=D_s\ge d(s).
\]
If the next states are maximally coupled from a shared state $s$, their one-step disagreement probability is $d(s)$. Equality $D_s=d(s)$ holds, in particular, for deterministic dynamics with an injective action-to-next-state map.
\end{proposition}
\begin{proof}
The expectation identity is proved in (ii). Both next-state laws are the push-forwards of the two action laws through the common kernel $a\mapsto\mathcal P(\cdot\mid s,a)$, so data processing gives $d(s)\le D_s$. Maximal coupling identifies $d(s)$ with one-step disagreement from that shared state. An injective deterministic map preserves TV, giving the stated sufficient equality case.
\end{proof}
This proposition is an expectation-level, one-step statement. It neither gives a samplewise upper bound on kernel disagreement nor orders the multi-step stopping times once the two state processes differ. The action-dependent surrogate contraction theorem is \cref{thm:surrogate-contraction}; it uses the surrogate's own survival quantities and the negative-mass fraction $\chi_\lambda$.
\endgroup

\subsection{Mixing-Time Parameters and On-Policy Look-Ahead}\label{app:mixing-background}

\paragraph{Minorization-based mixing parameter.}
The parameter $\tau_*^\beta\ge1$ in \cref{eq:mixing-bound} belongs to
the common data-generating and target chain in the on-policy analysis
of \citet{duan2022policy}. Their mixing condition requires a probability
measure $\nu_\beta$ and a finite constant $C_\nu\ge0$ such that
\[
\begin{gathered}
\mathcal P^\beta(s,B)\ge\frac{1}{\tau_*^\beta}\nu_\beta(B)
\quad\text{for all $s$ and measurable $B$},\\
\left\|\frac{d\nu_\beta}{d\mu_\beta}\right\|_\infty\le1+C_\nu.
\end{gathered}
\]
Thus all one-step transition distributions contain a common probability
component of weight $1/\tau_*^\beta$. This is a minorization condition,
rather than the TV-threshold definition of $\tau_{\mathrm{mix}}^\beta$
in \cref{sec:prelim-mixing}; the two parameters are not identical by
definition \citep[Section~3.1.1, Eq.~(11)]{duan2022policy}.

\paragraph{Scope of the look-ahead condition.}
\citet{duan2022policy} analyze multi-step kernel least-squares TD estimates.
Their error bounds distinguish martingale fluctuations from a
Bellman-residual contribution affected by temporal dependence.
In the uniformly bounded-reward setting, consistent with our assumption
$|r(s,a)|\le r_{\max}$, \cref{eq:mixing-bound} is sufficient for their
stated finite-sample bounds, subject to their mixing condition,
boundedness and regularity assumptions on the kernel and its
eigenfunctions, and sample-size requirements
\citep[Sections~3.1.1 and~3.3.2, Eq.~(29b)]{duan2022policy}.
The look-ahead condition alone is not an off-policy guarantee.

\paragraph{Behavior-chain mixing versus off-policy stopping.}
A coalescent coupling of two behavior chains from different initial
states bounds $\tau_{\mathrm{mix}}^\beta$. The behavior--target first
agreement instead controls \texttt{MINT} trace survival. These two
couplings and their stopping times are distinguished in
\cref{app:two-couplings}.

\section{Proofs}\label{app:proofs}

\subsection{Proof of the Variance Lemma (\texorpdfstring{\cref{lem:variance}}{Variance lemma})}\label{app:variance-lemma}
\begingroup\color{black}
\paragraph{Joint sampling and filtration.}
Let $\mathcal H_t$ contain both state histories, all past actions and rewards, and coupling randomness revealed through time $t$, before sampling the current behavior action. At policy-generated steps, draw $a_t\sim\beta(\cdot\mid s_t)$ conditionally on $\mathcal H_t$, and set $\mathcal G_t=\mathcal H_t\vee\sigma(a_t)$. For $t\ge1$, the arrival-state mask $A_t$ and cumulative mask $W_t$ are $\mathcal H_t$-measurable, hence $\mathcal G_t$-measurable; $W_0=1$ is deterministic. The enlarged history contains no information from the next reward or transition beyond that allowed by the environment law. We require
\[
\mathcal L(r_t,s_{t+1}\mid\mathcal G_t)
=\mathcal L(r_t,s_{t+1}\mid s_t,a_t).
\]
This compatibility is part of the construction, not a consequence of marginal coupling alone. For example, let $\Gamma_t(u,v)$ be a maximal coupling of $\mathcal P^\beta(s_t,\cdot)$ and $\mathcal P^\pi(S_t^\pi,\cdot)$, given $\mathcal H_t$. First draw $a_t\sim\beta(\cdot\mid s_t)$ and the behavior reward and next state $u$ from the environment. Then draw the auxiliary next state $v$ from
$\Gamma_t(u,v)/\mathcal P^\beta(s_t,u)$ on positive-denominator rows. Averaging over $a_t$ recovers $\Gamma_t$ and both policy-induced state marginals; the behavior transition law given $\mathcal G_t$ is unchanged. Zero-denominator rows can be assigned any conditional law, since they have zero probability under a policy-generated step. If the initial pair $x$ is prescribed, use the induced conditional initial coupling, with a valid behavior transition on any null-support extension. All $\Pr_x(T>t)$ refer to that specified initialization.

\paragraph{Conditional and marginal coupling.}
For $i\ge1$, let $p_i:=\Pr(A_i=1\mid\mathcal H_{i-1})$. At a policy-generated step, conditional maximal coupling gives
\[
p_i=\mathrm{TV}\!\left(\mathcal P^\beta(S_{i-1}^\beta,\cdot),
\mathcal P^\pi(S_{i-1}^\pi,\cdot)\right).
\]
When the initial action is prescribed, the first-step law is instead the conditional initial coupling specified above.
Conditional maximal coupling at each step does not imply maximal coupling of the unconditional time-$i$ marginals. In general,
\[
\mathrm{TV}\!\left(\mathcal L(S_i^\beta),\mathcal L(S_i^\pi)\right)
\le\Pr(A_i=1).
\]

\paragraph{Survival functional and its bounds.}
For $T=T_{\beta,\pi}$, summing the geometric series along each stopped trajectory gives
\[
M_x(\gamma,\lambda)
=\E_x\!\left[\frac{1-(\gamma\lambda)^T}{1-\gamma\lambda}\right]
\le\min\!\left\{\E_x[T],\frac1{1-\gamma\lambda}\right\},
\]
with $(\gamma\lambda)^\infty=0$. For a distribution of starting pairs, $M$ without subscript denotes the corresponding average of $M_x$. The contraction bound in \cref{thm:contraction} instead retains each starting pair and takes a supremum; this average cannot replace that supremum.

\paragraph{Martingale and fixed point.}
Let $\delta_t=\delta_t^{Q^\pi}$, $W_t=\prod_{i=1}^tA_i=\mathbf1\{T>t\}$, and $\alpha=\gamma\lambda$.
The Bellman identity and the conditional environment law give
\[
\E_x[\delta_t\mid\mathcal G_t]
=(\mathcal T^\pi Q^\pi)(s_t,a_t)-Q^\pi(s_t,a_t)=0.
\]
The mask $W_t$ is $\mathcal G_t$-measurable, hence $\E_x[W_t\delta_t]=0$.
The deterministic bound $|\delta_t|\le C:=\|r\|_\infty+(1+\gamma)\|Q^\pi\|_\infty$ and $\alpha<1$ justify exchanging the sum and expectation. Thus $\E_x[\hat\Delta]=0$ and $\mathcal R_{p,\lambda}Q^\pi=Q^\pi$.

\paragraph{Cross terms and idempotence.}
For $u<t$, $W_u,W_t,\delta_u$ are $\mathcal G_t$-measurable. Therefore
\[
\E_x[W_uW_t\delta_u\delta_t]
=\E_x[W_uW_t\delta_u\E_x[\delta_t\mid\mathcal G_t]]=0.
\]
This orthogonality applies to other adapted weights as well. For the binary masks, $W_t^2=W_t$ additionally yields
\begin{equation}\label{eq:variance-explicit}
\begin{aligned}
\operatorname{Var}_x(\hat\Delta)=\E_x[\hat\Delta^2]
&=\sum_{t\ge0}\alpha^{2t}\E_x[W_t\delta_t^2]\\
&\le C^2\sum_{t\ge0}(\gamma^2\lambda^2)^t\Pr_x(T>t)
=C^2M_x(\gamma^2,\lambda^2).
\end{aligned}
\end{equation}
One can first expand finite sums and then pass to the limit: $|\hat\Delta|\le C/(1-\alpha)$ gives dominated convergence. Averaging over a specified starting distribution gives the corresponding $C^2M(\gamma^2,\lambda^2)$ bound. No claim of universal variance dominance over clipped trace methods follows from this calculation.
\endgroup

\subsection{Proof of \texorpdfstring{\cref{thm:contraction}}{Main contraction}}\label{app:contraction-proof}
\label{proof:contraction}\label{proof:contration}
\begingroup\color{black}
The normalized policy-TV average in \cref{sec:theory} is discounted and survival-weighted through $M_x,N_x$ of \cref{def:M}; $\bar D_{\mathrm{wgt}}$ takes a supremum over starting pairs, not individual times. By definition,
\[
N_x=D_{\mathrm{wgt}}^{\,x}(M_x-1)\le\bar D_{\mathrm{wgt}}(M_x-1),
\qquad \bar D_{\mathrm{wgt}}\le\sup_sD_s=:D\le1.
\]
Fix a starting pair $x$ and use the joint sampling conditions of \cref{sec:t4}. Set $W_t=\prod_{i=1}^tA_i$, $W_0=1$. The arrival-state mask $A_t$, $W_{t-1}$, and $s_t$ are known before $a_t$ is sampled; thus they are $\mathcal H_t$-measurable. The conditional action law is $\beta(\cdot\mid s_t)$. We retain $A_t$ in this calculation instead of replacing it by a conditional mean taken with respect to a different filtration.

\paragraph{Signed coefficients.}
For $q=Q-Q'$, expanding the TD-error sum and telescoping gives
\[
(\mathcal R_{p,\lambda}Q-\mathcal R_{p,\lambda}Q')(x)
=\sum_{t\ge1}\gamma^t\lambda^{t-1}\E_x\!\left[
W_{t-1}\sum_b\bigl(\pi(b\mid s_t)-\lambda A_t\beta(b\mid s_t)\bigr)q(s_t,b)\right].
\]
The conditional environment law justifies the shift of the bootstrap term to the next visited state; the pre-action conditioning justifies averaging $q(s_t,a_t)$ using $\beta$. Let $w_{x;y,b}$ be the resulting signed coefficients and define
\begin{equation}\label{eq:kappa-def}
b_x:=\sum_{y,b}|w_{x;y,b}|.
\end{equation}
For any $a\in\{0,1\}$,
\begin{equation}\label{eq:abs-decomp}
\begin{aligned}
\sum_b|\pi(b\mid s)-\lambda a\beta(b\mid s)|
&=1-\lambda a+2\sum_b[\lambda a\beta(b\mid s)-\pi(b\mid s)]_+\\
&\le1-\lambda a+2\lambda aD_s.
\end{aligned}
\end{equation}
Indeed, $\lambda a\le1$ and $[\lambda a\beta-\pi]_+\le\lambda a[\beta-\pi]_+$, whose action sum is $\lambda aD_s$. Consequently,
\begin{equation}\label{eq:kappa-decomp}
b_x\le M_{1,x}+2\Delta_{\lambda,x},
\end{equation}
where
\[
M_{1,x}:=\sum_{t\ge1}\gamma^t\lambda^{t-1}\E_x[W_{t-1}(1-\lambda A_t)],\qquad
\Delta_{\lambda,x}:=\sum_{t\ge1}(\gamma\lambda)^t\E_x[W_tD_{s_t}]=N_x.
\]
The survival identity $W_t=W_{t-1}A_t$ gives
\[
M_{1,x}=\gamma M_x-(M_x-1)=1-(1-\gamma)M_x.
\]
Combining the two terms proves
\begin{equation}\label{eq:kappa-bound}
b_x\le\kappa_x:=1-(1-\gamma)M_x+2N_x
\le1-(1-\gamma)M_x+2\bar D_{\mathrm{wgt}}(M_x-1).
\end{equation}
At $\lambda=0$ the same formula follows directly from $\mathcal R_{p,0}=\mathcal T^\pi$; equivalently, use the $t=1$ term in the telescoping expression. Consequently, for all bounded $Q,Q'$ and every starting pair $x$,
\begin{equation}\label{eq:state-pointwise-bound}
|\mathcal R_{p,\lambda}Q(x)-\mathcal R_{p,\lambda}Q'(x)|
\le\kappa_x\|Q-Q'\|_\infty.
\end{equation}
Taking the supremum over $x$ proves the claimed Lipschitz bound. The fixed-point identity of \cref{lem:variance} gives uniqueness and convergence when this supremum is below one.

Since $1\le M_x\le(1-\gamma\lambda)^{-1}$ uniformly,
\begin{equation}\label{eq:contraction-M}
\sup_x\{1-(1-\gamma)M_x+2N_x\}<1
\quad\Longleftrightarrow\quad
\sup_x\frac{2N_x}{M_x}<1-\gamma.
\end{equation}
The uniform bounds on $M_x$ preserve a strictly positive margin in both directions. Substituting $N_x\le\bar D_{\mathrm{wgt}}(M_x-1)$ gives the sufficient condition
\begin{equation}\label{eq:contraction-TV-condition}
\sup_x\left\{2\bar D_{\mathrm{wgt}}\frac{M_x-1}{M_x}\right\}<1-\gamma.
\end{equation}
This uses the survival-weighted identity above without requiring a bound on the TV average at each individual time. A value of $M$ averaged over an initial distribution cannot replace the supremum in \cref{eq:contraction-condition}. This proof concerns the state-coupled operator; replacing $p_i$ by $\hat p_i$ is not used. The action-dependent surrogate requires its own negative-weight bound, developed in \cref{app:surrogate-contraction}.

\paragraph{Connection to the $\lambda$-dependent effective horizon.}
For every starting pair,
\begin{equation}\label{eq:M-vs-K}
\begin{aligned}
M_x&\le K_\lambda^{\,x}:=\min\!\left(\E_x[T],\frac1{1-\gamma\lambda}\right),\\
K_\lambda&:=\sup_xK_\lambda^{\,x}\le\frac1{1-\gamma\lambda}.
\end{aligned}
\end{equation}
This follows from the pointwise inequality $\sum_{t=0}^{T-1}(\gamma\lambda)^t\le\min(T,(1-\gamma\lambda)^{-1})$. Since $(u-1)/u$ is increasing for $u\ge1$, a uniform sufficient condition is
\begin{equation}\label{eq:contraction-K}
2\bar D_{\mathrm{wgt}}\frac{K_\lambda-1}{K_\lambda}<1-\gamma.
\end{equation}
The scalar $K_\lambda$ here is a uniform bound over starting pairs, not an initial-distribution average. When the survival law is the same for all starting pairs, the formulas reduce to the corresponding scalar expressions.
\hfill\qed
\endgroup

\begin{figure}[!htbp]
\centering
\captionsetup{font=small,labelfont=bf,justification=raggedright,singlelinecheck=false,skip=5pt}
\includegraphics[width=0.82\linewidth]{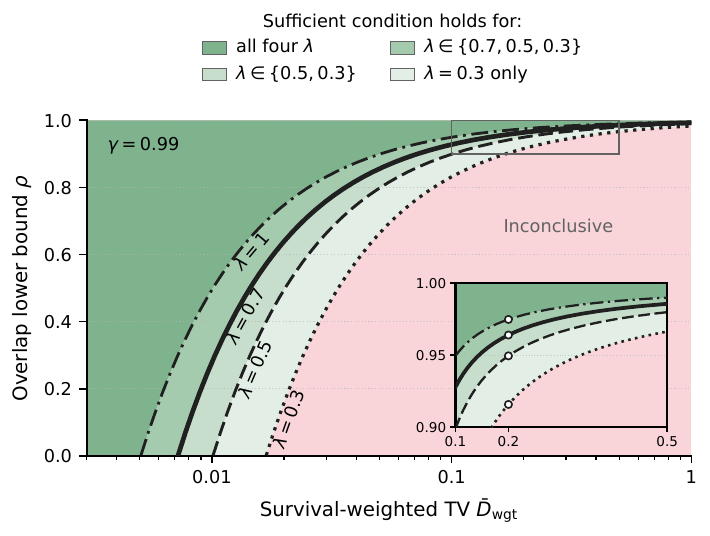}
\caption{\textbf{Geometric-overlap specialization at $\gamma=0.99$.}
Under the cross-Doeblin and uniform initial-step overlap hypotheses of
\cref{cor:geometric}, $\gamma[1+2\bar D_{\mathrm{wgt}}\lambda(1-\rho)]<1$
ensures contraction of the fixed-policy state-coupled operator.
Dark-to-light green bands identify the satisfying subsets of
$\lambda\in\{1,0.7,0.5,0.3\}$ specified in the legend.
Pink is inconclusive for all four displayed values; it does not indicate
divergence. For each $\lambda$, its curve marks equality and the condition
holds strictly above it. The inset enlarges the high-overlap region; open markers
are hypothetical boundary points at $\bar D_{\mathrm{wgt}}=0.2$.
The comparison holds the summary coordinate $\bar D_{\mathrm{wgt}}$ fixed.
These state-coupled conditions do not certify action-level surrogate
contraction or deep-RL convergence.}
\label{fig:phase-diagram}
\end{figure}

\subsection{Closed-Form Corollary and Its Proof}\label{app:cor-closed-form}

\begin{corollary}[Geometric meeting time — closed-form contraction]\label{cor:geometric}
Suppose \cref{ass:cross-doeblin} holds with parameter $\rho \in (0,1-d)$.
\revr{Also require $\Pr_x(A_1=0)\ge\rho$ uniformly over the prescribed starting pairs. This initial-step condition is needed when conditioning on the behavior's initial action; it is not implied by overlap of policy-averaged kernels alone.}
\revr{Then, for every starting pair $x$, $\Pr_x(T_{\beta,\pi}>t)\le(1-\rho)^t$, so $M_x\le[1-\gamma\lambda(1-\rho)]^{-1}$ uniformly.}
\revr{The uniform TV-based sufficient condition of \cref{thm:contraction} is ensured by the single-product condition}
\[
\gamma\bigl(1 + 2\bar D_{\mathrm{wgt}}\,\lambda(1-\rho)\bigr) \;<\; 1.
\]
None of the factors $\bar D_{\mathrm{wgt}}$, $\lambda$, $(1-\rho)$ needs to be small individually; only their product must be controlled.
\end{corollary}

\revr{\Cref{ass:cross-doeblin} asserts $\|\mathcal{P}^\beta(s,\cdot)-\mathcal{P}^\pi(s',\cdot)\|_{\mathrm{TV}} \le 1-\rho$ uniformly over state pairs. Conditional maximal coupling gives a meeting probability of at least $\rho$ at subsequent policy-generated transitions. Combining this with the explicit initial-step condition and iterating gives $\Pr_x(T_{\beta,\pi}>t)\le(1-\rho)^t$ for every $x$. Geometric summation gives}
\[
\textcolor{black}{M_x} \;\le\; \sum_{t\ge 0}(\gamma\lambda(1-\rho))^t \;=\; \frac{1}{1-\gamma\lambda(1-\rho)}.
\]
\revr{Writing $U=[1-\gamma\lambda(1-\rho)]^{-1}$, we have $(M_x-1)/M_x\le(U-1)/U=\gamma\lambda(1-\rho)$ uniformly in $x$. Substitution into \cref{eq:contraction-TV-condition} gives the sufficient condition}
\[
\gamma\bigl(1+2\bar D_{\mathrm{wgt}}\lambda(1-\rho)\bigr) \;<\; 1.
\]
\hfill\qed

\begingroup\color{black}
\subsection{Time-averaged disagreement bound}\label{proof:nearby}

\begin{lemma}[Time-averaged disagreement; nearby chains]\label{thm:nearby}
Consider a co-adapted coupling of the $\beta$ and $\pi$ state processes, with conditional maximal coupling at each policy-generated transition. Write $\mathcal F_k$ for the joint history through time $k$ before the next transition, $A_k=\mathbf1\{S_k^\beta\ne S_k^\pi\}$, and $u_k=\E[A_k]$. Under \cref{assumption:2,assumption:3}, set $\eta=\rho+d\in(0,1)$. For any initial joint law at time $1$ and every integer $t\ge1$,
\begin{equation}\label{eq:time-averaged}
\frac1t\sum_{k=1}^t u_k
\le \frac d\eta+
\frac{1-(1-\eta)^t}{\eta t}\left(u_1-\frac d\eta\right).
\end{equation}
\end{lemma}

\begin{proof}
Conditional maximal coupling and the same-state kernel bound give
\[
\Pr(A_{k+1}=1\mid\mathcal F_k)
\le d(1-A_k)+(1-\rho)A_k.
\]
Indeed, on $A_k=0$ both kernels start at the same state, whereas on $A_k=1$ the cross-state overlap is at least $\rho$. Taking expectations yields the scalar recursion
\begin{equation}\label{eq:disagreement-recursion}
u_{k+1}\le d+(1-\rho-d)u_k.
\end{equation}
Let $r=1-\eta\in(0,1)$ and $q_*=d/\eta$, so $d+rq_*=q_*$. Subtracting $q_*$ and iterating the inequality, using $r\ge0$, gives
\[
u_k-q_*\le r^{k-1}(u_1-q_*).
\]
This remains valid when $u_1<q_*$. Summing for $k=1,\ldots,t$ and using $\sum_{j=0}^{t-1}r^j=(1-r^t)/\eta$ proves \cref{eq:time-averaged}.
\end{proof}

This recursion gives the expectation bound associated with the nearby-chain analysis of \citet[Theorem~9]{johndrow2017coupling}, without a Poisson-equation construction. Its implication is
\[
\limsup_{t\to\infty}\frac1t\sum_{k=1}^t\E[A_k]\le\frac d{\rho+d}.
\]
The correction in the displayed upper bound has magnitude at most $|u_1-q_*|/(\eta t)$, hence is $O(t^{-1})$ for fixed $(d,\rho)$. This does not assert convergence of the actual average to $q_*$, or monotone decay: the upper bound increases toward $q_*$ if $u_1<q_*$. The first-agreement survival bound is proved separately in \cref{thm:truncation_length}; it is not a consequence of the Cesaro bound.
\endgroup

\begingroup\color{black}
\subsection{Perturbed Markov chain bound}
\label{proof:delta_bar}

\begin{assumption}[One-step Doeblin condition for the behavior kernel]\label{ass:beta-doeblin}
For the fixed behavior kernel, there exists $a_\beta\in(0,1)$ such that
\begin{equation}\label{eq:beta-doeblin}
\delta(\mathcal P^\beta):=
\sup_{s,s'}\|\mathcal P^\beta(s,\cdot)-\mathcal P^\beta(s',\cdot)\|_{\mathrm{TV}}
\le1-a_\beta.
\end{equation}
\end{assumption}

This is the one-step Doeblin condition in \citet[Assumption~1]{johndrow2017coupling}. It implies
\[
\|\nu\mathcal P^\beta-\nu'\mathcal P^\beta\|_{\mathrm{TV}}
\le(1-a_\beta)\|\nu-\nu'\|_{\mathrm{TV}}
\]
for probability laws $\nu,\nu'$. It is stronger than merely specifying an exponential envelope with arbitrary $C_{\mathrm{erg}}$ as in \cref{app:notation}; $a_\beta$ is not identified with that envelope's $\alpha_{\mathrm{erg}}$. The assumption is local to the perturbation results below and is not an additional hypothesis of \cref{thm:contraction}.

\begin{lemma}[Stationary perturbation bound]\label{lemma:two_stationary}
Under \cref{assumption:2,ass:beta-doeblin}, let $\mu_\beta$ and $\mu_\pi$ be stationary laws of $\mathcal P^\beta$ and $\mathcal P^\pi$, respectively. Then
\[
\|\mu_\beta-\mu_\pi\|_{\mathrm{TV}}\le\min\{1,d/a_\beta\}.
\]
\end{lemma}

\begin{proof}\label{proof:lemma_two_stationary}
Stationarity, the triangle inequality, and the one-step contraction give
\begin{align*}
\|\mu_\beta-\mu_\pi\|_{\mathrm{TV}}
&\le\|(\mu_\beta-\mu_\pi)\mathcal P^\beta\|_{\mathrm{TV}}
  +\|\mu_\pi(\mathcal P^\beta-\mathcal P^\pi)\|_{\mathrm{TV}}\\
&\le(1-a_\beta)\|\mu_\beta-\mu_\pi\|_{\mathrm{TV}}+d.
\end{align*}
Rearrange and use that TV is at most one. No mixing rate for $\pi$ is needed for this stationary comparison, provided its stationary law exists.
\end{proof}

\begin{lemma}[Perturbed Markov chain TV bound]\label{thm:delta_bar}
Under the hypotheses of \cref{lemma:two_stationary}, define
$b_\pi(t)=\sup_s\|(\mathcal P^\pi)^t(s,\cdot)-\mu_\pi\|_{\mathrm{TV}}$.
For every initial state $s$ and integer $t\ge0$,
\begin{equation}\label{eq:tv_stationary}
\begin{aligned}
\|(\mathcal P^\beta)^t(s,\cdot)-\mu_\beta\|_{\mathrm{TV}}
&\le\|(\mathcal P^\beta)^t(s,\cdot)-(\mathcal P^\pi)^t(s,\cdot)\|_{\mathrm{TV}}\\
&\quad+b_\pi(t)+\min\{1,d/a_\beta\}.
\end{aligned}
\end{equation}
If, additionally, $\pi$ satisfies its own uniform-ergodicity bound $b_\pi(t)\le C_\pi(1-a_\pi)^t$, this envelope may replace $b_\pi(t)$ on the right.
\end{lemma}

\begin{proof}
Insert $(\mathcal P^\pi)^t(s,\cdot)$ and $\mu_\pi$ between $(\mathcal P^\beta)^t(s,\cdot)$ and $\mu_\beta$, apply the triangle inequality, and use \cref{lemma:two_stationary}.
\end{proof}

The target-chain mixing term is explicit: behavior-chain ergodicity alone does not supply the same rate for $\pi$. For example, if $d<a_\beta/2$, the triangle inequality implies
\[
\delta(\mathcal P^\pi)\le\delta(\mathcal P^\beta)+2d
\le1-(a_\beta-2d),
\]
so one may take $C_\pi=1$ and $a_\pi=a_\beta-2d$. Alternatively, the all-policy cross-Doeblin condition of \cref{assumption:3}, applied separately to $(\beta,\beta)$ and $(\pi,\pi)$, supplies one-step constants $a_\beta=a_\pi=\rho$ when that stronger condition is imposed.

For a coupling of the two different kernels, the first term in \cref{eq:tv_stationary} is at most $\Pr(S_t^\beta\ne S_t^\pi)$. This is a time-$t$ disagreement probability, not generally bounded above by $\Pr(T_{\beta,\pi}>t)$: the chains can separate after their first agreement. Thus this perturbation decomposition does not by itself turn the \texttt{MINT} stopping time into a behavior-chain mixing-time bound.
\endgroup

\begingroup\color{black}
\subsection{Behavior-chain mixing and off-policy truncation: two couplings}
\label{proof:thm_truncation_length}\label{app:two-couplings}

\paragraph{Behavior--behavior coupling for estimating $\tau_{\mathrm{mix}}^\beta$.}
Let $(X_t,Y_t)$ be a co-adapted coupling of two copies of $\mathcal P^\beta$, started at arbitrary states $x,y$. Choose the coupling to be coalescent: once the states agree, use the same transition for both copies thereafter. Its coalescence time is
\[
\tau_c^\beta:=\inf\{t\ge0:X_t=Y_t\}.
\]
Unlike an off-policy coupling, both marginals here have the same kernel. Fix such a coupling for each initial pair and set
$m_\beta=\sup_{x,y}\E_{x,y}[\tau_c^\beta]$.

\begin{proposition}[Behavior-chain mixing bound]\label{prop:beta-coalescence}
Suppose $\mathcal P^\beta$ has stationary law $\mu_\beta$. For the coalescent couplings above and every integer $t\ge0$,
\begin{equation}\label{eq:beta-coalescence-tail}
\sup_x\|(\mathcal P^\beta)^t(x,\cdot)-\mu_\beta\|_{\mathrm{TV}}
\le\sup_{x,y}\Pr_{x,y}(\tau_c^\beta>t).
\end{equation}
If $m_\beta<\infty$, then
\begin{equation}\label{eq:tau-mix-bound}
\tau_{\mathrm{mix}}^\beta\le\lceil4m_\beta\rceil.
\end{equation}
\end{proposition}

\begin{proof}
Coalescence implies $\{X_t\ne Y_t\}=\{\tau_c^\beta>t\}$. The coupling inequality bounds
$\|(\mathcal P^\beta)^t(x,\cdot)-(\mathcal P^\beta)^t(y,\cdot)\|_{\mathrm{TV}}$
by this tail. Integrate the second initial state against $\mu_\beta$ and use stationarity to obtain \cref{eq:beta-coalescence-tail}. For $t\ge1$, Markov's inequality bounds its right-hand side by $m_\beta/t$. Taking $t=\lceil4m_\beta\rceil$ proves \cref{eq:tau-mix-bound}; if $m_\beta=0$, all initial states already agree and the claim is immediate.
\end{proof}

This is the same-kernel coupling route to mixing-time bounds \citep[Section~5.2]{levin2017markov}. It preserves the purpose of using two chains to bound the behavior policy's mixing time. It is an analytic upper bound; estimating the supremum or its tail from trajectories requires additional statistical control, and is not supplied by an observed \texttt{MINT} stopping time.

\paragraph{Behavior--target coupling for the \texttt{MINT} survival horizon.}
Return to the distinct-kernel processes of \cref{def:meeting-time}, with
$T_{\beta,\pi}=\inf\{t\ge1:S_t^\beta=S_t^\pi\}$ and starting pair $x=(s_0,a_0)$.
Here the cumulative mask is killed at first agreement, while the state processes may later separate. This stopping time controls the state-coupled trace and is distinct from $\tau_c^\beta$.

\begin{theorem}[Expected state-coupled truncation horizon]\label{thm:truncation_length}
Suppose the conditional meeting probability at each transition after time $1$, on survival, is at least $\rho>0$; conditional maximal coupling under \cref{assumption:3} ensures this. Write $u_{1,x}=\Pr_x(T_{\beta,\pi}>1)$. Then, for every integer $t\ge1$,
\begin{equation}\label{eq:offpolicy-survival-tail}
\Pr_x(T_{\beta,\pi}>t)\le u_{1,x}(1-\rho)^{t-1},
\qquad \E_x[T_{\beta,\pi}]\le1+\frac{u_{1,x}}\rho.
\end{equation}
If the initial conditional coupling also satisfies $u_{1,x}\le1-\rho$, these simplify to
\[
\Pr_x(T_{\beta,\pi}>t)\le(1-\rho)^t\quad(t\ge0),
\qquad \E_x[T_{\beta,\pi}]\le\rho^{-1}.
\]
For a deterministic integer cap $L\ge1$, define $K_{\mathrm{state}}=\min\{L,T_{\beta,\pi}\}$. Under this additional initial-step condition,
\[
\E_x[K_{\mathrm{state}}]\le\min\{L,\rho^{-1}\},
\qquad M_x(\gamma,\lambda)\le\frac1{1-\gamma\lambda(1-\rho)}.
\]
\end{theorem}

\begin{proof}
For $t\ge1$, $\{T_{\beta,\pi}>t\}$ is measurable with respect to the joint history through time $t$. The conditional meeting bound gives
\[
\Pr_x(T_{\beta,\pi}>t+1)
=\E_x\!\left[\mathbf1\{T_{\beta,\pi}>t\}
\Pr(A_{t+1}=1\mid\mathcal F_t)\right]
\le(1-\rho)\Pr_x(T_{\beta,\pi}>t).
\]
Iterate from time $1$. The tail-sum identity gives
$\E_x[T_{\beta,\pi}]=\sum_{t\ge0}\Pr_x(T_{\beta,\pi}>t)\le1+u_{1,x}/\rho$.
If the initial-step condition holds, the geometric tail starts at time $0$. Since $K_{\mathrm{state}}\le L$ and $K_{\mathrm{state}}\le T_{\beta,\pi}$, its expectation satisfies the stated cap bound. Finally, sum the discounted tail to bound $M_x$.
\end{proof}

The initial-step condition matters when conditioning on a prescribed action $a_0$: overlap of policy-averaged kernels alone does not by itself imply overlap of that conditional first transition. Without it, the same argument still gives
\[
\E_x[K_{\mathrm{state}}]\le\min\{L,1+u_{1,x}/\rho\},\qquad
M_x(\gamma,\lambda)\le1+
\frac{\gamma\lambda u_{1,x}}{1-\gamma\lambda(1-\rho)}.
\]
These bounds concern the state coupling. They do not bound the implemented action-level surrogate's stopping time, whose survival law and contraction are analyzed separately in \cref{app:surrogate-contraction}.

\paragraph{Why the two times cannot be substituted.}
Even when $\beta=\pi$, a common-state synchronous coupling has $T_{\beta,\pi}=1$ while the behavior chain can mix slowly. For example, with
\[
\mathcal P^\beta=\mathcal P^\pi=
\begin{pmatrix}1-\varepsilon&\varepsilon\\\varepsilon&1-\varepsilon\end{pmatrix},
\qquad 0<\varepsilon<\tfrac12,
\]
the stationary law is uniform and
$\|(\mathcal P^\beta)^t(0,\cdot)-\mu_\beta\|_{\mathrm{TV}}=\tfrac12(1-2\varepsilon)^t$.
Thus $\tau_{\mathrm{mix}}^\beta$ grows without bound as $\varepsilon\downarrow0$ although this same-start first agreement remains $1$. The valid mixing bound uses the worst-initial-pair $\beta$--$\beta$ coalescence time in \cref{prop:beta-coalescence}; the \texttt{MINT} horizon uses the $\beta$--$\pi$ survival time in \cref{thm:truncation_length}.
\endgroup

\begingroup\color{black}
\subsection{Online Control with Markov Traces}
\label{proof:online_t4}

This subsection distinguishes Markovianity of a coupled pair from a Markov trace on the original state--action space. The online result below concerns the latter class of expected operators. It is a conditional control result, separate from the state-coupled evaluation theorem in \cref{thm:contraction}.

\paragraph{State coupling on an augmented space.}
For fixed $\beta,\pi$, let $Z_t=(S_t^\beta,S_t^\pi)$ be a Markov coupling of the policy-induced state kernels, with joint transition kernel $K_{\beta,\pi}$ on $\mathcal S\times\mathcal S$. Its marginals are $\mathcal P^\beta(\cdot\mid S_t^\beta)$ and $\mathcal P^\pi(\cdot\mid S_t^\pi)$; maximal coupling is one such construction. For $z'=(s'_\beta,s'_\pi)$, define the killed kernel
\begin{equation}\label{eq:paired-killed-kernel}
\widetilde K_{\beta,\pi}(z,z')
:=K_{\beta,\pi}(z,z')\mathbf1\{s'_\beta\ne s'_\pi\}.
\end{equation}
If $T=\inf\{t\ge1:S_t^\beta=S_t^\pi\}$, then
\begin{equation}\label{eq:paired-survival-resolvent}
\begin{aligned}
\Pr_z(T>t)&=(\widetilde K_{\beta,\pi}^{\,t}\mathbf1)(z),\\
\sum_{t\ge0}(\gamma\lambda)^t\Pr_z(T>t)
&=\bigl[(I-\gamma\lambda\widetilde K_{\beta,\pi})^{-1}\mathbf1\bigr](z).
\end{aligned}
\end{equation}
Killing here terminates the trace at the first meeting; it does not require the two original chains to remain together after meeting. Thus the coupled construction can be Markov on the augmented space even when its survival cannot be described by a single keep function on the behavior state--action space. In particular, under maximal coupling the next disagreement probability depends on both coordinates through $\mathrm{TV}(\mathcal P^\beta(\cdot\mid s_\beta),\mathcal P^\pi(\cdot\mid s_\pi))$. Removing the target-policy state coordinate does not necessarily preserve a Markov trace representation. An online theorem for the augmented construction and a projected update of $Q(s,a)$ would require an additional operator argument; \cref{eq:paired-survival-resolvent} alone does not establish this extension.

\paragraph{Markov-trace setup for the control result.}
Let $\mathcal S,\mathcal A$ be finite, rewards be bounded, $\gamma\in(0,1)$, and $\lambda\in[0,1]$. At control iteration $n$, fix $\beta_n,\pi_n$ and a keep function $g_n:\mathcal S\times\mathcal A\to[0,1]$. These functions may be selected using $Q_n$, but are held fixed when applying that iteration's operator to any input $Q$. Given a behavior trajectory including its rewards, draw independent masks $B_i^{(n)}\sim\mathrm{Bernoulli}(g_n(s_i,a_i))$ with independent randomness. Write $W_t^{(n)}=\prod_{i=1}^t B_i^{(n)}$ and $W_0^{(n)}=1$. The masks do not inspect the following transition or reward. Define
\begin{equation}\label{eq:markov-trace-control-operator}
\mathcal R_nQ(x):=Q(x)+\E_{\beta_n,x}\!\left[
\sum_{t\ge0}(\gamma\lambda)^tW_t^{(n)}\delta_t^{Q,\pi_n}\right],
\end{equation}
where $\delta_t^{Q,\pi_n}=r_t+\gamma\E_{b\sim\pi_n(\cdot\mid s_{t+1})}Q(s_{t+1},b)-Q(s_t,a_t)$. The index $n$ denotes a control iteration, whereas $t$ denotes time within its behavior trajectory. This is an uncapped population operator. The implemented action-level rule is the case $g_n=h_n$ from \cref{eq:surrogate-keep-function}; the coupled-state indicators of \cref{thm:contraction} are not identified with $g_n$ by notation alone.

\begin{theorem}[Exact control convergence with Markov traces]\label{thm:online_t4}
Under the Markov-trace setup above, consider $Q_{n+1}=\mathcal R_nQ_n$. Assume:
\begin{description}
\item[\textup{(A1)}]\label{ass:online-A1} \emph{Evaluation fixed point.} $\mathcal R_nQ^{\pi_n}=Q^{\pi_n}$ for every $n$. This identity follows from the specified operator, as shown below.
\item[\textup{(A2)}]\label{ass:online-A2} \emph{Monotone policy improvement.} $\mathcal T^{\pi_{n+1}}Q_{n+1}\ge\mathcal T^{\pi_n}Q_{n+1}$ componentwise.
\item[\textup{(A3)}]\label{ass:online-A3} \emph{Initialization.} $\mathcal T^{\pi_0}Q_0\ge Q_0$ and $Q_0\le Q^*$ componentwise.
\item[\textup{(A4)}]\label{ass:online-A4} \emph{Coefficient overlap.} $\lambda\beta_n(a\mid s)g_n(s,a)\le\pi_n(a\mid s)$ for every $(s,a)$ and every $n$.
\item[\textup{(A5)}]\label{ass:online-A5} \emph{Asymptotic greediness.} For the Bellman optimality operator $\mathcal T$,
\[
\varepsilon_n:=\|\mathcal TQ_n-\mathcal T^{\pi_n}Q_n\|_\infty\longrightarrow0.
\]
This requirement is separate from (A2).
\end{description}
Then $Q_n\uparrow Q^*$ componentwise and $\|Q_n-Q^*\|_\infty\to0$.
\end{theorem}

\begin{proof}
\textbf{Markov factorization and the evaluation fixed point.}
For a keep function $g$ and behavior policy $\beta$, define
\begin{equation}\label{eq:Ppbeta-def}
C_{\beta,g}((s,a),(s',b))
:=\mathcal P(s'\mid s,a)\beta(b\mid s')g(s',b).
\end{equation}
Write $C_n=C_{\beta_n,g_n}$. Since $C_n\ge0$ and $C_n\mathbf1\le\mathbf1$, the resolvent
\[
\mathcal N_n:=(I-\gamma\lambda C_n)^{-1}
=\sum_{j\ge0}(\gamma\lambda C_n)^j
\]
exists and satisfies $\mathcal N_n\ge I$ entrywise and $\mathbf1\le\mathcal N_n\mathbf1\le(1-\gamma\lambda)^{-1}\mathbf1$. Conditional mask independence and the Markov property give
\begin{equation}\label{eq:markov-trace-factorization}
\E_{\beta_n,x}[W_t^{(n)}\delta_t^{Q,\pi_n}]
=\bigl[C_n^t(\mathcal T^{\pi_n}Q-Q)\bigr](x).
\end{equation}
The bounded geometric sum permits exchanging expectation and summation, hence
\begin{equation}\label{eq:markov-trace-resolvent}
\mathcal R_nQ=Q+\mathcal N_n(\mathcal T^{\pi_n}Q-Q).
\end{equation}
This proves (A1) directly. In fact, invertibility of $\mathcal N_n$ makes $Q^{\pi_n}$ the unique fixed point even before (A4) is used.

\textbf{Matrix order and non-negativity.}
We use only
\[
\mathcal N_n(I-\gamma\lambda C_n)
=(I-\gamma\lambda C_n)\mathcal N_n=I.
\]
Thus $\mathcal N_n$ commutes with $C_n$. No commutation with $\mathcal P^{\pi_n}$, or between operators from different control iterations, is assumed. The linear part of \cref{eq:markov-trace-resolvent} is
\begin{equation}\label{eq:H-identity}
\begin{aligned}
\mathcal R_nQ-\mathcal R_nQ'&=H_n(Q-Q'),\\
H_n&:=I+\mathcal N_n(\gamma\mathcal P^{\pi_n}-I)\\
&=\gamma\mathcal N_n(\mathcal P^{\pi_n}-\lambda C_n).
\end{aligned}
\end{equation}
Indeed, $I=\mathcal N_n(I-\gamma\lambda C_n)$ gives this identity without moving $\mathcal P^{\pi_n}$ across $\mathcal N_n$. Under (A4),
\[
(\mathcal P^{\pi_n}-\lambda C_n)((s,a),(s',b))
=\mathcal P(s'\mid s,a)[\pi_n(b\mid s')-\lambda\beta_n(b\mid s')g_n(s',b)]\ge0.
\]
Consequently $H_n\ge0$, so $\mathcal R_n$ preserves componentwise order. Also,
\begin{equation}\label{eq:H-on-one}
\begin{aligned}
H_n\mathbf1
&=\gamma\mathcal N_n\mathbf1-\gamma\lambda\mathcal N_nC_n\mathbf1\\
&=\mathbf1-(1-\gamma)\mathcal N_n\mathbf1\le\gamma\mathbf1.
\end{aligned}
\end{equation}
Therefore
\begin{equation}\label{eq:H-norm}
\|H_n\|_\infty\le\gamma,
\end{equation}
and the fixed-point identity yields
\begin{equation}\label{eq:online-layer1}
\|Q_{n+1}-Q^{\pi_n}\|_\infty
\le\gamma\|Q_n-Q^{\pi_n}\|_\infty.
\end{equation}

\textbf{Monotonicity of the iterates.}
Let $b_n:=\mathcal T^{\pi_n}Q_n-Q_n$. The residual after the update with the same target policy is
\begin{align*}
\mathcal T^{\pi_n}Q_{n+1}-Q_{n+1}
&=b_n+(\gamma\mathcal P^{\pi_n}-I)\mathcal N_nb_n\\
&=\gamma(\mathcal P^{\pi_n}-\lambda C_n)\mathcal N_nb_n.
\end{align*}
Here $\mathcal N_n-I=\gamma\lambda C_n\mathcal N_n$ is used with $\mathcal N_n$ remaining on the right. In particular, this expression is not obtained by commuting $\mathcal P^{\pi_n}$ and $\mathcal N_n$. Conditions (A2) and (A4) give
\begin{equation}\label{eq:bn-recursion}
b_{n+1}\ge\gamma(\mathcal P^{\pi_n}-\lambda C_n)\mathcal N_nb_n\ge0
\end{equation}
whenever $b_n\ge0$. Since $b_0\ge0$ by (A3), induction proves $b_n\ge0$ for every $n$, and
\begin{equation}\label{eq:Qn-monotone}
Q_{n+1}-Q_n=\mathcal N_nb_n\ge0.
\end{equation}

\textbf{Upper bound and optimality.}
Because $\mathcal T^{\pi_n}Q^*\le\mathcal TQ^*=Q^*$ and $\mathcal N_n\ge0$, \cref{eq:markov-trace-resolvent} implies $\mathcal R_nQ^*\le Q^*$. Order preservation then gives
\[
Q_n\le Q^*\quad\Longrightarrow\quad
Q_{n+1}=\mathcal R_nQ_n\le\mathcal R_nQ^*\le Q^*.
\]
Together with (A3), this bounds the monotone sequence above by $Q^*$. Finiteness of $\mathcal S\times\mathcal A$ gives convergence to some $Q_\infty\le Q^*$ in supremum norm, and hence $\|Q_{n+1}-Q_n\|_\infty\to0$. Since $\mathcal N_n\ge I$ entrywise and $b_n\ge0$,
\[
0\le b_n\le\mathcal N_nb_n=Q_{n+1}-Q_n,
\qquad \|b_n\|_\infty\to0.
\]
Now apply (A5):
\begin{equation}\label{eq:online-optimal-residual}
\|\mathcal TQ_n-Q_n\|_\infty
\le\varepsilon_n+\|b_n\|_\infty\longrightarrow0.
\end{equation}
The $\gamma$-contraction of $\mathcal T$ yields
\[
\|Q_n-Q^*\|_\infty
\le\frac{\|\mathcal TQ_n-Q_n\|_\infty}{1-\gamma}
\longrightarrow0,
\]
so $Q_\infty=Q^*$.
\end{proof}

\paragraph{Roles of the assumptions and the surrogate specialization.}
As in \citet[Theorem~2]{munos2016safe}, (A2) and (A5) perform distinct tasks: monotone policy improvement preserves the residual sign, whereas asymptotic greediness rules out convergence to a nonoptimal policy value. Choosing every $\pi_n$ greedy with respect to $Q_n$ satisfies (A5) with $\varepsilon_n=0$ and also supplies (A2) at the next update. The Markov-trace setup supplies (A1), while (A3) initializes monotonicity. Condition (A4) is a quantitative coefficient-overlap condition, not merely a common-support assumption: it makes both $H_n$ and the residual propagation matrix in \cref{eq:bn-recursion} nonnegative.

For $g_n=h_n=1-\min(1,\pi_n/\beta_n)$ on the support of $\beta_n$, $\beta_nh_n=[\beta_n-\pi_n]_+$. Thus (A4) is equivalent to
\begin{equation}\label{eq:A4-equivalent}
\pi_n(a\mid s)\ge\frac{\lambda}{1+\lambda}\beta_n(a\mid s)
\quad\text{for all $(s,a)$ with $\beta_n(a\mid s)>0$.}
\end{equation}
Indeed, where $\beta_n\le\pi_n$ the coefficient constraint is automatic; otherwise it is $\lambda(\beta_n-\pi_n)\le\pi_n$, which rearranges to \cref{eq:A4-equivalent}. At $\lambda=0$ the condition is automatic. This equivalence concerns the action-level surrogate; it is not a re-expression of a general paired-state survival indicator. Unlike the fixed-policy $\chi_\lambda$ condition of \cref{thm:surrogate-contraction}, (A4) is used here to obtain the order preservation needed for control.

\paragraph{Deep-RL extensions: entropy, KL constraints, and smoothing.}
Entropy-regularized actors, including SAC \citep{haarnoja2018soft}, and KL-based policy constraints motivate considering policy pairs with substantial overlap. To invoke this theorem for the surrogate, however, a quantitative bound such as $\inf_{s,a:\beta_n(a\mid s)>0}\pi_n(a\mid s)/\beta_n(a\mid s)\ge\lambda/(1+\lambda)$ must hold; full support, entropy regularization, or a finite KL bound alone is not used as a proof of (A4). This discussion concerns possible extensions of the operator analysis rather than an automatic guarantee for actor--critic training.

For deterministic-policy methods such as TD3, introducing explicitly specified smoothed policies can make density ratios available. An application to those smoothed policies would still require the coefficient-overlap and policy-improvement hypotheses above. Extending the result to the deterministic limit requires a separate continuity argument and is not established by the present theorem. Likewise, entropy-regularized optimal control requires the corresponding soft Bellman operator and policy-improvement conditions if its optimum is the object of interest.

\paragraph{Scope of the control result.}
The theorem concerns exact, uncapped, finite-dimensional updates of \cref{eq:markov-trace-control-operator}. It does not assert that arbitrary state-coupled traces admit this factorization on $(s,a)$, or that augmenting the pair state alone establishes the same control result after projection. Finite-cap evaluation retains terminal bootstrap terms as in \cref{thm:surrogate-contraction}; a finite-cap control extension requires its own corresponding argument. Sample-based updates, nonlinear critics, and simultaneously learned actors are further extensions. These distinctions preserve the state-coupled evaluation theorem as a separate main result.
\endgroup

\begingroup\color{black}
\subsection{Contraction of the Action-Level Surrogate}\label{app:surrogate-contraction}

This subsection analyzes the implemented action-level stopping rule directly, separately from the state-coupled operator of \cref{thm:contraction}. The behavior policy may be arbitrary; contraction requires the sufficient condition below. No transition-kernel coupling or cross-Doeblin hypothesis is used.

\paragraph{Setup and survival quantities.}
Let $\mathcal S,\mathcal A$ be finite, rewards be bounded, and $\beta,\pi$ be fixed policies. Fix $\gamma\in(0,1)$, $\lambda\in[0,1]$, and an integer hard cap $L=K_{\max}\ge1$. Write $\alpha=\gamma\lambda$ and
\begin{equation}\label{eq:surrogate-keep-function}
h(s,a):=1-\min\!\left(1,\frac{\pi(a\mid s)}{\beta(a\mid s)}\right).
\end{equation}
The ratio is needed only where $\beta(a\mid s)>0$; set $h=0$ otherwise. Thus $\beta(a\mid s)h(s,a)=[\beta(a\mid s)-\pi(a\mid s)]_+$ everywhere. Given a behavior trajectory, including its rewards, draw independent masks $\hat A_i\sim\mathrm{Bernoulli}(h(s_i,a_i))$ using independent randomness. In particular, the mask at $i$ does not inspect the next transition or reward. Define $W_t=\prod_{i=1}^t\hat A_i$, $W_0=1$, and $\hat T=\inf\{i\ge1:\hat A_i=0\}$, with $\inf\varnothing=\infty$. The expected surrogate operator is
\begin{equation}\label{eq:surrogate-population-operator}
\mathcal R^{\mathrm{sur}}_{\lambda,L}Q(x)
:=Q(x)+\E_x\!\left[\sum_{t=0}^{L-1}\alpha^tW_t\delta_t^Q\right],
\qquad x=(s_0,a_0).
\end{equation}
The cap in this definition is exactly the implementation's $\min(L,\hat T)$; it is not an uncapped approximation.

Let $D_s=\mathrm{TV}(\beta(\cdot\mid s),\pi(\cdot\mid s))$. Define the negative-coefficient mass and its normalized fraction by
\begin{align}\label{eq:surrogate-chi-definition}
e_\lambda(s)
&:=\sum_a[\lambda\beta(a\mid s)h(s,a)-\pi(a\mid s)]_+
=\sum_a[\lambda\beta(a\mid s)-(1+\lambda)\pi(a\mid s)]_+,\\
\chi_\lambda(s)&:=
\begin{cases}
e_\lambda(s)/(\lambda D_s),&\lambda D_s>0,\\
0,&\lambda D_s=0.
\end{cases}\nonumber
\end{align}
Since $0\le e_\lambda(s)\le\lambda D_s$, we have $\chi_\lambda(s)\in[0,1]$. This fraction measures the negative-weight correction left by the action-dependent keep rule; it is generally different from the policy TV $D_s$.

For every starting pair $x$, define
\begin{align}\label{eq:surrogate-survival-functionals}
\widehat M_L^x&:=\sum_{t=0}^{L-1}\alpha^t\Pr_x(\hat T>t),\\
\bar\chi_{\lambda,L}^{\,x}
&:=\frac{\sum_{t=1}^{L-1}\alpha^t\E_x[W_t\chi_\lambda(s_t)]}
{\widehat M_L^x-1},\nonumber
\end{align}
where the second quantity is $0$ if its denominator vanishes. All survival probabilities here belong to the surrogate's own stopping process. In particular, neither $\widehat M_L^x$ nor $\bar\chi_{\lambda,L}^{\,x}$ is inferred by substituting $\hat p_i$ into the state-level coupling analysis.

\begin{theorem}[Finite-cap surrogate contraction]\label{thm:surrogate-contraction}
Under the setup above, $\mathcal R^{\mathrm{sur}}_{\lambda,L}$ has the unique fixed point $Q^\pi$. Moreover, for all $Q,Q'$, its supremum-norm Lipschitz modulus is bounded by
\begin{equation}\label{eq:surrogate-chi-bound}
\kappa_{\chi,L}:=
\sup_x\left\{1-(1-\gamma)\widehat M_L^x
+2\bar\chi_{\lambda,L}^{\,x}(\widehat M_L^x-1)\right\}.
\end{equation}
Consequently, if $\kappa_{\chi,L}<1$, the exact iteration $Q_{n+1}=\mathcal R^{\mathrm{sur}}_{\lambda,L}Q_n$ satisfies
\[
\|Q_n-Q^\pi\|_\infty\le\kappa_{\chi,L}^{\,n}\|Q_0-Q^\pi\|_\infty.
\]
The condition $\kappa_{\chi,L}<1$ is sufficient, not necessary.
\end{theorem}

\begin{proof}
\textbf{Expected operator and fixed point.}
On state--action pairs define the nonnegative matrix
\begin{equation}\label{eq:surrogate-C}
C((s,a),(s',b)):=\mathcal P(s'\mid s,a)\beta(b\mid s')h(s',b),
\qquad
S_L:=\sum_{j=0}^{L-1}(\alpha C)^j,\quad S_0:=0.
\end{equation}
Because $C\mathbf1\le\mathbf1$ and $\alpha<1$, $\|\alpha C\|_\infty<1$. Conditional mask independence gives $\E[W_t\mid\text{trajectory}]=\prod_{i=1}^th(s_i,a_i)$. The Markov property and the conditional mean of the TD residual therefore yield
\[
\E_x[W_t\delta_t^Q]
=\bigl(C^t(\mathcal T^\pi Q-Q)\bigr)(x),
\qquad
\mathcal R^{\mathrm{sur}}_{\lambda,L}Q=Q+S_L(\mathcal T^\pi Q-Q).
\]
The factorization $S_L=(I-(\alpha C)^L)(I-\alpha C)^{-1}$ shows that $S_L$ is invertible. Hence $\mathcal R^{\mathrm{sur}}_{\lambda,L}Q=Q$ if and only if $\mathcal T^\pi Q=Q$, proving uniqueness of the fixed point $Q^\pi$.

\textbf{Finite-cap boundary.}
The linear part of the affine operator is
\begin{align}\label{eq:surrogate-H}
H_L&:=I+S_L(\gamma\mathcal P^\pi-I)\\
&=\gamma S_{L-1}(\mathcal P^\pi-\lambda C)
+\gamma(\alpha C)^{L-1}\mathcal P^\pi.\nonumber
\end{align}
This follows by expanding the finite sums; no commutativity between $C$ and $\mathcal P^\pi$ is used. The last term is the terminal bootstrap contribution, and $S_{L-1}$, rather than $S_L$, multiplies the signed interior contribution. Also,
\begin{equation}\label{eq:surrogate-row-mass}
S_L\mathbf1=(\widehat M_L^x)_x,
\qquad H_L\mathbf1=\mathbf1-(1-\gamma)S_L\mathbf1.
\end{equation}

\textbf{Negative mass.}
Let $E_\lambda:=(\lambda C-\mathcal P^\pi)_+$, with positive part and absolute values taken entrywise. Then
\[
|\mathcal P^\pi-\lambda C|=\mathcal P^\pi-\lambda C+2E_\lambda.
\]
All other factors in \cref{eq:surrogate-H} are nonnegative, so
\begin{equation}\label{eq:surrogate-matrix-bound}
|H_L|\mathbf1\le
\mathbf1-(1-\gamma)S_L\mathbf1
+2\gamma S_{L-1}E_\lambda\mathbf1.
\end{equation}
The state-level mass of $E_\lambda$ is exactly $e_\lambda$ from \cref{eq:surrogate-chi-definition}. If $\widetilde\chi(s,a):=\chi_\lambda(s)$, then
\[
(E_\lambda\mathbf1)(s,a)
=\sum_{s'}\mathcal P(s'\mid s,a)e_\lambda(s')
=\lambda(C\widetilde\chi)(s,a),
\]
where $\sum_b\beta(b\mid s')h(s',b)=D_{s'}$ was used. Therefore
\begin{align*}
\gamma(S_{L-1}E_\lambda\mathbf1)(x)
&=\sum_{t=1}^{L-1}\alpha^t(C^t\widetilde\chi)(x)\\
&=\sum_{t=1}^{L-1}\alpha^t\E_x[W_t\chi_\lambda(s_t)]\\
&=\bar\chi_{\lambda,L}^{\,x}(\widehat M_L^x-1).
\end{align*}
Substituting into \cref{eq:surrogate-matrix-bound} and taking the maximum row absolute sum proves $\|H_L\|_\infty\le\kappa_{\chi,L}$. Since $\mathcal R^{\mathrm{sur}}_{\lambda,L}Q-\mathcal R^{\mathrm{sur}}_{\lambda,L}Q'=H_L(Q-Q')$, contraction and the stated geometric convergence follow when $\kappa_{\chi,L}<1$.
\end{proof}

\begin{corollary}[A sufficient overlap condition]\label{cor:surrogate-overlap}
If $\pi(a\mid s)\ge\lambda\beta(a\mid s)/(1+\lambda)$ for all $(s,a)$, then $\mathcal R^{\mathrm{sur}}_{\lambda,L}$ is a $\gamma$-contraction for every $L\ge1$.
\end{corollary}
\begin{proof}
The condition is equivalent to $\lambda\beta h\le\pi$, so $e_\lambda=0$ and $E_\lambda=0$. \Cref{eq:surrogate-H} then gives $H_L\ge0$. Since $S_L\mathbf1\ge\mathbf1$, \cref{eq:surrogate-row-mass} gives $\|H_L\|_\infty\le\gamma$. This is the pointwise overlap condition appearing in \cref{eq:A4-equivalent}, used here solely for fixed-policy evaluation.
\end{proof}

\paragraph{Boundary cases and scope.}
At $L=1$ or $\lambda=0$ the surrogate operator reduces to $\mathcal T^\pi$. For an infinite cap, the same proof uses $S=(I-\alpha C)^{-1}$ in place of both $S_L$ and $S_{L-1}$, and the survival sums extend to infinity. A bound at or above $1$ is inconclusive: exact finite-dimensional iteration converges from every initial value if and only if $\rho(H_L)<1$, so the sufficient norm condition $\|H_L\|_\infty<1$ is not necessary. These are population results for fixed policies. Finite-sample concentration, sample-based stochastic approximation, changing policies, and nonlinear critic projection require additional analysis. In particular, averaging the complete action-dependent backup in a minibatch does not replace $h(s,a)$ by its state mean $D_s$ inside that backup.

\paragraph{Why a contraction condition is needed.}\label{app:surrogate-counterexample}
Consider two states $0,1$ and two deterministic actions: \emph{stay} preserves the state and \emph{flip} switches it. Let $\beta$ choose each action with probability $1/2$, let $\pi$ always flip, and set rewards to zero. Then $h(s,\mathrm{stay})=1$ and $h(s,\mathrm{flip})=0$. For $v(0)=1,v(1)=-1$, define $Q_0(s,a)=v(f(s,a))$, where $f$ is the deterministic next-state map. We have $CQ_0=Q_0/2$ and $\mathcal P^\pi Q_0=-Q_0$, hence
\[
H_LQ_0=\mu_LQ_0,
\qquad
\mu_L=1-(1+\gamma)\sum_{t=0}^{L-1}(\gamma\lambda/2)^t.
\]
At $\gamma=0.99$, $\lambda=0.7$, $L=5$, this eigenvalue is $\mu_L\approx-2.02993$, so exact expected iteration can diverge even with a finite cap. This example does not satisfy the stated sufficient contraction condition; it explains why that condition cannot be omitted. It makes no claim about small-step stochastic updates or the empirical SAC/TD3 training runs.
\endgroup

\begingroup\color{black}
\subsection{A Common Conservative TV Condition}\label{app:common-TV-condition}

\begin{corollary}[A common sufficient condition]\label{cor:common-TV-condition}
Let $D=\sup_s\mathrm{TV}(\beta(\cdot\mid s),\pi(\cdot\mid s))$. Under the respective fixed-policy hypotheses of \cref{thm:contraction,thm:surrogate-contraction}, the condition
\begin{equation}\label{eq:common-TV-condition}
\gamma(1+2\lambda D)<1
\end{equation}
ensures contraction of each expected evaluation operator toward $Q^\pi$; for the surrogate this holds for every finite cap $L\ge1$.
\end{corollary}

\begin{proof}
Set $\varepsilon_0=1-\gamma-2\gamma\lambda D>0$. For the state-coupled operator,
$N_x\le D(M_x-1)$ and $M_x-1\le\gamma\lambda M_x$ give
\[
1-(1-\gamma)M_x+2N_x\le1-\varepsilon_0M_x\le1-\varepsilon_0<1
\]
uniformly over $x$. For the surrogate, $e_\lambda(s)\le\lambda D_s\le\lambda D$, so $E_\lambda\mathbf1\le\lambda D\mathbf1$. Since $S_{L-1}\mathbf1\le S_L\mathbf1$, \cref{eq:surrogate-matrix-bound} gives
\[
|H_L|\mathbf1\le\mathbf1-\varepsilon_0S_L\mathbf1
\le(1-\varepsilon_0)\mathbf1.
\]
The separate fixed-point results complete both claims.
\end{proof}

For $\lambda>0$, \cref{eq:common-TV-condition} requires $D<(1-\gamma)/(2\gamma\lambda)$. At the main MuJoCo setting $(\gamma,\lambda)=(0.99,0.7)$, this is $D<0.00722$ uniformly over states. It is a restrictive shared baseline, not a certificate for the experimental regime. The survival-dependent state condition and the surrogate's negative-mass condition can be less restrictive; their relative restrictiveness depends on their stopping distributions and actionwise policy ratios. The common condition neither identifies the two stopping processes nor transfers one theorem to the other operator.
\endgroup

\section{Extended Theoretical Discussion}\label{app:regimes}

\paragraph{Reading the survival and overlap diagrams.}
\Cref{fig:phase-diagram-main} shows the general mismatch--horizon budget at
$\gamma=0.9$, without a cross-Doeblin assumption.
\Cref{fig:phase-diagram}, placed with the geometric corollary in
\cref{app:cor-closed-form}, instead shows the overlap specialization at
$\gamma=0.99$. The two diagrams describe sufficient conditions for the
state-coupled operator, not measured deep-RL operating points.

\subsection{Four Regimes of \texorpdfstring{\cref{thm:contraction}}{Theorem 1}}\label{app:four-regimes}

\revr{For the scalar examples below, suppose the survival law is the same for all starting pairs and write $M=M_x$. The uniform modulus reduces to $\kappa=1-(1-\gamma)M+2\bar D_{\mathrm{wgt}}(M-1)$, and the uniform TV-based sufficient condition \cref{eq:contraction-TV-condition} reduces to $2\bar D_{\mathrm{wgt}}(M-1)/M<1-\gamma$. For nonuniform survival laws, the supremum over starting pairs must be retained.} The main text discusses Regime 3 (geometric meeting); the other three regimes are recorded here for completeness.

\paragraph{Regime 1 — On-policy limit ($\bar D_{\mathrm{wgt}} \to 0$).}
\revr{For every starting pair, $N_x\le\bar D_{\mathrm{wgt}}(M_x-1)\le\bar D_{\mathrm{wgt}}\gamma\lambda/(1-\gamma\lambda)$, so the correction vanishes uniformly as $\bar D_{\mathrm{wgt}}\to0$ for fixed $\gamma,\lambda$. At $\bar D_{\mathrm{wgt}}=0$, $\kappa_x=1-(1-\gamma)M_x\le\gamma$ for every $x$. In the on-policy case, $M_x=1$ and $N_x=0$ for all $x$, recovering $\kappa=\gamma$.}

\paragraph{Regime 2 — Deterministic meeting at $T = 1$.}
$\Pr(T>0)=1$, $\Pr(T>t)=0$ for $t \ge 1$, so $M=1$ and $\kappa \le \gamma$. \emph{One-step Bellman is recovered for any $\bar D_{\mathrm{wgt}}$}, reflecting that with immediate meeting there is no off-policy correction to apply.

\paragraph{Regime 3 — Geometric meeting time, $T \sim \mathrm{Geom}(q)$.}
$\Pr(T>t) = (1-q)^t$, so $M = 1/\bigl(1-\gamma\lambda(1-q)\bigr)$ and $(M-1)/M = \gamma\lambda(1-q)$. Substituting into \cref{eq:contraction-TV-condition} gives the closed-form sufficient condition
\[
\boxed{\;\gamma\bigl(1 + 2\bar D_{\mathrm{wgt}}\lambda(1-q)\bigr) \;<\; 1.\;}
\]
The sufficient-condition region widens as the per-step meeting probability
$q$ grows: at $q=1$ we recover Regime 2; at $q=0$ we recover Regime 4 below.
Here the survival law is exactly geometric. In \cref{cor:geometric}, the
cross-Doeblin parameter $\rho$ instead supplies a geometric upper bound
on the survival tail; equality is not required.

\paragraph{Regime 4 — Worst case ($T = \infty$).}
The two coupled trajectories never meet. $\Pr(T>t) = 1$ for all $t$, so $M = 1/(1-\gamma\lambda)$ and $(M-1)/M = \gamma\lambda$. The condition becomes
\[
\boxed{\;\gamma\bigl(1 + 2\bar D_{\mathrm{wgt}}\lambda\bigr) \;<\; 1.\;}
\]
At $\bar D_{\mathrm{wgt}} = 1, \lambda = 1$ this gives $\gamma < 1/3$, the most pessimistic case. But at $\bar D_{\mathrm{wgt}} = 0.1, \lambda = 1$ it admits $\gamma < 1/1.2 \approx 0.833$; at $\bar D_{\mathrm{wgt}} = 0.01$ it admits $\gamma < 0.98$. Even when the trajectories never meet, reducing $\bar D_{\mathrm{wgt}}$ allows the sufficient condition to hold at larger discount factors.

\subsection{Practical Regime Corollaries}\label{app:practical-regime}

\Cref{thm:contraction} holds for any $\gamma \in (0,1)$, but the worst-case quantitative bound becomes vacuous when $\gamma$ is close to $1$ unless $\bar D_{\mathrm{wgt}}$ is correspondingly small. In this subsection we specialize to the moderate-discount regime $\gamma \le 0.9$ and give two explicit sufficient conditions: a worst-case ($\ell_\infty$) form and an averaged ($L^1$) form. \textcolor{black}{Both follow from the pointwise bound of \cref{thm:contraction}. The averaged result is an $L^\infty\!\to L^1$ one-step error bound; the averaging step requires neither cross-Doeblin nor ergodicity and is not an iteration-convergence argument.}

\paragraph{Untruncated TD-error reference.}
\revr{Setting $A_i\equiv1$ in \cref{eq:t4-operator} gives
$Q+(I-\gamma\lambda\mathcal P^\beta)^{-1}(\mathcal T^\pi Q-Q)$,
the $Q^\pi(\lambda)$ evaluation operator of \citet{harutyunyan2016q}.
It retains $Q^\pi$ as its fixed point. This is distinct from Peng's
mixture-of-returns operator in \cref{tab:operator-comparison}.
The no-survival reference $M_x=(1-\gamma\lambda)^{-1}$ gives the
sufficient bound $\gamma(1+2\bar D_{\mathrm{wgt}}\lambda)<1$ from the
current analysis. Comparisons with this expression concern sufficient
bounds; they do not identify a convergence region for Peng's algorithm.}

\begin{corollary}[Practical worst-case contraction at moderate discount]\label{cor:practical-worst}
\revr{Adopt the operator setup of \cref{thm:contraction}. Let $K_\lambda=\sup_x\min(\E_x[T_{\beta,\pi}],(1-\gamma\lambda)^{-1})$ be the uniform effective-horizon bound in \cref{eq:M-vs-K}. Fix $\gamma\in(0,1)$.}
\begin{itemize}
\item \textbf{(A) Free-trace regime.} If $\bar D_{\mathrm{wgt}} \le \tfrac{1-\gamma}{2}$, then $\kappa < 1$ for \emph{every} $K_\lambda \ge 1$ (equivalently, every meeting-time distribution and every $\lambda \in [0,1]$).
\item \textbf{(B) Bounded-trace regime.} If $\bar D_{\mathrm{wgt}} > \tfrac{1-\gamma}{2}$, then $\kappa < 1$ whenever
\begin{equation}\label{eq:Klam-bound}
K_\lambda \;<\; \frac{2\bar D_{\mathrm{wgt}}}{2\bar D_{\mathrm{wgt}} - (1-\gamma)} \;=:\; K_\lambda^{\max}(\gamma,\bar D_{\mathrm{wgt}}).
\end{equation}
In particular, choosing $\lambda < \lambda^*(\gamma,\bar D_{\mathrm{wgt}}) := \tfrac{1-\gamma}{2\gamma\bar D_{\mathrm{wgt}}}$ forces $\tfrac{1}{1-\gamma\lambda} < K_\lambda^{\max}$ and therefore guarantees contraction for every meeting-time distribution.
\end{itemize}
\end{corollary}

\begin{proof}
\revr{For every $x$, \cref{eq:kappa-bound} gives $\kappa_x\le1-(1-\gamma)M_x+2\bar D_{\mathrm{wgt}}(M_x-1)$. In Regime (A), the coefficient $2\bar D_{\mathrm{wgt}}-(1-\gamma)$ of $M_x$ is non-positive, so $M_x\ge1$ yields $\kappa_x\le\gamma$ uniformly. In Regime (B), this coefficient is positive, so $M_x\le K_\lambda$ yields a uniform upper bound strictly below $1$ when $K_\lambda<K_\lambda^{\max}$. The $\lambda^*$ statement follows from $(1-\gamma\lambda^*)^{-1}=K_\lambda^{\max}$.}
\end{proof}

\begin{remark}[Specialization to $\gamma = 0.9$]\label{rem:gamma09}
At $\gamma=0.9$ the free-trace threshold is $\bar D_{\mathrm{wgt}}\le0.05$.
In the bounded-trace regime, $\bar D_{\mathrm{wgt}}=0.10$ requires
$K_\lambda<2$, while $\bar D_{\mathrm{wgt}}=0.50$ requires
$K_\lambda<1.111\ldots$.
These are fixed-policy state-coupled guarantees. In the free-trace regime,
no additional meeting-time restriction is needed; this does not certify
the action-level surrogate or deep-RL training.
\end{remark}

\begin{table}[!htbp]
\centering
\caption{State-coupled contraction budget at $\gamma=0.9$
(\cref{cor:practical-worst}).
$\lambda^*=(1-\gamma)/(2\gamma\bar D_{\mathrm{wgt}})$ is the critical
$\lambda$ below which the sufficient condition holds for any meeting-time
distribution. Finite entries require $K_\lambda<K_\lambda^{\max}$;
$\infty$ denotes no additional horizon restriction beyond the discount-only
bound, not an infinite realized horizon.}
\label{tab:contraction-gamma09}
\smallskip
\begin{tabular}{ccccc}
\toprule
$\bar D_{\mathrm{wgt}}$ & Regime & $\lambda^*$ & $K_\lambda^{\max}$ ($\lambda{=}1$) & $K_\lambda^{\max}$ ($\lambda{=}0.5$) \\
\midrule
$0.001$ & (A) & any & $\infty$ & $\infty$ \\
$0.05$  & (A) & any & $\infty$ & $\infty$ \\
$0.10$  & (B) & $0.556$ & $2.00$ & $\infty$ \\
$0.30$  & (B) & $0.185$ & $1.20$ & $1.20$ \\
$0.50$  & (B) & $0.111$ & $1.11$ & $1.11$ \\
$1.00$  & (B) & $0.0556$ & $1.05$ & $1.05$ \\
\bottomrule
\end{tabular}
\end{table}

\begingroup\color{black}
\begin{corollary}[Averaged one-step error bound]\label{cor:practical-expected}
Adopt the state-coupled operator setup and the pointwise Lipschitz bound of \cref{thm:contraction}, without requiring its uniform contraction condition. Fix a probability distribution $\nu$ over starting pairs $x=(s,a)$, and write
\[
\bar M:=\E_{x\sim\nu}[M_x],\qquad
\|f\|_{L^1(\nu)}:=\E_{x\sim\nu}[|f(x)|].
\]
Then for every bounded $Q$,
\begin{equation}\label{eq:expected-contraction}
\|\mathcal R_{p,\lambda}Q-Q^\pi\|_{L^1(\nu)}
\le\kappa^\star\|Q-Q^\pi\|_\infty,
\end{equation}
where
\begin{equation}\label{eq:kappa-star}
\kappa^\star:=1-(1-\gamma)\bar M
                 +2\bar D_{\mathrm{wgt}}(\bar M-1).
\end{equation}
The coefficient is below one in either of the following regimes:
\begin{itemize}
\item If $\bar D_{\mathrm{wgt}}\le(1-\gamma)/2$, then $\kappa^\star\le\gamma<1$ for every admissible $\bar M\ge1$.
\item If $\bar D_{\mathrm{wgt}}>(1-\gamma)/2$, then $\kappa^\star<1$ whenever $\bar M<2\bar D_{\mathrm{wgt}}/[2\bar D_{\mathrm{wgt}}-(1-\gamma)]$.
\end{itemize}
No cross-Doeblin or mixing assumption is needed for this averaging step. If the behavior chain has a stationary law $\mu_\beta$, one may choose $\nu=\mu_\beta\otimes\beta$ as a special case.
\end{corollary}

\begin{proof}
The fixed-point identity and the pointwise bound of \cref{thm:contraction} give
\[
|\mathcal R_{p,\lambda}Q(x)-Q^\pi(x)|
\le\left[1-(1-\gamma)M_x+2\bar D_{\mathrm{wgt}}(M_x-1)\right]
\|Q-Q^\pi\|_\infty.
\]
Integrating with respect to $\nu$ gives \cref{eq:expected-contraction}. The two regimes follow by considering the sign of $2\bar D_{\mathrm{wgt}}-(1-\gamma)$ and using $\bar M\ge1$.
\end{proof}

This is an $L^\infty\!\to L^1(\nu)$ one-step error bound. Even if $\kappa^\star<1$, it is not an $L^1(\nu)$ contraction, since the right-hand side uses the supremum norm. Consequently it does not by itself prove convergence of repeated updates; the uniform sufficient condition of \cref{thm:contraction} supplies that separate guarantee.
\endgroup

\paragraph{\rev{Relation to the empirical deep-RL settings.}}
\revr{The free-trace threshold $(1-\gamma)/2$ is $0.005$ at $\gamma=0.99$. The RWRL runs report mean action-level keep probabilities between $0.70$ and $0.99$ depending on the arm, but these statistics do not identify the state-coupled survival quantities or certify the $\chi_\lambda$ condition of \cref{thm:surrogate-contraction}. The measured $\hat\kappa=0.84\pm0.13$ at $K_{\max}=5$, below $1$ on $90\%$ of probes (\cref{app:kappa-measured}), is empirical evidence about visited-distribution backup gains, not an estimate of $\kappa^\star$ or a certificate of $\kappa_{\chi,L}<1$.}

\section{Extended Toy-Chain Validation}\label{app:toy-extended}

\subsection{Diagnostic study on tabular CliffWalking}\label{app:cliffwalking}

\Cref{assumption:2,assumption:3} specify a kernel-level TV bound $d$ and uniform cross-policy overlap $\rho$ for the coupling analysis. Our diagnostic study on tabular \texttt{CliffWalking} (\cref{fig:tv-diagnostic}\,(a)) illustrates their relation to transition structure: policy deviations induce near-deterministic branching and absorbing transitions, with state-wise TV distances close to $1$ and cross-state overlap near zero. As shown in \cref{fig:tv-diagnostic}\,(b), both the state-wise discrepancies $d(s)$ and the overlap matrix $\rho(s,s')$ between an optimal policy and a uniformly random policy exhibit extreme mismatch across most states---reflecting structural properties of the MDP rather than policy proximity. \textcolor{black}{\Cref{fig:tv-diagnostic}\,(c) evaluates the RHS of \cref{thm:nearby} at several illustrative $(d,\rho)$ pairs. These curves bound time-averaged expected disagreement under their stated hypotheses; they do not verify a positive uniform $\rho$ for CliffWalking or imply a meeting-time tail. The latter requires the separate conditional overlap assumptions in \cref{thm:truncation_length}.}

\begin{figure}[!htbp]
\centering
\begin{subfigure}[t]{0.58\linewidth}
    \centering
    \includegraphics[width=\linewidth]{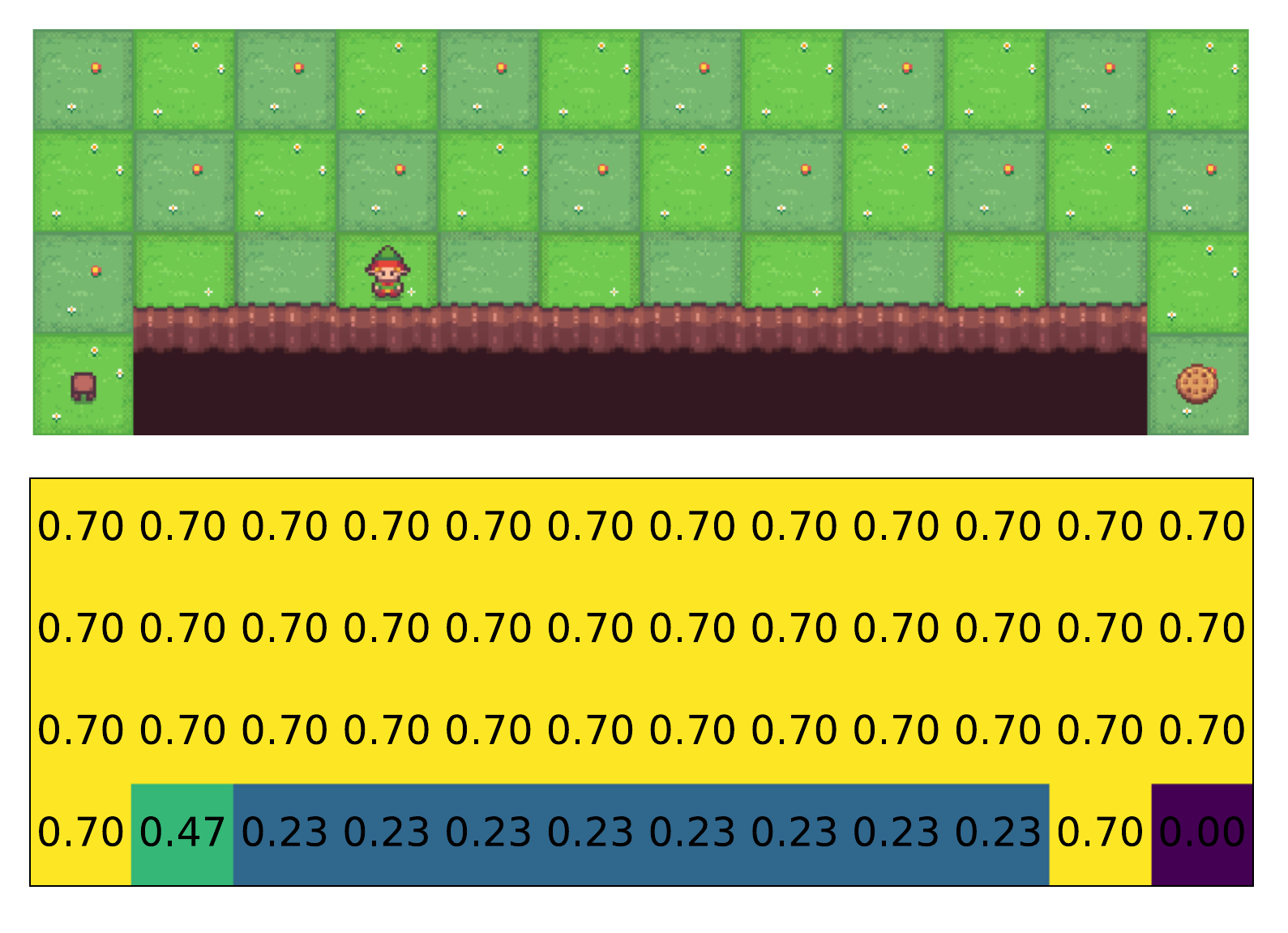}
    \caption{}
    \label{fig:tv-cliff}
\end{subfigure}\\[6pt]
\begin{subfigure}[t]{0.44\linewidth}
    \centering
    \includegraphics[width=\linewidth]{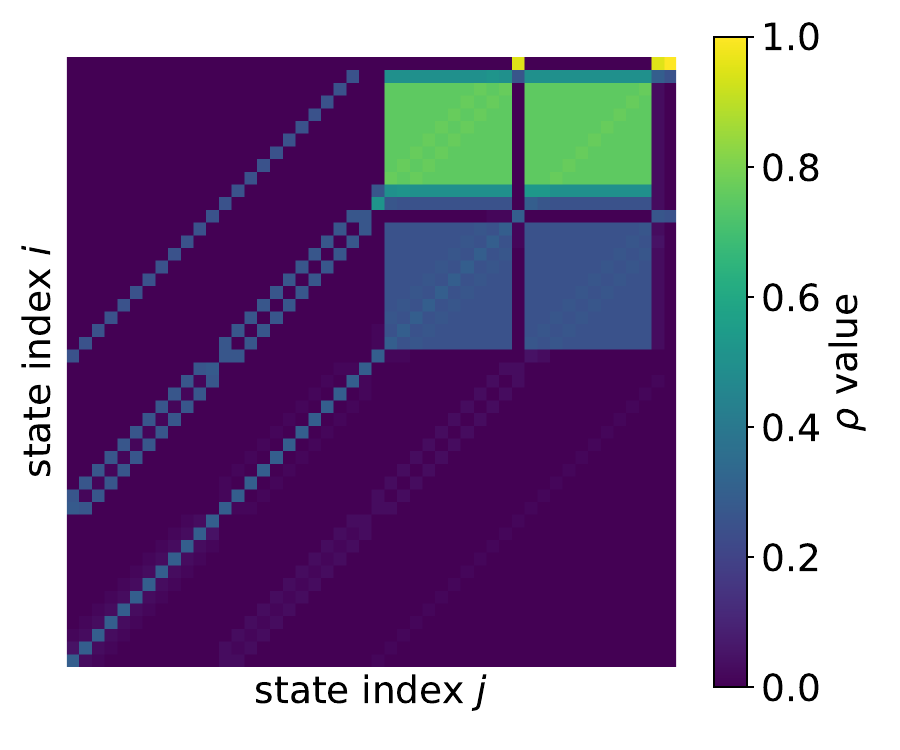}
    \caption{}
    \label{fig:tv-rho}
\end{subfigure}\hfill
\begin{subfigure}[t]{0.50\linewidth}
    \centering
    \includegraphics[width=\linewidth]{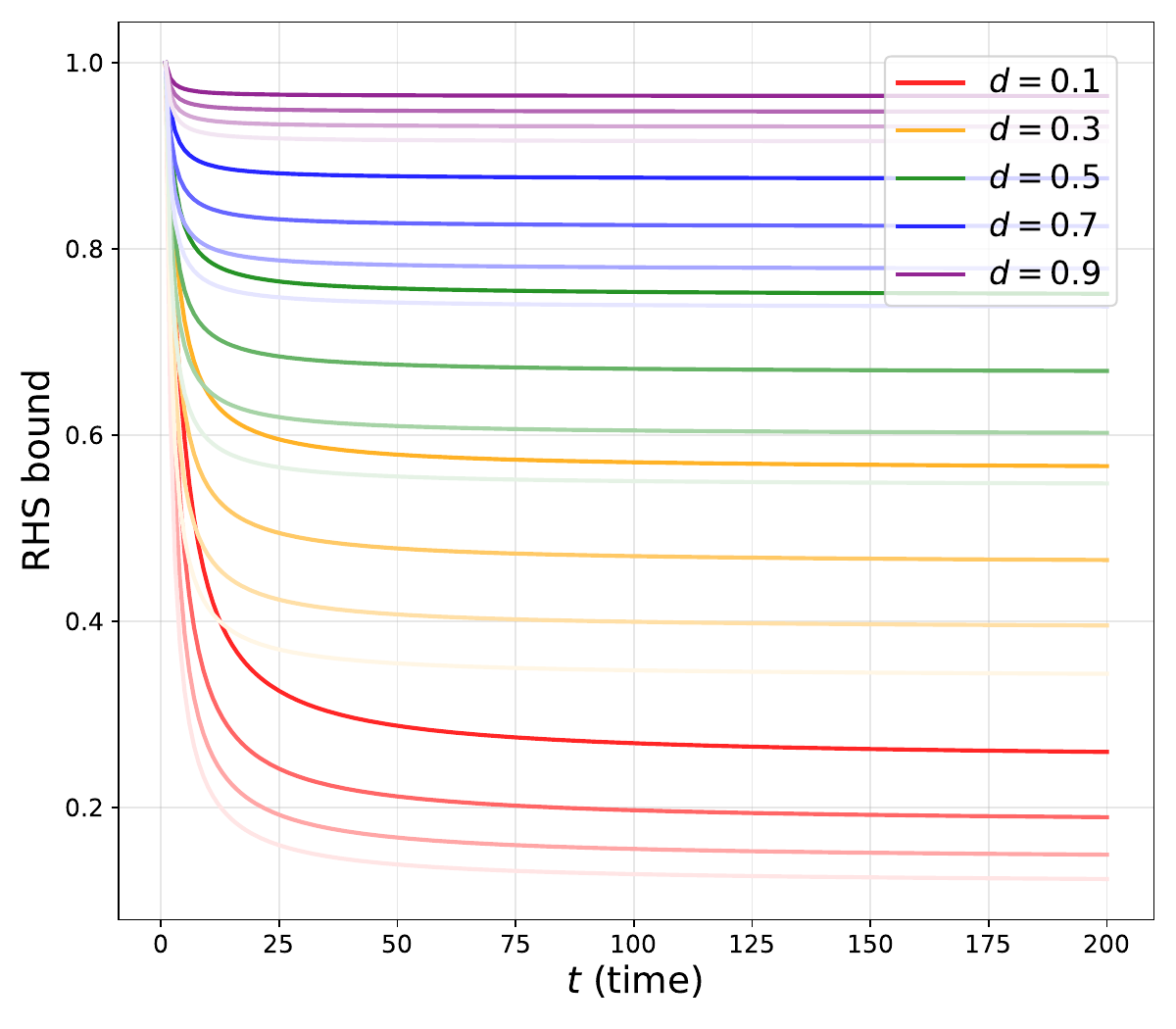}
    \caption{}
    \label{fig:tv-lemma2}
\end{subfigure}
\caption{\textbf{Diagnostic study on tabular CliffWalking.}
\textbf{(a-top)} Environment visualization for \texttt{CliffWalking}.
\textbf{(a-bottom)} State-wise total variation distance $d(s)$ between an optimal policy and a uniformly random policy, illustrating pronounced off-policy discrepancies near the cliff region.
\textbf{(b)} Pairwise overlap matrix $\rho(s,s')$ between transition kernels induced by the two policies; near-zero overlap across most state pairs highlights a strong structural mismatch.
\textcolor{black}{\textbf{(c)} Time-averaged expected-disagreement bounds from \cref{thm:nearby} for illustrative $(d,\rho)$ pairs; these curves do not certify a meeting-time bound for the displayed environment.}}
\label{fig:tv-diagnostic}
\end{figure}

\subsection{Single-Trap chain refinements}

\paragraph{Main-text second-moment experiment.}\label{app:single-trap-moment}
For \cref{fig:variance-empirical}, the chain uses $N=20$, states $\{0,\ldots,N\}$, reward $r=+1$ at the trap state $N$, slippery transitions with $\xi=0.1$, and reflecting boundaries. The behavior policy is uniform over $\{-1,+1\}$; the target is identical except $\pi(+1\mid N)=1$. We evaluate the multi-step Bellman-residual estimator at trace lengths $\{1,5,10,20,50\}$ with $\gamma=0.9$ and $\lambda=1$. Both axes are logarithmic. Error bars are $\pm1$ standard deviation over 30 replicates and are smaller than the plot markers. At $t=50$, the second moments of Retrace and Peng's $Q(\lambda)$ are $1.43\times$ and $2.28\times$ that of \texttt{MINT}; these comparisons concern only the tested trace lengths and this localized-mismatch setup.

We extend the Single-Trap chain validation of the main text with two additional refinements: data-processing gap (refinement (ii)) and cross-Doeblin restart (refinement (iii)). Setup follows the main text: Single-Trap chain with states $\{0,\ldots,N\}$, reflective boundaries, two actions $\{-1,+1\}$, $\beta$ uniform, $\pi$ identical to $\beta$ except $\pi(+1\mid N) = 1$ at the trap state. The per-state action TV is $D_s = 0$ everywhere except $D_N = 0.5$, so the worst-case bound is $\bar D = 0.5$ regardless of dynamics.

\subsection{Refinement (ii): Data-processing gap on slippery transitions}

\revr{With slippery transitions (intended action with probability $1-\xi$, a uniform random action otherwise), the action-level TV $\bar D=0.5$ is unchanged, while the one-step kernel TV at the trap shrinks to $0.25$ at $\xi=0.5$. The plotted kernel-weighted diagnostic drops from $0.11$ to $0.06$, while the policy-TV diagnostic remains around $0.12$ (\cref{fig:slippery-chain}). \textcolor{black}{This illustrates a one-step data-processing gap. Kernel TV cannot replace policy TV in the correction term of \cref{thm:contraction}: that proof bounds signed coefficients on state--action values before any push-forward through the dynamics. The weighted kernel quantity is a diagnostic, not a contraction certificate. The implemented surrogate has a separate $\chi_\lambda$ correction (\cref{thm:surrogate-contraction}).}}

\begin{figure}[!htbp]
\centering
\includegraphics[width=0.95\linewidth]{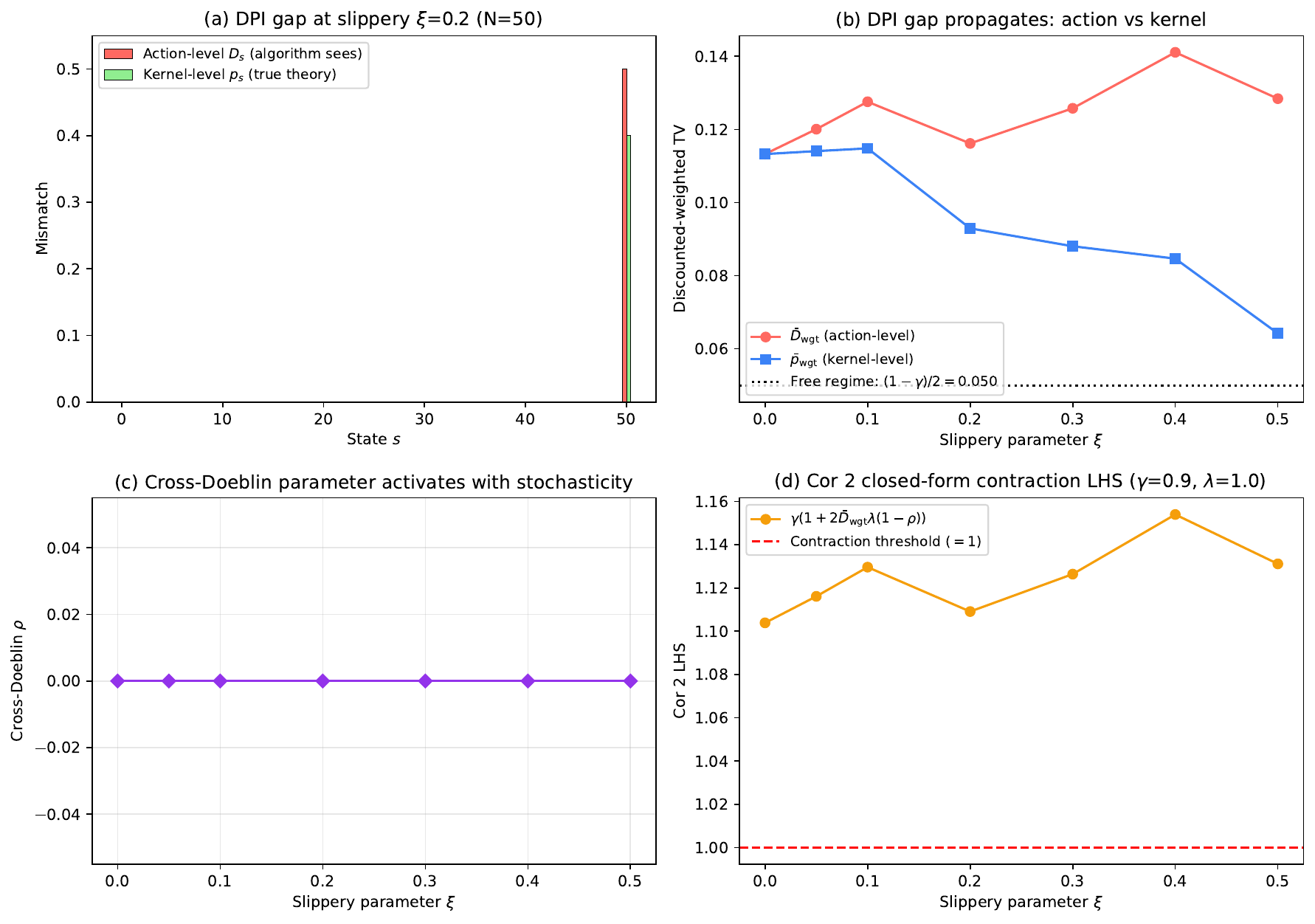}
\caption{Data-processing gap on the slippery Single-Trap chain ($N{=}50$, $\gamma{=}0.9$, $\lambda{=}1$, slippery $\xi$ varied). \textbf{(a)} Action-level $D_s$ (red) is unchanged by slippery dynamics, while kernel-level $p_s$ (green) shrinks at the trap state. \textcolor{black}{\textbf{(b)} The plotted policy-TV and kernel-TV weighted diagnostics separate with $\xi$; the smaller kernel diagnostic is not substituted into the contraction condition.} \textbf{(c)} Cross-Doeblin $\rho = 0$ for the chain MDP without restart. \textbf{(d)} The left-hand side of the sufficient condition in \cref{cor:geometric} exceeds 1 throughout; the restart modification introduces $\rho>0$.}
\label{fig:slippery-chain}
\end{figure}

\subsection{Refinement (iii): Cross-Doeblin restart and the no-survival reference}

We add a uniform-restart component (with probability $q_{\mathrm{rst}}$, transition is uniform over all states; with probability $1-q_{\mathrm{rst}}$, slippery dynamics) to obtain $\rho \approx q_{\mathrm{rst}}$ (\cref{fig:restart-chain}\,(a)). We compare \texttt{MINT}'s contraction LHS $\gamma(1+2\bar D_{\mathrm{wgt}}\lambda(1-\rho))$ against the no-survival reference LHS $\gamma(1+2\bar D_{\mathrm{wgt}}\lambda)$ (the $\rho=0$ specialization of this sufficient bound). \Cref{tab:region-ratio} reports the measured region ratio $1/(1-\rho)$ alongside both LHS values. An illustrative parameter setting is $q_{\mathrm{rst}} = 0.20$: \texttt{MINT}'s LHS drops to $0.985 < 1$ (contraction), while the reference LHS is $1.006 > 1$ (no contraction guarantee). The empirical region ratio matches the theoretical $1/(1-\rho)$ to three decimal places across $\rho \in \{0.1, 0.3, 0.5\}$, providing a quantitative comparison of these sufficient bounds.

\begin{table}[!htbp]
\centering
\caption{Uniform-restart diagnostic: \texttt{MINT} and the no-survival reference bound on the restart chain ($N{=}30$, $\gamma{=}0.9$, $\lambda{=}1$, $\xi{=}0.1$). The contraction condition is $\gamma(1+2\bar D_{\mathrm{wgt}}\lambda(1-\rho)) < 1$ for \texttt{MINT} and the $\rho{=}0$ specialization for the reference. At $q_{\mathrm{rst}} = 0.20$: \texttt{MINT} contracts ($\checkmark$) while the reference bound is inconclusive.}
\label{tab:region-ratio}
\smallskip
\begin{tabular}{ccccc}
\toprule
$q_{\mathrm{rst}}$ & measured $\rho$ & \texttt{MINT} LHS & Reference LHS & $1/(1-\rho)$ \\
\midrule
$0.00$ & $0.000$ & $1.101$ & $1.101$ & $1.000$ \\
$0.05$ & $0.050$ & $1.064$ & $1.072$ & $1.053$ \\
$0.10$ & $0.100$ & $1.023$ & $1.036$ & $1.111$ \\
\textbf{0.20} & \textbf{0.200} & $\mathbf{0.985\ \checkmark}$ & $1.006$ & $1.250$ \\
$0.30$ & $0.300$ & $0.957\ \checkmark$ & $0.982\ \checkmark$ & $1.429$ \\
$0.50$ & $0.500$ & $0.930\ \checkmark$ & $0.960\ \checkmark$ & $2.000$ \\
\bottomrule
\end{tabular}
\end{table}

\begin{figure}[!htbp]
\centering
\includegraphics[width=0.95\linewidth]{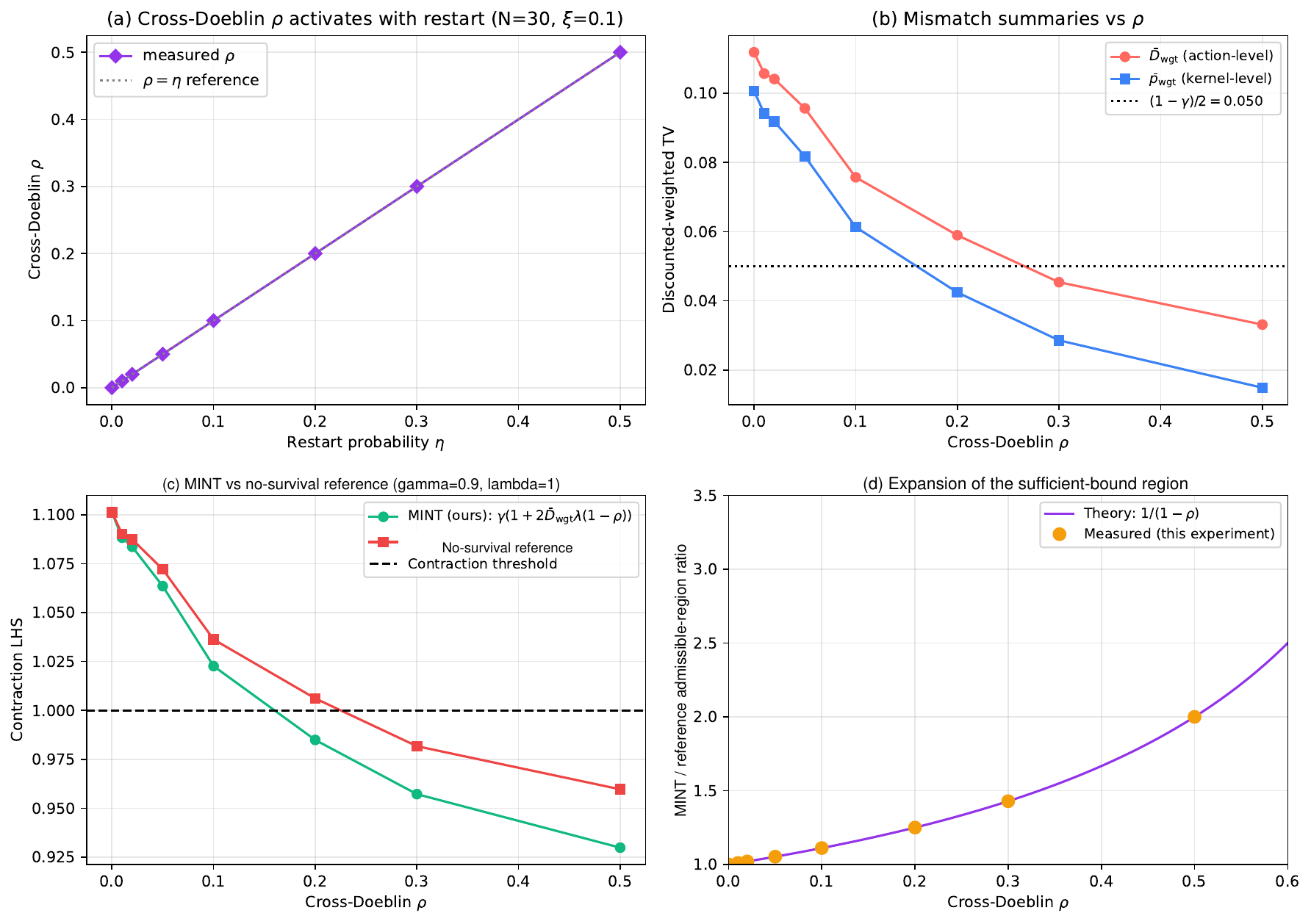}
\caption{Uniform-restart diagnostic: cross-Doeblin overlap and the region satisfying \texttt{MINT}'s sufficient contraction condition on the restart chain ($N{=}30$, $\gamma{=}0.9$, $\xi{=}0.1$). \textbf{(a)} Measured $\rho$ matches restart probability $q_{\mathrm{rst}}$ exactly. \textcolor{black}{\textbf{(b)} The plotted policy-TV and kernel-TV diagnostics decrease with $\rho$. Only the policy-TV quantity enters the stated contraction bound.} \textbf{(c)} \texttt{MINT}'s contraction LHS (green) drops below $1$ at $q_{\mathrm{rst}} = 0.20$, while the no-survival reference LHS (red) does not until $q_{\mathrm{rst}} = 0.30$. \textbf{(d)} Empirical region ratio matches the theoretical $1/(1-\rho)$ to three decimal places.}
\label{fig:restart-chain}
\end{figure}

\paragraph{Summary of refinements.}
\begingroup\color{black}
The diagnostics have separate roles. Survival weighting gives the plotted policy-TV value $\bar D_{\mathrm{wgt}}\approx0.1$, compared with the worst-case action TV $\bar D=0.5$. Slippery dynamics give a smaller kernel-TV diagnostic, $\bar p_{\mathrm{wgt}}\approx0.06$. For any common nonnegative normalized weights $\omega_s$, DPI implies $\sum_s\omega_sd(s)\le\sum_s\omega_sD_s$, but this does not justify replacing $\bar D_{\mathrm{wgt}}$ by $\bar p_{\mathrm{wgt}}$ in \cref{thm:contraction}. Cross-Doeblin overlap separately changes the geometric-tail sufficient bound by the factor $1-\rho$; the displayed threshold ratio $1/(1-\rho)$ compares those two state-coupled sufficient bounds with the same policy-TV parameter. These observations do not compose into a multiplicative contraction improvement and do not transfer a state-coupled guarantee to the implemented surrogate.
\endgroup

\begingroup\color{black}
\subsection{High-Mismatch Ablations on the Single-Trap Chain}\label{app:ablations}

We compare stopping rules in a controlled high-mismatch stress test, using the single-trap dynamics but different policy pairs from the localized-mismatch refinements in \cref{fig:slippery-chain,fig:restart-chain}. The policies below differ at every state. Results from these settings are not pooled with, or compared in absolute scale to, the localized-mismatch experiments.

\paragraph{Setup and measured quantities.}
We use $N=20$, $\gamma=0.9$, $\lambda=1$, slippery transitions with probability $0.1$, and a $50$-step evaluation window. The representative setting is
\[
\beta(+1\mid s)=0.05,\qquad \pi(+1\mid s)=0.95,
\qquad D_s=0.90\quad\text{for every }s.
\]
An additional comparison uses $\beta(+1\mid s)=0.005$ and $\pi(+1\mid s)=0.995$, giving $D_s=0.99$ at every state. Within each setting, all variants use the same pool of behavior trajectories: $20$ replicates of $20{,}000$ trajectories. The anchor is always retained, $W_0=1$, and $W_t=\prod_{i=1}^tB_i$ for continuation indicators $B_i$. We report the mean retained horizon and the sampling variance of
\[
\widehat\Delta_t=\sum_{u=0}^{t-1}(\gamma\lambda)^uW_u\delta_u^{Q^\pi}
\]
at $t\in\{1,5,10,20,50\}$. The variance plots concern evaluation at $Q^\pi$; the separate perturbed-critic experiment below measures sampled backup MSE.

\paragraph{Stopping rules and controls.}
\begin{itemize}
\item \textbf{Action-level surrogate:} draw $B_i\sim\mathrm{Bernoulli}(h(s_i,a_i))$, where $h=1-\min(1,\pi/\beta)$ is the keep probability in \cref{eq:practical-disagreement}. Set $\hat T=\inf\{i\ge1:B_i=0\}$. With no additional cap we retain $\min(50,\hat T)$ terms; the capped variant retains $\min(10,\hat T)$ terms.
\item \textbf{Constant-$p$ controls:} draw independent $B_i\sim\mathrm{Bernoulli}(p)$, with $p$ denoting keep probability. The sweep uses $p\in\{0.1,0.5,0.9\}$. Separate rate-matched controls choose $p$ so that $\sum_{u=0}^{49}p^u$ matches either the surrogate or the state-coupled reference horizon. The two matching targets are distinguished in the legends.
\item \textbf{State-coupled reference (``oracle'' in the legends):} use the first-agreement survival mask from the coupled state processes. This is a reference requiring state-coupling information, not a minimum-MSE estimator or an upper bound on practical performance.
\item \textbf{Symmetric log-ratio heuristic (``symmetric TV'' in the legends):} use
\begin{equation}\label{eq:ablation-symmetric-keep}
h_{\mathrm{sym}}(s,a)
=\operatorname{clip}\!\left(\tfrac12\left|\log\pi(a\mid s)-\log\beta(a\mid s)\right|,0,1\right)
\end{equation}
as the keep probability. This heuristic is not an exact TV estimator, and the surrogate contraction theorem is not asserted for this alternative rule.
\end{itemize}

\paragraph{A: Comparison with horizon-matched random truncation.}
\Cref{fig:ablation-panels}\,(a,d) includes controls matched separately to the surrogate and to the state-coupled reference. In these state-independent policy-mismatch settings, the surrogate does not show a consistent advantage over its horizon-matched constant-$p$ control across variance and backup MSE. Thus these tests do not support a general claim that the surrogate improves on random truncation at a matched horizon. They also do not establish equivalence of the estimators: the surrogate mask depends on the sampled action, whereas the constant-$p$ mask is independent of that action. A setting with constant $D_s$ provides limited evidence about benefits from adapting to statewise variation in policy mismatch.

\paragraph{B: Saturation of the symmetric heuristic.}
For the opposing policies here, $\tfrac12|\log(\pi/\beta)|=\tfrac12\log19\approx1.472$ at $D_s=0.90$, and $\tfrac12\log199\approx2.647$ at $D_s=0.99$, for either action. Hence \cref{eq:ablation-symmetric-keep} is identically one: the symmetric heuristic retains the full $50$-step window. In the representative setting its variance is approximately $0.160$, versus $0.129$ for the surrogate, and its backup MSE is higher (\cref{fig:ablation-panels}\,(b,d)). The variance curves are closer at $D_s=0.99$ (\cref{fig:ablation-panels-high-mismatch}), indicating that their separation depends on the policy pair. This comparison diagnoses saturation of this particular heuristic, rather than proving that one clipping form is uniformly superior.

\paragraph{C: The cap reduces horizon and sampled backup MSE.}
For a state-independent pair $\beta(+1)=b$, $\pi(+1)=q>b$, the surrogate has
\[
h(+1)=0,\qquad h(-1)=\frac{q-b}{1-b},\qquad
\E_\beta[h\mid s]=q-b=D.
\]
Conditional on the pre-action history, the mean keep probability is therefore always $D$. Iterating the survival recursion gives
\begin{equation}\label{eq:ablation-mismatch-horizon}
\Pr(\hat T>u)=D^u,\qquad
\E[\min(L,\hat T)]=\sum_{u=0}^{L-1}D^u=\frac{1-D^L}{1-D}.
\end{equation}
At $D=0.90$, the analytic horizons are $9.948$ for $L=50$ and $6.513$ for $L=10$, consistent with the measured $10.0$ and $6.52$. At $D=0.99$, they are $39.499$ and $9.562$, consistent with $39.5$ and $9.57$. The probabilities of reaching the cap are $\Pr(\hat T\ge10)=D^9$, namely $38.7\%$ and $91.4\%$, consistent with the insets. These marginal survival identities do not make the surrogate independent of the trajectory.

For backup MSE, set $Q_0=Q^\pi+\xi$ with $\xi\sim\mathcal N(0,I)$ over state--action pairs. Each of $20$ perturbation seeds fixes one $\xi$ shared across the trajectory replicates and variants. We measure
\begin{equation}\label{eq:ablation-backup-MSE}
\mathrm{MSE}_{\mathrm{backup}}
=\E\!\left[\bigl(\widehat{\mathcal R}Q_0(s_0,a_0)-Q^\pi(s_0,a_0)\bigr)^2\right],
\end{equation}
where $\widehat{\mathcal R}Q_0$ is the sampled finite-window TD-error backup. The table reports mean $\pm$ standard deviation across perturbation seeds. Figure panels (a)--(c) show $\pm1$ standard deviation of the variance estimates across $20$ trajectory replicates; panel (d) shows bootstrap percentile $95\%$ confidence intervals for the mean across the $20$ perturbation seeds. The perturbation scale in panel (d) is $\sigma=1$, distinct from the slippery-transition probability.

\begin{table}[!htbp]
\color{black}\captionsetup{font={small,color=black}}
\centering\small
\caption{High-mismatch surrogate and cap comparison at a $50$-step evaluation window. Backup MSE is mean $\pm$ SD over $20$ perturbation seeds; variance is measured at $Q^\pi$. No additional cap means that the trace is still limited by the $50$-step window.}
\label{tab:ablation-summary}
\begin{tabular}{clrrr}
\toprule
$D_s$ & Variant & $K_{\mathrm{eff}}$ & Variance & Backup MSE \\
\midrule
$0.90$ & Surrogate, no additional cap & $10.0$ & $0.129$ & $13.8\pm12.1$ \\
       & Surrogate, cap $10$ & $6.52$ & $0.128$ & $9.5\pm6.5$ \\
\midrule
$0.99$ & Surrogate, no additional cap & $39.5$ & $0.143$ & $40.8\pm46.5$ \\
       & Surrogate, cap $10$ & $9.57$ & $0.141$ & $16.1\pm12.6$ \\
\bottomrule
\end{tabular}
\end{table}

In the representative $D_s=0.90$ case, cap $10$ lowers mean backup MSE from $13.8$ to $9.5$ ($31.2\%$), while variance changes only from $0.129$ to $0.128$ (about $0.8\%$). In the $D_s=0.99$ comparison, it lowers mean backup MSE from $40.8$ to $16.1$ ($60.5\%$), while variance changes from $0.143$ to $0.141$ (about $1.4\%$). The mean MSE decrease is accompanied by lower MSE in $16$ of the $20$ reported perturbation seeds in the latter setting, with a median paired capped/uncapped ratio of $0.49$. These observations support the cap's role in controlling the sampled backup under the tested mismatch and perturbation distribution, not a claim of large variance reduction or an optimal choice of cap.

The state-coupled reference remains long: its measured horizons are $40.4$ and $44.5$, with backup MSE $29.2\pm32.4$ and $41.7\pm48.3$ at $D_s=0.90$ and $0.99$, respectively. Thus the cap does not match the surrogate horizon to the reference horizon. The smaller MSE of a much shorter capped backup does not establish an ordering between the state-coupled and surrogate contraction conditions. More aggressive constant-$p$ truncation can yield still smaller MSE in this perturbation experiment, as shown in panel (d); the experiment does not measure the credit-assignment benefit of long backups in learned-critic training.

\paragraph{Scope of the MSE comparison.}
For a fixed $Q_0$ and starting pair, the sampled MSE decomposes as
\[
\E[(\widehat{\mathcal R}Q_0-Q^\pi)^2]
=(\mathcal RQ_0-Q^\pi)^2+\operatorname{Var}(\widehat{\mathcal R}Q_0),
\qquad \mathcal RQ_0=\E[\widehat{\mathcal R}Q_0].
\]
It therefore combines expected-backup error with sampling variance. A reduction in this quantity does not prove sup-norm contraction of the expected operator or convergence of repeated updates. These are stress tests of fixed-policy estimators; they do not supply a contraction certificate for the deep-RL runs. These perturbation experiments use $\sigma=1$; no claim of robustness across perturbation scales is made from these figures.

\begin{figure}[!htbp]
\color{black}\captionsetup{font={small,color=black}}
\centering
\includegraphics[width=0.90\linewidth]{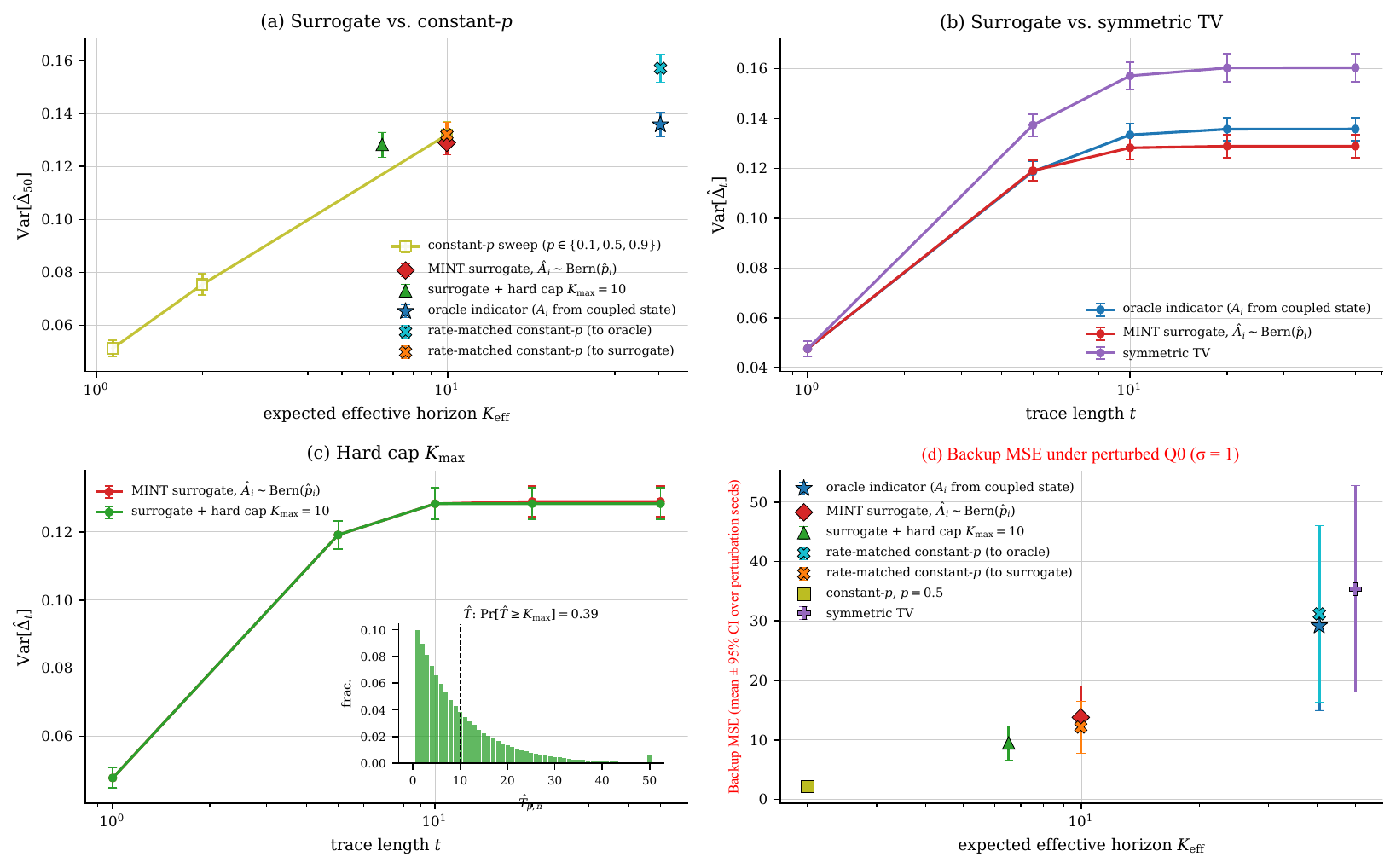}
\caption{\textbf{Representative high-mismatch ablation: $D_s=0.90$.} Policies are $\beta(+1\mid s)=0.05$ and $\pi(+1\mid s)=0.95$ at every state. \textbf{(a)} Variance versus mean retained horizon, including random controls matched separately to the surrogate and state-coupled reference. \textbf{(b)} The symmetric log-ratio heuristic saturates and retains all $50$ steps; the surrogate has lower variance here. \textbf{(c)} Cap $10$ reduces the measured horizon from $10.0$ to $6.52$ with little change in variance. The inset displays the right-censored stopping time $\min(\hat T,50)$; the final bin includes $\hat T\ge50$, and about $39\%$ of traces reach cap $10$. \textbf{(d)} Sampled backup MSE at $Q_0=Q^\pi+\xi$, $\xi\sim\mathcal N(0,I)$; the cap lowers the mean from $13.8$ to $9.5$. Error bars: (a)--(c), $\pm1$ SD across $20$ trajectory replicates; (d), bootstrap percentile $95\%$ CI for the mean across $20$ perturbation seeds. These MSEs are not expected-operator contraction measurements.}
\label{fig:ablation-panels}
\end{figure}

\paragraph{Additional comparison at $D_s=0.99$.}
\Cref{fig:ablation-panels-high-mismatch} uses the more extreme opposing policy pair. It produces a much longer surrogate horizon and a larger gap between capped and uncapped backup MSE, while retaining the same qualitative conclusions about the matched-random comparison and cap-induced variance change. The comparisons are within this policy setting; they do not identify the absolute variance or MSE of different policy settings with one another.

\begin{figure}[!htbp]
\color{black}\captionsetup{font={small,color=black}}
\centering
\includegraphics[width=0.90\linewidth]{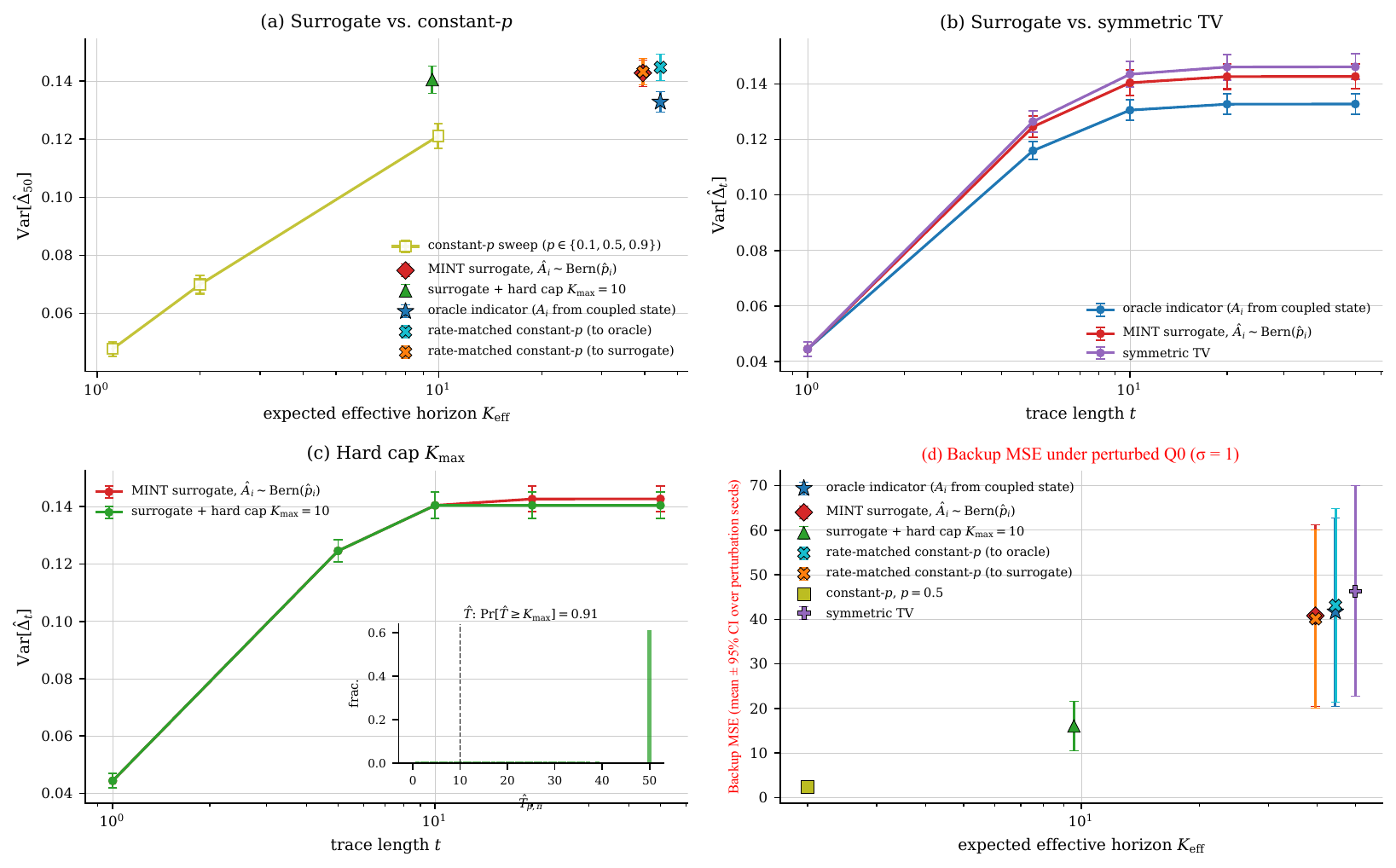}
\caption{\textbf{Additional high-mismatch comparison: $D_s=0.99$.} Policies are $\beta(+1\mid s)=0.005$ and $\pi(+1\mid s)=0.995$ at every state; panel definitions and error-bar conventions match \cref{fig:ablation-panels}. Cap $10$ reduces the measured surrogate horizon from $39.5$ to $9.57$ and mean backup MSE from $40.8$ to $16.1$, while variance changes only from $0.143$ to $0.141$. About $91\%$ of traces reach cap $10$. The inset is right-censored at $50$, so the final bin includes $\hat T\ge50$. The state-coupled reference has horizon $44.5$ and is not a minimum-MSE benchmark. The matched-random controls and variance comparisons do not establish a general surrogate advantage across policy pairs.}
\label{fig:ablation-panels-high-mismatch}
\end{figure}

\paragraph{Reproducibility.}
The experiment and plotting entry points are \path{scripts/single_trap_ablation.py} and \path{paper_figures/ablation_single_trap.py}. The $D_s=0.99$ run is identified by \path{outputs/single_trap_ablation/mismatch_b0.005_q0.995.json}; the policy pair stated above identifies the representative $D_s=0.90$ setting. Trajectory-length curves and stopping-time histograms use the same high-mismatch policy pairs and protocols as the corresponding backup comparisons.
\clearpage
\endgroup

\section{Implementation Details}\label{sec:implementation_detail}

\begin{algorithm}[H]
\caption{\texttt{MINT} multi-step target construction --- \textcolor{black}{ordinary-residual form matching \cref{alg:t4}; SAC/TD3 substitutions are specified below.}}
\label{alg:t4-full}
\begin{algorithmic}[1]
\setcounter{ALG@line}{100}
\Require target policy $\pi$, current critic $Q_\theta$, target critic $Q_{\bar\theta}$, sampled trajectory $(s_0, a_0, r_0, s_1, a_1, r_1, \ldots, s_H)$ from replay $\mathcal{B}$ \rev{together with the stored behavior log-probabilities $\log\beta(a_i\mid s_i)$ recorded at collection time (so $\beta$ is the data-generating policy of that trajectory, not the current actor)}, discount $\gamma$, decay $\lambda \in [0,1]$, hard cap $K_{\max}$ \revr{(a protocol constant, fixed to $5$ in the main continuous-control comparisons)}.
\Ensure multi-step target $\hat y$ for $(s_0, a_0)$.
\State \textbf{[Disagreement estimate]} For $i = 1, \ldots, K_{\max}-1$, compute
\[
\hat p_i \;\gets\; 1 - \min\!\bigl(1,\ \pi(a_i\mid s_i)/\beta(a_i\mid s_i)\bigr)
\quad\text{(action-level surrogate, \cref{eq:practical-disagreement}).}
\]
\State \textbf{[Sample stochastic mask]} Draw $\hat A_i \sim \mathrm{Bernoulli}(\hat p_i)$ independently for $i = 1, \ldots, K_{\max}-1$ using independent randomness.
\State \textbf{[Effective horizon]} Let $\hat T\gets\inf\{1\le i<K_{\max}:\hat A_i=0\}$, with $\inf\varnothing=\infty$; set $\hat K\gets\min(K_{\max},\hat T)$.
\State \textbf{[Trace-mask product]} For $t = 0, 1, \ldots, \hat K - 1$, define $w_t \gets (\gamma\lambda)^t \prod_{i=1}^{t} \hat A_i$, so $w_t = (\gamma\lambda)^t$ until step $\hat T-1$ and $w_t = 0$ afterwards.
\State \textbf{[Bellman residuals]} For $t = 0, \ldots, \hat K - 1$, compute the Bellman residual $\delta_t^{Q}$ (cf. \cref{eq:retrace-general})
\[
\delta_t \;\gets\; r_t + \gamma\,\E_{a' \sim \pi(\cdot\mid s_{t+1})} Q_{\bar\theta}(s_{t+1}, a') - Q_{\bar\theta}(s_t, a_t).
\]
\State \textbf{[Multi-step target]}
\[
\hat y \;\gets\; Q_{\bar\theta}(s_0, a_0) + \sum_{t=0}^{\hat K - 1} w_t\, \delta_t.
\]
\State \textbf{[Critic update]} Take a gradient step on $(Q_\theta(s_0, a_0) - \hat y)^2$; refresh $Q_{\bar\theta}$ and update $\pi$ as in the base actor--critic algorithm.
\end{algorithmic}
\end{algorithm}

In this section, we describe the full implementation details of \texttt{MINT}.
Following the standard practice in off-policy RL,
we use the PyTorch version of the implementations in OpenAI SpinningUp \citep{SpinningUp2018}.

\paragraph{Experimental Setup}

We compare \texttt{MINT} with four baseline methods, a conventional one-step method, uncorrected multi-step method, Retrace \citep{munos2016safe} and Peng's $Q(\lambda)$ \citep{kozuno2021revisiting}.
Given a randomly sampled trajectory $(s_0,a_0,r_0,s_1,a_1,r_1,s_2,\cdots)$, where $Q_{\theta^-}$ denotes the target $Q$-function, and $\tilde{a}_\phi(s_i)$ is a sample from $\pi_\phi(\cdot|s_i)$. The detailed targets for the $Q$-function of all algorithms are described in \cref{tab:update_pseudocode}.
\textcolor{black}{Multi-step targets train the critic at the sampled anchor $(s_0,a_0)$. The actor retains its backbone objective at replay states using fresh actor actions; for SAC this maximizes the critic value minus the log-policy term, as specified in \cref{eq:sac-actor-objective}.}

\paragraph{Training and evaluation.}
For all algorithms,
we use $[256,256]$-sized multi-layer perceptrons (MLPs) for all neural networks.
We train with 1M environment steps for OpenAI MuJoCo and evaluate the agent every 1000 steps using a deterministic policy in 10 episodes.

\begingroup\color{black}
\paragraph{Finite-cap target definitions and backward recursion.}
All formulas below use a fixed target critic $Q_{\bar\theta}$ during target construction. Write $q_i=Q_{\bar\theta}(s_i,a_i)$ and $v_i=\E_{a\sim\pi(\cdot\mid s_i)}Q_{\bar\theta}(s_i,a)$. With twin target critics, the fixed function used here can be $Q_{\bar\theta}=\min_{j=1,2}Q_{\bar\theta_j}$. The stored actions $a_i$ and fresh target-policy actions in $v_i$ have different roles. In particular, $v_i$ is an expectation at $s_i$, not a maximization over the current state.

Let $L\ge1$ be the available deterministic cap and let $c_i$ exclude $\lambda$, as in \cref{eq:retrace-general}. Set
\[
b_0=1,\qquad b_t=\prod_{j=1}^t c_j,\qquad
\delta_i=r_i+\gamma v_{i+1}-q_i.
\]
The finite-cap TD-error target is
\begin{equation}\label{eq:finite-cap-td-target}
Y_L(c)=q_0+\sum_{i=0}^{L-1}(\gamma\lambda)^i b_i\delta_i.
\end{equation}
For \texttt{MINT}, $c_i=\hat A_i$, so $b_i=\mathbf1\{\hat T>i\}$ and terms after $\hat K-1$, with $\hat K=\min(L,\hat T)$, vanish. Thus \cref{eq:finite-cap-td-target} is exactly the target in the main-text \cref{alg:t4} and the full \cref{alg:t4-full}; it includes the anchor $q_0$ and uses $\lambda$ exactly once per continuation step.

An equivalent backward computation is
\begin{equation}\label{eq:finite-cap-backward}
\begin{aligned}
G_{L-1}&=r_{L-1}+\gamma v_L,\\
G_i&=r_i+\gamma v_{i+1}
       +\gamma\lambda c_{i+1}(G_{i+1}-q_{i+1}),
       \qquad i=L-2,\ldots,0,\\
Y_L(c)&=G_0.
\end{aligned}
\end{equation}
To verify the identity, subtract $q_i$ from the recursion and substitute backward: each continuation contributes a factor $\gamma\lambda c_{i+1}$, giving \cref{eq:finite-cap-td-target}. For \texttt{MINT} one may start instead at $G_{\hat K-1}=r_{\hat K-1}+\gamma v_{\hat K}$ and recurse only over the retained prefix. The boundary uses a bootstrap at $s_{\hat K}$ even when the trace ends because of the cap. There is no subtraction of an unsampled $q_L$ at the boundary.

For comparison with reward-return formulas, define $\tilde c_i=\lambda c_i$ for $1\le i<L$ and impose $\tilde c_L=0$. Expanding \cref{eq:finite-cap-td-target} gives
\begin{equation}\label{eq:finite-cap-expanded}
Y_L(c)=\sum_{i=0}^{L-1}\gamma^i
\left(\prod_{j=1}^i\tilde c_j\right)
\left[r_i+\gamma\bigl(v_{i+1}-\tilde c_{i+1}q_{i+1}\bigr)\right].
\end{equation}
The term $\tilde c_Lq_L$ is defined to be zero and can be omitted from evaluation. This terminal convention prevents an extra subtraction at a deterministic cap; it does not remove the final value bootstrap. At $L=1$, every row below reduces to $r_0+\gamma v_1$.

\begin{table}[!htbp]
\centering\small
\caption{Consistent finite-cap targets for the ordinary Bellman residual used in the main text. Here $L$ is deterministic; \texttt{MINT}'s realized prefix length is $\hat K\le L$. The maximum-entropy substitutions are specified below.}
\label{tab:update_pseudocode}
\begin{tabular}{@{}p{0.23\linewidth}p{0.71\linewidth}@{}}
\toprule
Method & Target or continuation rule \\
\midrule
One-step & $r_0+\gamma v_1$. \\[3pt]
Uncorrected $L$-step & $\sum_{i=0}^{L-1}\gamma^i r_i+\gamma^L v_L$. \\[3pt]
Retrace & $Y_L(c)$ with $c_i=\min\{1,\pi(a_i\mid s_i)/\beta(a_i\mid s_i)\}$; use \cref{eq:finite-cap-backward}. \\[3pt]
Peng's $Q(\lambda)$ & $P_{L-1}=r_{L-1}+\gamma v_L$; $P_i=r_i+\gamma[(1-\lambda)v_{i+1}+\lambda P_{i+1}]$ for $i=L-2,\ldots,0$; return $P_0$. \\[3pt]
MINT & $Y_L(c)$ with $c_i=\hat A_i$, $\hat A_i\sim\mathrm{Bernoulli}(h(s_i,a_i))$; equivalently $q_0+\sum_{i=0}^{\hat K-1}(\gamma\lambda)^i\delta_i$. \\
\bottomrule
\end{tabular}
\end{table}

Peng's row is a finite mixture of ordinary multi-step returns \citep{kozuno2021revisiting}, not the $c_i\equiv1$ specialization of \cref{eq:finite-cap-td-target} with target $\pi$. Its expanded form is
\begin{equation}\label{eq:peng-finite-cap}
P_0=\sum_{i=0}^{L-1}(\gamma\lambda)^i
       [r_i+\gamma(1-\lambda)v_{i+1}]
       +(\gamma\lambda)^L v_L.
\end{equation}
The last term is essential: at $\lambda=1$ this is the uncorrected $L$-step return including $\gamma^L v_L$; at $\lambda=0$ it is one-step. By contrast, $c_i\equiv1$ in \cref{eq:finite-cap-td-target} is the finite-cap TD-error trace with unit coefficients $c_i=1$, retaining $(\gamma\lambda)^i$ weights and target $\pi$.

\paragraph{SAC soft residual and actor update.}
For SAC \citep{haarnoja2018soft}, replace $v_i$ in the TD-error constructions by
\begin{equation}\label{eq:sac-soft-bootstrap}
v_i^{\mathrm{soft}}=
\E_{a\sim\pi_\phi(\cdot\mid s_i)}
\left[\min_{j=1,2}Q_{\bar\theta_j}(s_i,a)
      -\alpha_{\mathrm{ent}}\log\pi_\phi(a\mid s_i)\right],
\end{equation}
and set $\delta_i^{\mathrm{soft}}=r_i+\gamma v_{i+1}^{\mathrm{soft}}-q_i$, with $q_i=\min_jQ_{\bar\theta_j}(s_i,a_i)$. An action drawn freshly from $\pi_\phi$ gives the usual Monte Carlo estimate of this bootstrap. MINT and Retrace then use the same sum and backward recursion with this soft residual. The stopping rule and cap boundary are unchanged. This is an explicitly entropy-regularized extension; the main-text evaluation results concern the ordinary Bellman residual and $Q^\pi$ stated there.

For a full maximum-entropy version of the reward-return baselines, intermediate entropy terms must also be included. With $e_i=-\alpha_{\mathrm{ent}}\log\pi_\phi(a_i\mid s_i)$ for stored actions $1\le i<L$, the corresponding uncorrected soft return is
\[
\sum_{i=0}^{L-1}\gamma^i r_i
 +\sum_{i=1}^{L-1}\gamma^i e_i+\gamma^L v_L^{\mathrm{soft}}.
\]
The finite mixture of these soft returns has boundary $P_{L-1}^{\mathrm{soft}}=r_{L-1}+\gamma v_L^{\mathrm{soft}}$ and recursion
\[
P_i^{\mathrm{soft}}=r_i+\gamma\left[(1-\lambda)v_{i+1}^{\mathrm{soft}}
                    +\lambda(P_{i+1}^{\mathrm{soft}}+e_{i+1})\right]
\quad(i<L-1).
\]
These formulas specify full soft-return versions; inserting a soft value only at the end of an ordinary reward sum defines a different target. No entropy of the fixed starting action is added to a $Q$-target.

The SAC actor is optimized by gradient ascent on
\begin{equation}\label{eq:sac-actor-objective}
\E_{s\sim\mathcal B,\,a\sim\pi_\phi(\cdot\mid s)}
\left[\min_jQ_{\theta_j}(s,a)
      -\alpha_{\mathrm{ent}}\log\pi_\phi(a\mid s)\right],
\end{equation}
using online critics and reparameterized actor actions. The multi-step critic target is detached during critic regression; it is not substituted for the actor's differentiable critic objective. Here $\alpha_{\mathrm{ent}}$ is the SAC entropy coefficient, distinct from the discount--trace product $\gamma\lambda$.

\paragraph{TD3 bootstrap and the policy used in the ratio.}
For TD3 \citep{pmlr-v80-fujimoto18a}, the bootstrap has no entropy term and uses the clipped double target critic at a smoothed target action:
\[
\begin{aligned}
\tilde a_i&=\operatorname{clip}_{\mathcal A}
\!\left(\mu_{\bar\phi}(s_i)+\operatorname{clip}(\epsilon_i,-c,c)\right),
\qquad \epsilon_i\sim\mathcal N(0,\sigma_{\mathrm{targ}}^2I),\\
v_i^{\mathrm{TD3}}&=\E[\min_jQ_{\bar\theta_j}(s_i,\tilde a_i)].
\end{aligned}
\]
During TD3 data collection, the behavior policy adds Gaussian exploration noise with standard deviation $0.1$ to the actor output and clips the action to the valid bounds. Its log-probability $\log\beta(a_i\mid s_i)$ is computed at collection time and stored with the transition. At replay, the same stored action is evaluated under the smoothed target-actor law above, with $\sigma_{\mathrm{targ}}=0.2$ and noise clip $c=0.5$. Both probability evaluations account for clipping, including the induced probability masses at clipping boundaries. The stopping rule uses the stored behavior value and the replay-time target value through $\min(1,\pi/\beta)=\exp(\min(0,\log\pi-\log\beta))$. During uniform-random warm-up, the stored behavior value is that of the uniform action law. Thus the ratio compares the stochastic collection and target-action laws, rather than assigning a Gaussian density to the deterministic actor itself.

For true terminal transitions set the subsequent bootstrap and continuation to zero. A deterministic trace cap retains the bootstrap at its last observed next state; a time-limit truncation should not be treated as a true terminal transition solely because the recording window ended.
\endgroup

\paragraph{Methods and Hyperparameters.}
We use two one-step RL algorithms, SAC and TD3, for the multi-step extension.

\begin{enumerate}
    \item \textbf{Twin-Delayed Deep Deterministic Policy Gradient (TD3).\quad}
TD3 \citep{pmlr-v80-fujimoto18a} adopts the same training pipeline and neural network architecture as DDPG, but introduces several improvements to address overestimation bias in Q-learning. TD3 uses two critic networks, $Q_{\theta_1}(s, a)$ and $Q_{\theta_2}(s, a)$, with independent parameter sets $\theta_1$ and $\theta_2$. This twin-critic design follows double Q-learning \citep{hasselt2010double}.
\item \textbf{Soft Actor-Critic (SAC).\quad}
SAC \citep{haarnoja2018soft} adopts the same training pipeline and architecture as DDPG/TD3 but introduces a maximum-entropy objective. SAC also maintains two critic networks to reduce overestimation bias.
\end{enumerate}

We adopt the default hyperparameters from the OpenAI SpinningUp codebase. The cap length denotes the upper limit of the sub-trajectory length for the baseline algorithms (uncorrected $n$-step, Retrace, PQL). Detailed values are reported below.

\begin{table}[!htbp]
\centering
\caption{TD3 Hyperparameters}
\begin{tabular}{ll}
\toprule
\textbf{Hyperparameter} & \textbf{Value} \\
\midrule
Actor learning rate & $1 \times 10^{-3}$ \\
Critic learning rate & $1 \times 10^{-3}$ \\
Batch size & 100 \\
Replay buffer size & $1 \times 10^6$ \\
Discount factor $\gamma$ & 0.99 \\
Polyak averaging coefficient ($\tau$) & 0.995 \\
Target policy noise (stddev) & 0.2 \\
Target noise clip & 0.5 \\
Policy update delay (frequency) & 2 steps \\
Exploration noise (initial stddev) & 0.1 \\
Action range & [-1, 1] \\
Start steps (before training begins) & 10000 \\
Max episode length & 1000 \\
cap length & 5 \\
$\lambda$ & 0.7 \\
\bottomrule
\end{tabular}
\label{tab:td3_hyperparams}
\end{table}

\begin{table}[!htbp]
\centering
\caption{SAC Hyperparameters \rev{(MuJoCo main experiments). The entropy coefficient is \emph{fixed} at $\alpha{=}0.2$ (the SpinningUp default) for \texttt{MINT} and all multi-step baselines; no automatic entropy tuning is used in these runs. \textcolor{black}{The DMC ablation and RWRL experiments use their separately stated entropy settings; the fixed-$\alpha$ statement here is restricted to the main MuJoCo SAC comparison.}}}
\begin{tabular}{ll}
\toprule
\textbf{Hyperparameter} & \textbf{Value} \\
\midrule
Actor learning rate & $1 \times 10^{-3}$ \\
Critic learning rate & $1 \times 10^{-3}$ \\
Entropy coefficient $\alpha$ & \rev{$0.2$ (fixed; no automatic tuning)} \\
Batch size & 100 \\
Replay buffer size & $1 \times 10^6$ \\
Discount factor $\gamma$ & 0.99 \\
Polyak averaging coefficient ($\tau$) & 0.995 \\
Target update interval & Every 1 step \\
Start steps (before policy used) & 10000 \\
Action range & [-1, 1] \\
Max episode length & 1000 \\
cap length & 5 \\
$\lambda$ & 0.7 \\
\bottomrule
\end{tabular}
\label{tab:sac_hyperparams}
\end{table}

\paragraph{Algorithm.}
The detailed multi-step target construction is given as \cref{alg:t4} in the main text (\cref{sec:t4-practical}). It replaces the multi-step critic target in the SAC and TD3 implementations evaluated here; \textcolor{black}{the trace construction computes $\hat p_i$, draws the keep masks, and selects the retained prefix. Its critic target is the finite-cap residual sum in \cref{eq:finite-cap-td-target}, with the explicit SAC or TD3 bootstrap substitution above.}

\paragraph{Experimental Details.}
We implement \texttt{MINT} and the baselines in PyTorch and evaluate them using the standard off-policy RL protocol in Google Dopamine \citep{castro18dopamine}. \rev{Code is available at \url{https://github.com/seungyubhan/mint}.}

\paragraph{\rev{Numerical treatment of the policy ratio.}}
Here, log-probability denotes log-mass for a discrete policy and log-density for a continuous policy. \rev{For SAC's tanh-squashed Gaussian actors, evaluating $\pi(a_i\mid s_i)/\beta(a_i\mid s_i)$ at a stored action requires three precautions. (i) \emph{Stored behavior log-probability.} $\log\beta(a_i\mid s_i)$ is computed and stored at collection time (during the uniform warm-up it is the log-probability under the uniform box distribution), so $\beta$ is always the data-generating policy of that transition and no density model of the drifting replay distribution is needed (\cref{alg:t4-full}). (ii) \emph{Squash correction.} The log-probability of a stored action under $\pi$ is evaluated by inverting the squash, $u=\operatorname{atanh}(a/a_{\max})$ with the argument clipped to $[-1+10^{-6},\,1-10^{-6}]$ to guard the saturated boundary, and adding the tanh log-Jacobian; the policy log-standard-deviation is clamped to $[-20,2]$. (iii) \emph{Log-space ratio.} The clipped ratio is formed as $\min(1,\pi/\beta)=\exp\bigl(\min(0,\log\pi-\log\beta)\bigr)$: clamping the log-ratio at $0$ is exactly the minimum against $1$ and removes overflow, while underflow of the exponential for very negative log-ratios yields $\hat p_i\to1$, the correct limit rather than a NaN. $\hat p_i$ is finally clamped to $[0,1]$ and the anchor step $t=0$ is always kept. These steps keep the ratio stable when $\pi$ becomes sharply peaked relative to $\beta$ (the saturation regime of \cref{app:cap-signal}).}

\addtocontents{apc}{\protect\newpage}
\section{Additional Experimental Results}\label{sec:additional_exp}
\addtocontents{apc}{\protect\setcounter{tocdepth}{3}}

\subsection{MuJoCo: Evaluation Protocol and Additional Results}\label{app:mujoco-results}

\begin{figure}[!htbp]
    \centering
    \includegraphics[width=\textwidth]{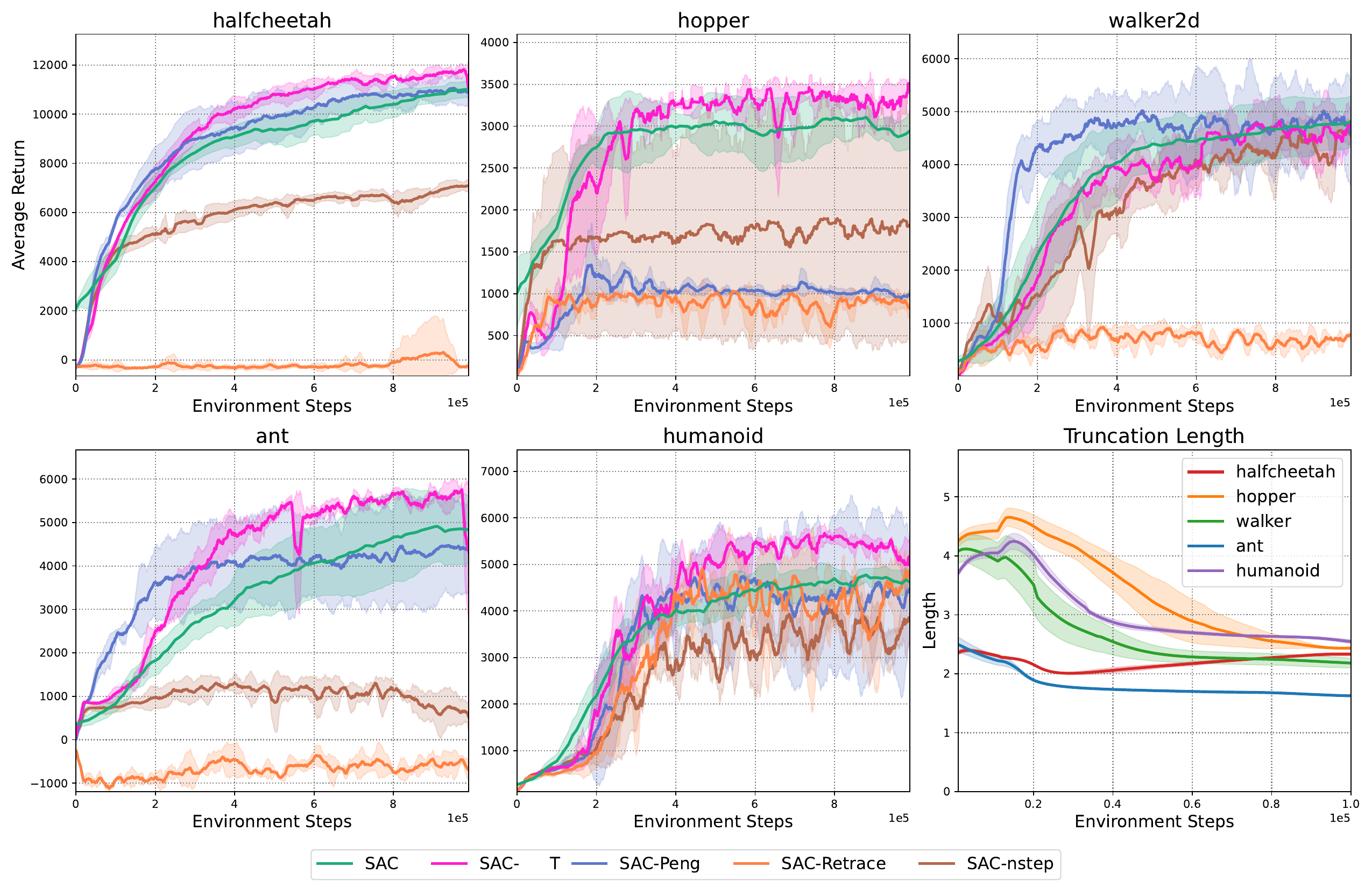}
    \caption{\textbf{Performance with stochastic truncation.}  Across five MuJoCo benchmarks, \texttt{MINT} attains higher returns than the multi-step baselines on most tasks and at earlier training checkpoints. For baseline comparisons, we follow the convention of \citep{kozuno2021revisiting} and fix the cap length to $n=5$ for all multi-step methods.}
    \label{fig:mujoco_sac}
\end{figure}

\begin{figure}[!htbp]
    \centering
    \includegraphics[width=\textwidth]{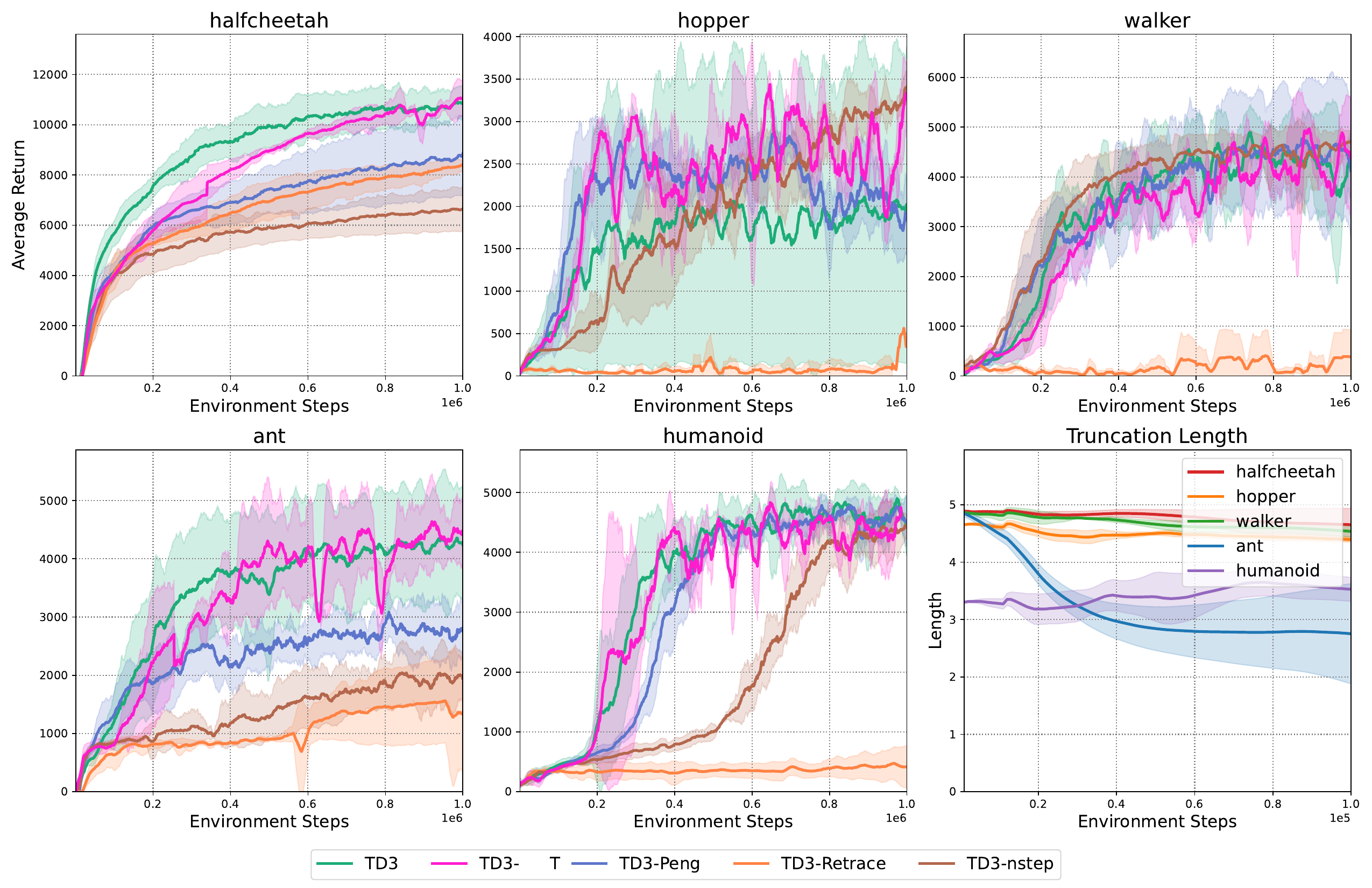}
    \caption{\textbf{Evaluation of TD3 variants over OpenAI MuJoCo environments.}}
    \label{fig:mujoco_td3}
\end{figure}

\begin{figure}[!htbp]
    \centering
    \includegraphics[width=\linewidth]{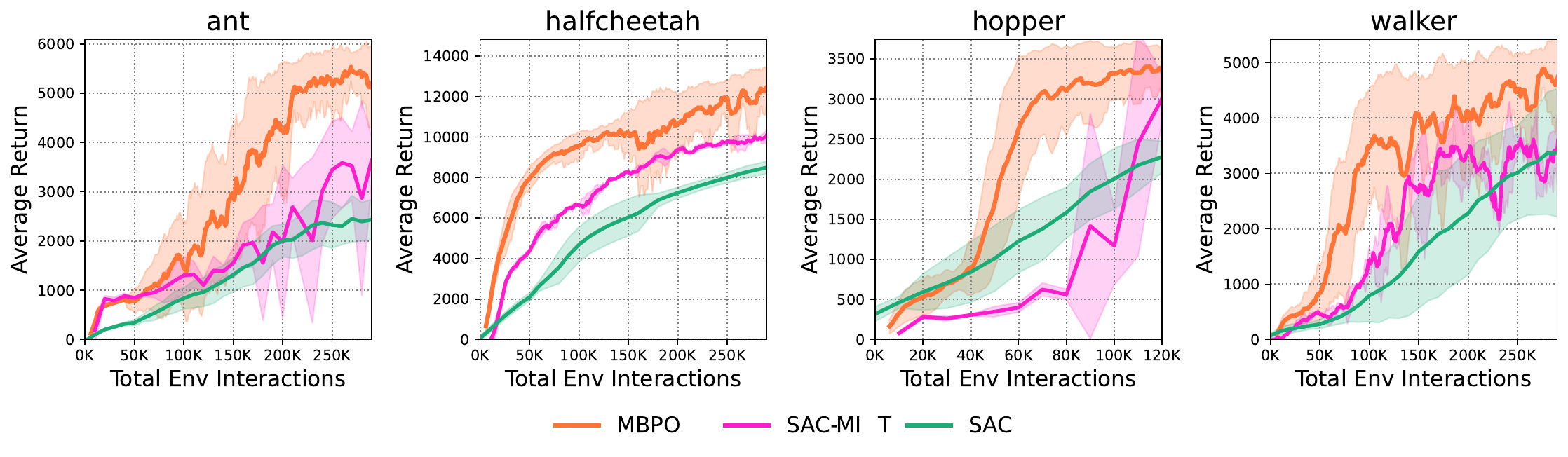}
    \caption{\textbf{Comparison of SAC, SAC-\texttt{MINT}, and MBPO across four MuJoCo benchmarks.} We use different multi-step configurations per environment: \textit{halfcheetah} and \textit{walker} use a standard multi-step setting (cap length $=5$, $\lambda=0.7$), whereas \textit{ant} and \textit{hopper} follow the longer-horizon model-based configuration inspired by MBPO (cap length $=25$, $\lambda=0.1$). The curves compare the sample efficiency of SAC, \texttt{MINT}, and MBPO under these task-specific settings; \texttt{MINT} often attains returns comparable to MBPO while using a model-free algorithm.}
    \label{fig:mbpo}
\end{figure}

\subsubsection{Evaluation Protocol and Score Normalization}\label{app:iqm}

We conducted experiments following the evaluation protocol proposed in \citep{agarwal2021deep}. This protocol emphasizes robust evaluation through interquartile mean (IQM) scores and normalized performance.

\textcolor{black}{The main MuJoCo comparison uses the five-seed-per-method-and-task protocol stated in \cref{app:compute}, with scores aggregated after min--max normalization. Additional experiments report their own seed counts.} The main learning curves cover 1{,}000K environment interactions; \cref{tab:ref_scores} lists the reference scores used for normalization. Metric computation was conducted using the official codebase of \citep{agarwal2021deep}. For min--max normalization, expert and random scores were taken from the Minari extension \citep{minari2022} of D4RL \citep{fu2020d4rl}; the Humanoid benchmark follows Minari scores, while the remaining tasks use D4RL values.

Across both SAC- and TD3-based experiments, \texttt{MINT} attains higher IQM at earlier training checkpoints than the compared multi-step methods and baselines. Only \texttt{MINT} surpasses the expert score within the reported 1{,}000K-step budget.

\begin{table}[!htbp]
\centering
\caption{\textbf{Reference scores for min--max normalization.} Random and expert performance values are taken from D4RL \citep{fu2020d4rl} and Minari \citep{minari2022}.}
\vspace{0.3em}
\begin{tabular}{lcc}
\toprule
Environment & Random Score & Expert Score \\
\midrule
\texttt{Hopper}      & $-20.27$   & $3234.3$ \\
HalfCheetah & $-280.18$  & $12135.0$ \\
\texttt{Walker}2d    & $1.63$     & $4592.3$ \\
Ant         & $-325.6$   & $3879.7$ \\
Humanoid    & $78.85$    & $9024.95$ \\
\bottomrule
\end{tabular}
\label{tab:ref_scores}
\end{table}

\subsubsection{Sensitivity to \texorpdfstring{$\lambda$}{lambda}}\label{app:lambda-sweep}

$\lambda$ is shared by \texttt{MINT}, Peng's $Q(\lambda)$, and Retrace: the main table runs \texttt{MINT}, Peng's $Q(\lambda)$, and Retrace at the same $\lambda=0.7$, chosen to match the Peng's $Q(\lambda)$ protocol rather than to optimize our operator. It is nevertheless a hyperparameter, setting the $(\gamma\lambda)^t$ weight on every term of the backup. \Cref{tab:lambda-sweep} reports the full sweep on MuJoCo \texttt{Ant-v2} and \texttt{Hopper-v2} (fixed $\alpha=0.2$, 1M steps, final return averaged over the last five evaluations, mean $\pm$ s.d.\ across seeds) without selecting favorable settings. Return decreases with $\lambda$ on \texttt{Ant-v2} and is comparatively stable on \texttt{Hopper-v2} (about $9\%$ variation over the range); from $\lambda=0.7$ to $0.9$ the change is $-12.9\%$ on \texttt{Ant-v2} and $+3.5\%$ on \texttt{Hopper-v2}, with overlapping mean $\pm$ s.d.\ ranges in both cases. Two qualifications apply: $\lambda=0.7$ is not the best setting in this sweep (the two highest mean returns on \texttt{Ant-v2} occur at $0.1$ and $0.3$), consistent with its having been fixed for protocol matching; and because the baselines were not swept, we make no claim about how the gap between methods changes at other $\lambda$. \citet{kozuno2021revisiting} likewise find Peng's $Q(\lambda)$ sensitive to $\lambda$ on DMControl (their Appendix J.8), with the best overall performance near $\lambda=0.7$, so $\lambda$ sensitivity is not unique to \texttt{MINT}. In the reported sweep the cap is fixed at $5$ independently of $\lambda$ (\cref{eq:effective-K}). For fixed policies and a fixed trajectory, $\lambda$ changes the $(\gamma\lambda)^t$ weights without changing the stopping rule; across training runs, it can also affect the learned critic and policy.

\begin{table}[!htbp]
\centering\small
\caption{$\lambda$ sweep for SAC-\texttt{MINT} (fixed $\alpha=0.2$, $K_{\max}=5$, 1M steps; mean $\pm$ s.d.\ across \revr{2--3} seeds \revr{per cell}). The main-table value is $\lambda=0.7$.}
\label{tab:lambda-sweep}
\begin{tabular}{lccccc}
\toprule
task & $\lambda=0.1$ & $\lambda=0.3$ & $\lambda=0.5$ & $\lambda=0.7$ & $\lambda=0.9$ \\
\midrule
\texttt{Ant-v2}    & $5812\pm81$ & $5991\pm92$ & $5466\pm98$ & $3922\pm553$ & $3417\pm1594$ \\
\texttt{Hopper-v2} & $3453\pm8$  & $3291\pm60$ & $3280\pm3$  & $3151\pm161$ & $3262\pm137$ \\
\bottomrule
\end{tabular}
\end{table}

\subsection{DMC: Main Results and Additional Diagnostics}
\label{app:dmc11}

\subsubsection{11-Task Evaluation: Setup and Task Selection}\label{app:dmc11-setup}
\paragraph{Task selection and scope.}\label{app:dmc-task-selection}
Our eleven-task evaluation is a study-specific subset of the DeepMind Control Suite \citep{tassa2018deepmindcontrolsuite}, rather than a separate official benchmark. We selected tasks from the 20-task DMC Easy \& Medium set used by SimBa \citep{lee2025simba}: six tasks with indicator or height-gated reward (\texttt{Cartpole swingup sparse}, \texttt{Pendulum swingup}, \texttt{Finger turn easy}/hard, and \texttt{Hopper stand}/hop), and five dense-reward tasks with action dimension at most six (\texttt{Walker walk}/run, \texttt{Cheetah run}, \texttt{Acrobot swingup}, and \texttt{Fish swim}). The dense-task choice considered published SAC performance indicating potential for improvement and the availability of existing task implementations; this is a purposive subset, not a random or exhaustive sample of DMC. The comparison applies a common protocol to all eleven selected tasks. The evaluated subset excludes Ball-in-Cup catch and \texttt{Acrobot swingup} sparse. \Cref{tab:dmc11-tasks} identifies the exact evaluation set. The reward-frequency groups reported later are a separate descriptive measurement under a uniform-random policy; they are not official DMC task classes. Conclusions are restricted to this set and protocol.

\paragraph{Common protocol.}
We train eleven DMC tasks with five seeds per method/task (165 runs) for 500K environment steps. All methods use $\gamma=0.99$, automatic entropy tuning initialized at $\alpha=0.2$, batch size 256, two 256-unit hidden layers, learning rates $3\times10^{-4}$, and one update per step after 5K steps. We evaluate ten episodes every 10K steps. Final performance averages the last five evaluations within each run; task tables report the mean and standard deviation over five seeds. Aggregate scores divide returns by $1000$. No action repeat is applied. \texttt{MINT} uses $\lambda=0.9$, $K_{\max}=10$, and stored collection-time behavior log-probabilities. SAC-multistep is a pure uncorrected ten-step return with $\gamma^{10}$ bootstrapping; it does not use $\lambda$ weighting. Consequently, this comparison does not isolate stochastic stopping as the only changed target ingredient. Only runs using this common protocol contribute to the reported aggregates.

\begin{table}[!htbp]
\centering\small
\setlength{\tabcolsep}{4pt}
\caption{\textbf{DMC evaluation protocols.} Supplementary evaluation scores are not pooled into the primary eleven-task aggregates. The supplementary evaluation's five-seed curves and \texttt{CartPole} re-run retain their separately reported configurations.}
\label{tab:dmc-protocols}
\begin{tabular}{@{}p{0.15\textwidth}p{0.22\textwidth}p{0.32\textwidth}p{0.23\textwidth}@{}}
\toprule
\raggedright Evaluation & \raggedright Steps and seeds & \raggedright Target settings & \raggedright Score window\tabularnewline
\midrule
\raggedright Primary eleven-task & \raggedright 500K; five seeds per method/task & \raggedright Cap $10$, $\gamma=0.99$, automatic $\alpha$; \texttt{MINT} $\lambda=0.9$ & \raggedright Last five evaluations; aggregate return/$1000$\tabularnewline[4pt]
\raggedright Supplementary five-task & \raggedright 1M; three-seed table, separate five-seed curves and \texttt{CartPole} re-run & \raggedright Cap $5$ on \texttt{CartPole}, $10$ elsewhere; $\gamma=0.999$ on sparse tasks, $0.99$ on \texttt{Hopper-Stand}; automatic $\alpha$, \texttt{MINT} $\lambda=0.9$ & \raggedright Last 100K steps (mean$_{10}$); raw returns\tabularnewline
\bottomrule
\end{tabular}
\end{table}

\subsubsection{11-Task Performance and Task Groups}\label{app:dmc11-performance}

\begin{table}[!htbp]\centering\small
\caption{\textbf{11-task DMC final returns by reward accessibility.} Last-five-evaluation average at 500K steps; mean $\pm$ standard deviation across five seeds. Groups use the fraction of uniform-random episodes containing positive reward. Gray literature stars in the task-curve figure use a different protocol and are not matched baselines.}
\label{tab:dmc11-tasks}
\begin{tabular}{lrrr}\toprule
Task & SAC 1-step & SAC 10-step & SAC-\texttt{MINT}\\\midrule
\multicolumn{4}{l}{\textbf{Rare} ($<5\%$; 2 tasks)}\\[2pt]
\texttt{Cartpole swingup sparse} & $431\pm353$ & $291\pm229$ & $493\pm196$\\
\texttt{Pendulum swingup} & $249\pm176$ & $560\pm264$ & $284\pm253$\\
\midrule
\multicolumn{4}{l}{\textbf{Occasional} ($10$--$50\%$; 4 tasks)}\\[2pt]
\texttt{Finger turn easy} & $492\pm54$ & $498\pm43$ & $663\pm62$\\
\texttt{Finger turn hard} & $323\pm59$ & $288\pm74$ & $541\pm105$\\
\texttt{Hopper stand} & $345\pm138$ & $25\pm39$ & $374\pm364$\\
\texttt{Hopper hop} & $65\pm32$ & $2\pm3$ & $105\pm75$\\
\midrule
\multicolumn{4}{l}{\textbf{Frequent} ($100\%$; 5 tasks)}\\[2pt]
\texttt{Walker walk} & $864\pm121$ & $636\pm63$ & $920\pm79$\\
\texttt{Walker run} & $554\pm156$ & $273\pm81$ & $635\pm59$\\
\texttt{Cheetah run} & $711\pm22$ & $476\pm48$ & $626\pm105$\\
\texttt{Acrobot swingup} & $4\pm2$ & $48\pm20$ & $50\pm10$\\
\texttt{Fish swim} & $183\pm46$ & $232\pm90$ & $195\pm26$\\
\bottomrule\end{tabular}\end{table}

\paragraph{Aggregate learning and distribution.}
The aggregate in \cref{fig:dmc11-main} uses final-window scores. The IQM curve instead aggregates scores at each evaluation checkpoint; its final point can differ from the final-window IQM. The performance profile pools the 55 task/seed final scores per method. At normalized score $0.5$, \texttt{MINT} exceeds the threshold in $28/55$ runs, versus $21/55$ and $15/55$ for one-step and ten-step. These are descriptive threshold comparisons, not proof of dominance at every threshold.

\begin{figure}[!htbp]
\centering
\includegraphics[width=\textwidth]{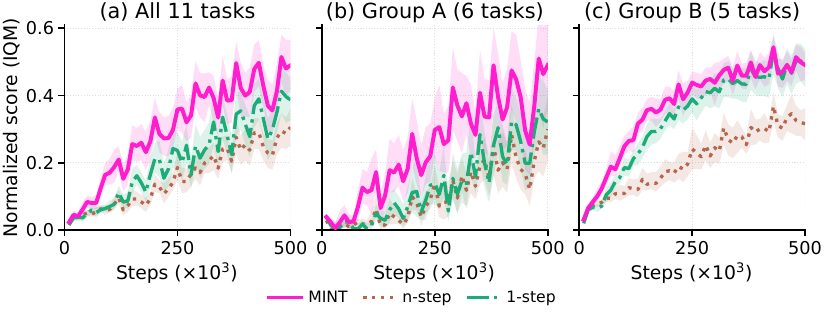}
\caption{\textbf{11-task DMC evaluation-time IQM.} All eleven tasks and the original six-/five-task A/B split. Bands are pointwise $95\%$ stratified-bootstrap intervals; final-window aggregate scores average five evaluations per run.}
\label{fig:dmc11-learning}
\end{figure}
\begin{figure}[!htbp]
\centering
\includegraphics[width=\textwidth]{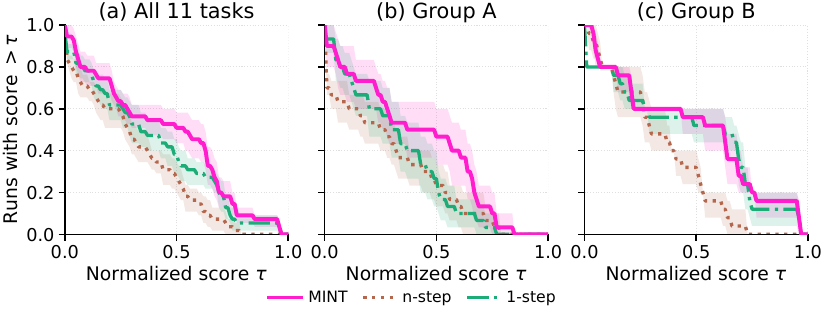}
\caption{\textbf{11-task DMC final-score profiles.} Fraction of pooled task/seed runs above return/$1000$ threshold $\tau$, with pointwise $95\%$ bootstrap bands. Each task contributes five seeds.}
\label{fig:dmc11-profile}
\end{figure}
\begin{figure}[!htbp]
\centering
\includegraphics[width=\textwidth]{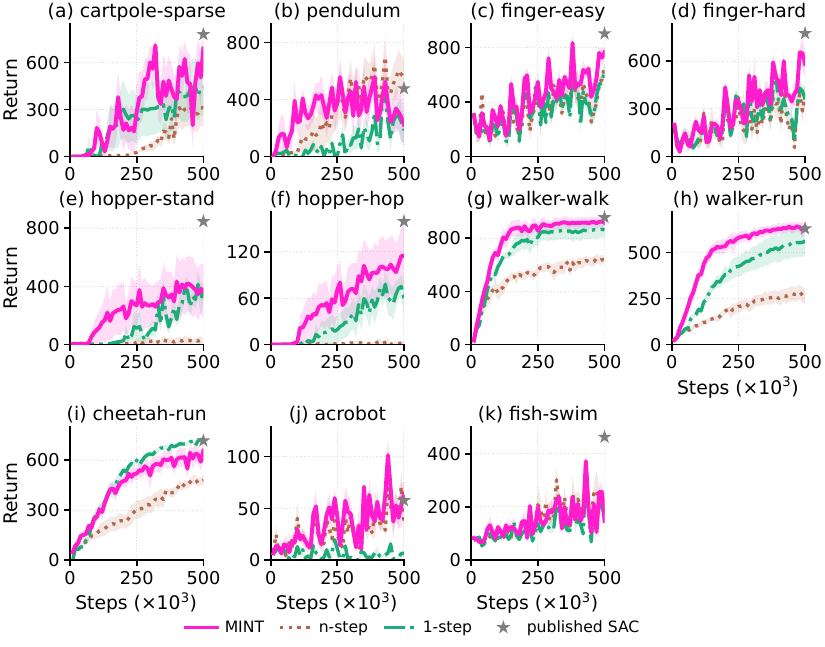}
\caption{\textbf{11-task DMC task learning curves.} Mean over five seeds; bands are one standard error, not $95\%$ intervals. Each evaluation averages ten episodes. Gray stars are published SAC reference values with a different action-repeat protocol.}
\label{fig:dmc11-tasks}
\end{figure}

\paragraph{Task-specific gains and exceptions.}
\texttt{Finger easy}/hard and \texttt{Walker walk}/run favor \texttt{MINT} over both baselines at the reported endpoint. \texttt{Hopper} recovers relative to pure ten-step, but \texttt{Hopper-Stand} has large seed variability. One-step is stronger on \texttt{Cheetah}, and ten-step is stronger on \texttt{Pendulum} and \texttt{Fish}. \texttt{Acrobot-Swingup} has low absolute return despite improvement over one-step; this evaluation uses the dense-reward task, distinct from \texttt{Acrobot-Swingup-Sparse}.

\paragraph{Descriptive task groups.}
Reward accessibility is measured as the fraction of uniform-random episodes containing positive reward. Rare ($<5\%$) comprises \texttt{Cartpole sparse} and \texttt{Pendulum}; occasional ($10$--$50\%$) comprises \texttt{Finger easy}/hard and \texttt{Hopper stand}/hop; frequent ($100\%$) comprises \texttt{Walker walk}/run, \texttt{Cheetah}, \texttt{Acrobot}, and \texttt{Fish}. The corresponding \texttt{MINT} IQMs are $0.392$, $0.453$, and $0.491$, versus one-step $0.320/0.322/0.487$ and ten-step $0.440/0.157/0.327$. The occasional group has probabilities of improvement $0.78\,[0.63,0.91]$ over one-step and $0.88\,[0.74,1.00]$ over ten-step. Action groups contain 3/2/6 tasks; the two-dimensional group is entirely \texttt{Finger}. The learned-$K$ grouping is post hoc. Across the six rare/occasional tasks, \texttt{MINT} improves mean return over one-step in all six and over ten-step in five; \texttt{Pendulum} is the exception to the latter comparison. The strictly rare group contains only two tasks and is reported separately. These partitions describe reward accessibility under a random policy, not the exploration taxonomy used for Atari, and do not establish causal effects of reward frequency, action dimension, or length.

\begin{figure}[!htbp]
\centering
\includegraphics[width=\textwidth]{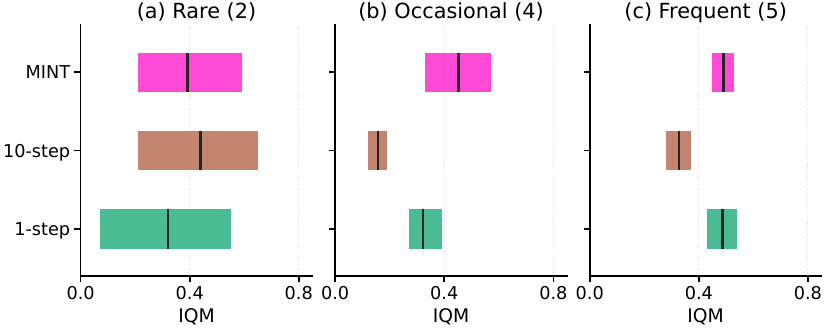}
\caption{\textbf{DMC reward-accessibility groups.} Endpoint IQM with $95\%$ confidence-interval rectangles and black estimate ticks. Group sizes are 2/4/5; grouping is descriptive rather than a universal exploration taxonomy.}
\label{fig:dmc11-reward}
\end{figure}
\begin{figure}[!htbp]
\centering
\includegraphics[width=\textwidth]{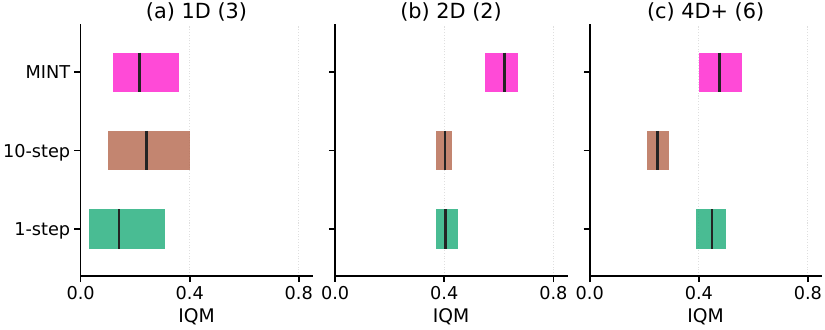}
\caption{\textbf{DMC action-dimensionality groups.} Endpoint IQM, bootstrap intervals and estimates. The two-dimensional group contains only \texttt{Finger} tasks, confounding action dimension with task identity.}
\label{fig:dmc11-action}
\end{figure}
\begin{figure}[!htbp]
\centering
\includegraphics[width=\textwidth]{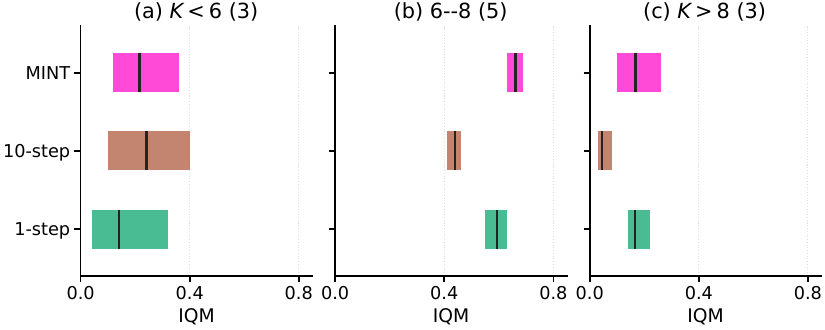}
\caption{\textbf{DMC learned-horizon groups.} Endpoint IQM grouped by \texttt{MINT} mean length after training. Grouping is post hoc and does not certify an optimal horizon.}
\label{fig:dmc11-khat}
\end{figure}

\FloatBarrier
\subsubsection{Mismatch--Horizon Diagnostics and Reference Construction}\label{app:dmc11-proxy}
\Cref{fig:dmc11-trace-tradeoff} relates measured trace lengths to conditional mismatch references. For logged mean keep $p$ and $\lambda=0.9$, a state-homogeneous symmetric-binary closure gives
\[
\chi_\lambda^{\rm proxy}=\frac{[(1+2\lambda)p-1]_+}{2\lambda p}.
\]
The vertical coordinate $J$ is the logged mean length. Let $S_L(u)=\sum_{t=0}^{L-1}u^t$, solve $S_{10}(q)=J$, and set $M_\gamma(J)=S_{10}(\gamma\lambda q)$. Green contours count the fraction of $\gamma\in\{0.01,0.02,\ldots,0.99\}$ satisfying
\[
2\chi_\lambda^{\rm proxy}\bigl(1-M_\gamma(J)^{-1}\bigr)<1-\gamma.
\]
Logged $p$ and reconstructed $q$ can differ. Thus these contours combine two conditional approximations and are neither sample-wise contraction certificates, convergence probabilities, nor measured operator gains. Gray curves fit monotone cubic-spline 90th-quantile regression to all positive-proxy records within each panel (2,696 records in total), then add $0.6$ steps for display. They are not confidence bounds; solid/dashed portions distinguish supported and sparse ranges. Of these records, 457 are displayed; hollow/filled marks distinguish steps at or below/above 140K.

At $\gamma_0=0.99$, blue guides combine a checkpoint block-transition proxy $\hat\rho_{\rm eff}$ on visited support with trace-mass bounds. This is not uniform minorization. Task-specific $C_e$ is calibrated to at least $50\%$ coverage of broadly visited checkpoint cells; uncapped $K_{\rm ref}$ values are summarized by task medians and equal-task group means ($3.82$, $6.08$, $7.77$). The two \texttt{Finger} tasks have only three and two calibration cells, respectively. Same-data calibration provides conditional references, not independent evidence of noise reduction. The on-policy motivation from \citet{duan2022policy} does not transfer its guarantee to off-policy learning or these empirical proxies.

\subsubsection{Target Operators and Bias--Variance Diagnostics}\label{app:dmc11-bv}
\paragraph{Weighted operator definitions.}
For a shared behavior segment, let $q_t=Q_{\bar\theta}(s_t,a_t)$ and let $v_{t+1}$ be the sampled soft target-policy value used by the diagnostic code. Define $\delta_t=r_t+\gamma v_{t+1}-q_t$. The fixed-length $\lambda$-weighted TD-error trace (\texttt{lam} in the diagnostic code) is
\begin{equation}\label{eq:dmc11-lambda-target}
G_K^\lambda=q_0+\sum_{t=0}^{K-1}(\gamma\lambda)^t\delta_t
=r_0+\gamma v_1+\sum_{t=1}^{K-1}(\gamma\lambda)^t(r_t+\gamma v_{t+1}-q_t).
\end{equation}
This is a finite-cap $Q^\pi(\lambda)$-type TD-error trace \citep{harutyunyan2016q}, retaining the $\lambda$ weights but using unit importance-trace coefficients. The code evaluates the same sum as $\sum_{t<K}(\gamma\lambda)^t[r_t+\gamma v_{t+1}-\gamma\lambda\mathbf1\{t<K-1\}q_{t+1}]$. \texttt{MINT} selects $G_{\hat K}^\lambda$, where $\hat K$ is the retained prefix from Bernoulli draws with keep probability $1-\min(1,\pi(a_t\mid s_t)/\beta(a_t\mid s_t))$. Its random length depends on the segment. Pure $n$-step instead uses $N_n=\sum_{t<n}\gamma^tr_t+\gamma^n v_n$.

Peng's finite-cap mixture is $P_K^\lambda=(1-\lambda)\sum_{n=1}^{K-1}\lambda^{n-1}N_n+\lambda^{K-1}N_K$ \citep{kozuno2021revisiting}, as in \cref{eq:peng-finite-cap}. With shared $v_t$, the pathwise difference is
\begin{equation}\label{eq:dmc11-peng-difference}
G_K^\lambda-P_K^\lambda=\sum_{t=1}^{K-1}(\gamma\lambda)^t(v_t-q_t).
\end{equation}
Thus the diagnostic's \texttt{lam} curve, including its $K=10$ square, is not a measurement of Peng's return mixture off policy. The planes use ``fixed $\lambda$-TD trace'' and ``fixed trace ($K=10$)'' accordingly. No Peng-mixture target comparison is inferred from those points. This distinction does not change the independently trained Peng baselines in the MuJoCo experiments.

\paragraph{Shared evaluation and aggregation.}
Policy and learned target critics are frozen at 100K, 200K, 300K, or 500K checkpoints of the SAC-\texttt{MINT} run; these are not oracle-critic measurements. Behavior segments are generated with that checkpoint and its available predecessors, rather than drawn from the replay buffer. Operators share segments within each evaluated cell. Bias is mean target minus a finite-rollout soft-$Q^\pi$ Monte-Carlo reference; finite-rollout and sampling error remain in that reference. The plane plots the absolute value of the seed-aggregated signed bias, rather than mean absolute error. Target spread is the within-anchor standard deviation aggregated over anchors and behavior checkpoints by a root-mean-variance calculation within each seed, then averaged over seeds. \texttt{MINT} spread also includes randomness in the stopping draws. Neither coordinate has a confidence interval in these planes. The plotted bias is a finite-backup diagnostic with a learned critic, not the fixed-point bias of the population operator. Its change therefore need not contradict preservation of $Q^\pi$ in fixed-policy evaluation. These target coordinates complement the control returns rather than establish their cause. Values are expressed as percentages relative to $100=1/(1-\gamma)$; this is not an upper bound on entropy-regularized value. The same seed set is used for all operators in each task--stage panel.

\paragraph{Stage-specific exclusions.}
The aggregation excludes a seed at a stage if any included behavior checkpoint has absolute one-step target bias divided by $100$ above $0.5$. This is a diagnostic filter, not an independent certification that the remaining critic is accurate. It applies to target diagnostics only; the control-performance aggregates keep all five seeds. \Cref{tab:dmc11-bv-seeds} gives retained counts and excluded seed identities. In particular, \texttt{Pendulum} at 500K retains only seed 0; \texttt{Finger-Hard} retains two seeds. These late-stage panels are descriptive conditional estimates.

\begin{table}[!htbp]\centering\small
\caption{\textbf{Seeds retained in target diagnostics.} Each cell gives the retained count, followed by excluded zero-based seed IDs in parentheses when nonempty. Training and control-score aggregates use all five seeds.}
\label{tab:dmc11-bv-seeds}
\begin{tabular}{lrrrr}\toprule
Task & 100K & 200K & 300K & 500K\\\midrule
\texttt{Cartpole sparse} & 5 & 5 & 5 & 5\\
\texttt{Pendulum} & 5 & 4 (2) & 3 (2,3) & 1 (1,2,3,4)\\
\texttt{Finger easy} & 4 (1) & 5 & 5 & 4 (2)\\
\texttt{Finger hard} & 3 (0,1) & 3 (0,4) & 4 (3) & 2 (0,1,2)\\
\texttt{Hopper stand} & 5 & 5 & 4 (1) & 4 (1)\\
\texttt{Hopper hop} & 5 & 4 (1) & 4 (1) & 4 (1)\\
\texttt{Walker walk} & 5 & 5 & 5 & 5\\
\texttt{Walker run} & 5 & 5 & 5 & 5\\
\texttt{Cheetah} & 5 & 5 & 5 & 5\\
\texttt{Acrobot} & 5 & 5 & 5 & 5\\
\texttt{Fish} & 4 (0) & 4 (0) & 5 & 5\\
\bottomrule\end{tabular}\end{table}

\paragraph{Aggregate spread and the bias trade-off.}
At 200K/300K/500K there are 33 task--stage cases. Fixed-$K$ $\lambda$-weighted TD trace spread increases monotonically with length in 32; the median length-ten/length-one standard-deviation ratio is $4.09$. \texttt{MINT} is below the cap-ten TD trace in $26/33$ (median ratio $0.916$), and below pure ten-step in $27/33$ ($0.759$). For multidimensional-action tasks, both counts are $23/24$, with median ratios $0.911$ and $0.733$, respectively. In $20/33$ cases, a lower spread than pure ten-step coexists with no smaller absolute bias. Against an interpolated fixed-$K$ $\lambda$-weighted TD trace at \texttt{MINT}'s measured mean length, the median ratio is $1.077$ (below one in $10/33$). That interpolation is not a randomized matched-length control. The aggregate statistics therefore support reductions relative to long fixed backups without establishing uniform bias reduction or an additional equal-length benefit from adaptive stopping.

\begin{figure}[!htbp]
\centering
\includegraphics[width=\textwidth]{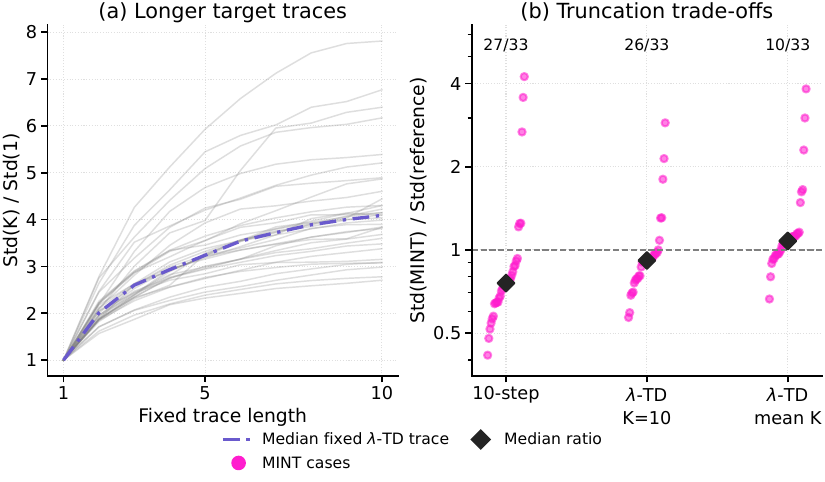}
\caption{\textbf{DMC target-spread diagnostics with the weighted TD-error reference.} 33 task--stage cases at 200K/300K/500K, conditional on the seed filter in \cref{tab:dmc11-bv-seeds}. (a) Fixed-$K$ $\lambda$-weighted TD trace standard deviation divided by its length-one value; gray lines are cases and the violet line is their pointwise median. (b) \texttt{MINT}/reference standard-deviation ratios; black diamonds are medians. The cap-ten reference is the fixed TD trace, not the Peng mixture. The mean-$K$ comparison interpolates the fixed-length standard-deviation curve; it is not a randomized matched-length estimator.}
\label{fig:dmc11-spread}
\end{figure}

\paragraph{Selected lower-spread examples and counterexamples.}
\Cref{fig:dmc11-bv-main} shows \texttt{Finger-Turn-Hard}, \texttt{Hopper-Hop}, \texttt{Cartpole-Swingup-Sparse}, and \texttt{Walker-Run} at 200K. The one-step-bias filter retains 3/4/5/5 seeds: \texttt{Finger-Turn-Hard} excludes seeds 0 and 4; \texttt{Hopper-Hop} excludes seed 1. No additional filter is introduced. \texttt{MINT}/fixed-$K=10$ standard-deviation ratios are $0.699$, $0.806$, $0.704$, and $0.760$. Within-anchor variance is lower in 3/3, 4/4, 3/5, and 5/5 retained seeds, respectively. In particular, the aggregated reduction on \texttt{Cartpole-Swingup-Sparse} does not imply a decrease in every seed. Absolute aggregated bias is higher on \texttt{Finger-Turn-Hard} ($11.051$ versus $10.689$), \texttt{Hopper-Hop} ($1.364$ versus $1.317$), and \texttt{Walker-Run} ($0.714$ versus $0.700$), but lower on \texttt{Cartpole-Swingup-Sparse} ($0.211$ versus $0.218$). These descriptive comparisons do not establish statistical significance or represent all tasks.

All-task planes are collected in \cref{app:dmc11-full-planes}. At 300K, \texttt{Finger-Easy} reduces standard deviation from $1.095$ to $0.650$ relative to the cap-ten TD trace while bias rises from $11.647$ to $11.947$. \texttt{Walker-Walk} reduces both spread ($0.418$ to $0.381$) and bias ($0.253$ to $0.224$); \texttt{Fish} has a modest spread reduction ($0.638$ to $0.593$) with higher bias ($16.485$ to $16.754$). \texttt{Acrobot} is a counterexample: spread rises from $0.068$ to $0.122$ and bias from $2.484$ to $2.595$; pure ten-step is also lower in both coordinates (\cref{fig:dmc11-bv-300k}). The all-task results and $26/33$ lower-spread count provide the broader context for the selected main-text examples.

\FloatBarrier
\subsubsection{Reward Retention and Target Variance Diagnostics}\label{app:dmc-reward-retention}

\paragraph{Protocol and scope.}
We evaluate five training seeds (0--4), fixing the policy and learned target critic from each \texttt{MINT} checkpoint. \Cref{tab:dmc-retention-main-protocol} summarizes the measurement. Each behavior distribution is a single checkpoint rollout, rather than training replay. Policy pairs are reported separately, without pooling the $\pi=\beta$ boundary with off-policy pairs or applying the stage-specific seed filter of \cref{app:dmc11-bv}. The reward-frequency table covers five tasks; the subsequent retention and target tables report four of these tasks (FH, CS, WR, and AS), excluding HH. Task codes are FH: \texttt{Finger-Turn-Hard}; CS: \texttt{Cartpole-Swingup-Sparse}; WR: \texttt{Walker-Run}; AS: \texttt{Acrobot-Swingup}; and HH: \texttt{Hopper-Hop}.

\begin{table}[!htbp]
\centering\small
\caption{Main reward-retention and target-diagnostic protocol.}
\label{tab:dmc-retention-main-protocol}
\setlength{\tabcolsep}{4pt}
\begin{adjustbox}{max width=\linewidth}
\begin{tabular}{@{}lp{0.73\linewidth}@{}}
\toprule
Item & Setting \\
\midrule
Tasks and training seeds & Five tasks (FH, CS, WR, AS, HH); seeds 0--4 \\
Target/behavior pairs $(\pi,\beta)$ & $(200,100)$, $(500,400)$, $(500,100)$, $(500,500)$K steps \\
Measurement cells & $5$ tasks $\times\,5$ seeds $\times\,4$ pairs $=100$ \\
Samples per cell & $64$ anchors; $16$ windows per anchor; $32$ stopping draws per window \\
Monte-Carlo reference & $32$ rollouts; horizon $1000$ \\
Target family & TD-error prefix; $\gamma=0.99$, $\lambda=0.9$, cap $L=10$; no clipping \\
Behavior source & Rollouts from one checkpoint policy per cell \\
Independent repeats & Training seeds; windows and draws are conditional samples \\
Reported intervals & $95\%$ percentile intervals from $2000$ training-seed resamples \\
Missing retention intervals & Not reported when retention is defined in fewer than five seeds \\
\bottomrule
\end{tabular}
\end{adjustbox}

\end{table}

\paragraph{Reward frequency and conditional retention.}
For a reward-bearing window, define
\[
R_{\rm ret}(K)=\frac{\sum_{t<K}(\gamma\lambda)^t|r_t|}{\sum_{t<L}(\gamma\lambda)^t|r_t|},
\qquad \sum_{t<L}(\gamma\lambda)^t|r_t|>0.
\]
The shuffled-$K$ control preserves \texttt{MINT}'s stopping-length distribution while removing its alignment with the behavior window. Retention measures use of already encountered rewards, separately from reward discovery. \Cref{tab:dmc-retention-main-frequency} reports reward availability before conditioning on reward-bearing windows.

\begin{table}[!htbp]
\centering\small
\caption{Reward availability in the sampled behavior windows. Policy-pair entries are in thousands of environment steps.}
\label{tab:dmc-retention-main-frequency}
\setlength{\tabcolsep}{4pt}
\begin{adjustbox}{max width=\linewidth}
\begin{tabular}{@{}lcrr@{}}
\toprule
Task & $(\pi,\beta)$ & Fraction of reward-free windows & Fraction of steps with $r>0$ \\
\midrule
FH & $(200,100)$ & $0.808$ & $0.186$ \\
FH & $(500,400)$ & $0.503$ & $0.487$ \\
FH & $(500,100)$ & $0.809$ & $0.183$ \\
FH & $(500,500)$ & $0.411$ & $0.582$ \\
CS & $(200,100)$ & $0.851$ & $0.105$ \\
CS & $(500,400)$ & $0.749$ & $0.241$ \\
CS & $(500,100)$ & $0.863$ & $0.095$ \\
CS & $(500,500)$ & $0.400$ & $0.574$ \\
WR & $(200,100)$ & $0.000$ & $1.000$ \\
WR & $(500,400)$ & $0.000$ & $1.000$ \\
WR & $(500,100)$ & $0.000$ & $1.000$ \\
WR & $(500,500)$ & $0.000$ & $1.000$ \\
AS & $(200,100)$ & $0.000$ & $1.000$ \\
AS & $(500,400)$ & $0.000$ & $1.000$ \\
AS & $(500,100)$ & $0.000$ & $1.000$ \\
AS & $(500,500)$ & $0.000$ & $1.000$ \\
HH & $(200,100)$ & $0.969$ & $0.019$ \\
HH & $(500,400)$ & $0.747$ & $0.168$ \\
HH & $(500,100)$ & $0.977$ & $0.014$ \\
HH & $(500,500)$ & $0.753$ & $0.148$ \\
\bottomrule
\end{tabular}
\end{adjustbox}

\end{table}

\paragraph{Target moments.}
For each anchor $x=(s_0,a_0)$, let $\mu_m(x)$ and $v_m(x)$ be the conditional mean and variance of method $m$'s target over behavior windows and stopping draws. The reported squared bias averages $[\mu_m(x)-Q_{\rm MC}^\pi(x)]^2$ over anchors, rather than squaring an aggregated signed error. Variance averages $v_m(x)$, and target MSE combines these two quantities; finite-rollout reference error remains. These are learned-critic finite-backup diagnostics, not operator fixed-point bias. This distinction also separates the present squared-bias statistic from the horizontal coordinate of \cref{fig:dmc11-bv-main}.

For stochastic targets, with $D=32$ draws per window, the variance estimator accounting for within-window sampling is
\[
\widehat v=\widehat{\operatorname{Var}}_{w}(\bar Y_w)
+(1-D^{-1})\operatorname{Avg}_{w}\widehat{\operatorname{Var}}_{d}(Y_{wd}),
\]
where both sample variances use their corresponding independent sample counts. Fixed-length targets use the sample variance across windows. \Cref{tab:dmc-retention-main-absolute} gives \texttt{MINT}'s absolute values in unnormalized target units; values are seed means.

\begin{table}[!htbp]
\centering\small
\caption{\texttt{MINT} target statistics in the main measurement. Squared bias is computed per anchor before averaging.}
\label{tab:dmc-retention-main-absolute}
\setlength{\tabcolsep}{4pt}
\begin{adjustbox}{max width=\linewidth}
\begin{tabular}{@{}lcrrrrr@{}}
\toprule
Task & $(\pi,\beta)$ & Mean $K$ & $R_{\rm ret}$ & Bias$^2$ & Variance & MSE \\
\midrule
FH & $(200,100)$ & $7.08$ & $0.827$ & $1{,}179$ & $5.029$ & $1{,}184$ \\
FH & $(500,400)$ & $6.84$ & $0.754$ & $243.8$ & $1.166$ & $245.0$ \\
FH & $(500,100)$ & $7.97$ & $0.927$ & $300.9$ & $1.127$ & $302.0$ \\
FH & $(500,500)$ & $1.00$ & $0.159$ & $508.9$ & $0.065$ & $509.0$ \\
CS & $(200,100)$ & $3.66$ & $0.549$ & $83.51$ & $0.095$ & $83.61$ \\
CS & $(500,400)$ & $4.39$ & $0.677$ & $69.33$ & $0.112$ & $69.44$ \\
CS & $(500,100)$ & $5.96$ & $0.845$ & $50.03$ & $0.139$ & $50.17$ \\
CS & $(500,500)$ & $1.00$ & $0.149$ & $242.7$ & $0.011$ & $242.7$ \\
WR & $(200,100)$ & $8.75$ & $0.898$ & $2.354$ & $0.109$ & $2.464$ \\
WR & $(500,400)$ & $4.78$ & $0.560$ & $1.898$ & $0.049$ & $1.947$ \\
WR & $(500,100)$ & $9.37$ & $0.950$ & $2.336$ & $0.217$ & $2.553$ \\
WR & $(500,500)$ & $1.00$ & $0.159$ & $2.126$ & $0.004$ & $2.130$ \\
AS & $(200,100)$ & $4.98$ & $0.560$ & $14.56$ & $0.026$ & $14.59$ \\
AS & $(500,400)$ & $4.04$ & $0.483$ & $40.54$ & $0.042$ & $40.58$ \\
AS & $(500,100)$ & $4.52$ & $0.533$ & $23.76$ & $0.050$ & $23.81$ \\
AS & $(500,500)$ & $1.00$ & $0.186$ & $36.58$ & $0.001$ & $36.58$ \\
\bottomrule
\end{tabular}
\end{adjustbox}

\end{table}

\begin{table}[!htbp]
\centering\small
\caption{Reward retention: shuffled-$K$ minus \texttt{MINT}. Negative differences favor \texttt{MINT}; intervals are for paired seed differences.}
\label{tab:dmc-reward-retention}
\setlength{\tabcolsep}{4pt}
\begin{adjustbox}{max width=\linewidth}
\begin{tabular}{@{}lcrrrcr@{}}
\toprule
Task & $(\pi,\beta)$ & MINT & Shuffled & $\Delta R_{\rm ret}$ & $95\%$ interval & Valid seeds \\
\midrule
FH & $(200,100)$ & $0.827$ & $0.734$ & $-0.092$ & $[-0.136{,} -0.048]$ & $5$ \\
FH & $(500,400)$ & $0.754$ & $0.729$ & $-0.025$ & $[-0.071{,} 0.014]$ & $5$ \\
FH & $(500,100)$ & $0.927$ & $0.821$ & $-0.106$ & $[-0.180{,} -0.043]$ & $5$ \\
CS & $(200,100)$ & $0.549$ & $0.513$ & $-0.036$ & --- & $3$ \\
CS & $(500,400)$ & $0.677$ & $0.476$ & $-0.200$ & --- & $4$ \\
CS & $(500,100)$ & $0.845$ & $0.712$ & $-0.133$ & --- & $3$ \\
WR & $(200,100)$ & $0.898$ & $0.898$ & $0.000$ & $[0.000{,} 0.001]$ & $5$ \\
WR & $(500,400)$ & $0.560$ & $0.560$ & $0.000$ & $[0.000{,} 0.000]$ & $5$ \\
WR & $(500,100)$ & $0.950$ & $0.950$ & $0.000$ & $[0.000{,} 0.001]$ & $5$ \\
AS & $(200,100)$ & $0.560$ & $0.563$ & $0.003$ & $[-0.003{,} 0.011]$ & $5$ \\
AS & $(500,400)$ & $0.483$ & $0.483$ & $0.000$ & $[-0.006{,} 0.006]$ & $5$ \\
AS & $(500,100)$ & $0.533$ & $0.526$ & $-0.007$ & $[-0.017{,} 0.003]$ & $5$ \\
\bottomrule
\end{tabular}
\end{adjustbox}
\par\smallskip{\footnotesize Retention is defined only for reward-bearing windows. Dashes indicate fewer than five valid training seeds.}
\end{table}

\paragraph{Variance at a matched length distribution.}
The variance point estimates favor \texttt{MINT} over shuffled-$K$ in $11/12$ off-policy task--pair conditions: all three pairs on FH, CS, and AS, and two on WR. On FH, the reductions relative to shuffled-$K$ are approximately $6\%$, $45\%$, and $23\%$ for $(200,100)$, $(500,400)$, and $(500,100)$K, respectively. The corresponding reductions are approximately $64$--$68\%$ on CS and $65$--$86\%$ on AS; WR differences are small and mixed. These percentages are reconstructed from rounded summary values. Unlike the interpolated mean-length comparison in \cref{app:dmc11-bv}, this control preserves the full marginal stopping-length distribution. The results therefore support a descriptive variance benefit from the alignment between stopping length and trajectory in this measurement. The reported intervals concern MSE and retention, not variance differences, so the $11/12$ count is not a significance count.

FH also has higher retention than shuffled-$K$ in the two $\beta=100$K conditions, with differences $0.092$ and $0.106$ and intervals excluding zero. Its $(500,400)$K retention interval includes zero. CS retention differences favor \texttt{MINT}, but are defined in only three or four seeds. AS shows substantial variance reductions with little retention difference, so the variance pattern is not confined to reward-retention gains.

\paragraph{Fixed-length references and error decomposition.}
For completeness, \cref{tab:dmc-retention-main-shuffled,tab:dmc-retention-main-fixed10,tab:dmc-retention-main-fixed1} reproduce the target-MSE comparisons and their squared-bias and variance differences. Every difference is reference method minus \texttt{MINT}. The interval and direction columns apply only to $\Delta\mathrm{MSE}$. The variance point estimates do not establish dominance over fixed traces: \texttt{MINT} has lower variance than fixed-$10$ in only $1/12$ off-policy conditions, and higher variance than fixed-$1$ in all $12$. At the $\pi=\beta$ boundary it is approximately one-step and has lower variance than fixed-$10$ in all four tasks. Thus the matched-distribution benefit is distinct from a general variance advantage over long fixed backups. Comparisons with \cref{app:dmc11-bv} must also account for that diagnostic's averaging across current and older behavior checkpoints, seed exclusions, and standard-deviation aggregation.

\FloatBarrier
\begin{table}[H]
\centering\small
\caption{Target comparison: shuffled-$K$ minus \texttt{MINT}. All four tasks use five training seeds per policy pair.}
\label{tab:dmc-retention-main-shuffled}
\setlength{\tabcolsep}{4pt}
\begin{adjustbox}{max width=\linewidth}
\begin{tabular}{@{}lcrrrclrr@{}}
\toprule
Task & $(\pi,\beta)$ & MINT MSE & Ref. MSE & $\Delta$MSE & $95\%$ interval & MSE direction & $\Delta$bias$^2$ & $\Delta$var \\
\midrule
FH & $(200,100)$ & $1{,}184$ & $1{,}156$ & $-27.79$ & $[-39.93{,} -15.65]$ & Lower & $-28.09$ & $0.303$ \\
FH & $(500,400)$ & $245.0$ & $241.3$ & $-3.610$ & $[-6.668{,} -0.183]$ & Lower & $-4.557$ & $0.947$ \\
FH & $(500,100)$ & $302.0$ & $299.1$ & $-2.975$ & $[-4.898{,} -0.983]$ & Lower & $-3.314$ & $0.339$ \\
FH & $(500,500)$ & $509.0$ & $509.0$ & $0.000$ & $[0.000{,} 0.000]$ & Includes 0 & $0.000$ & $0.000$ \\
CS & $(200,100)$ & $83.61$ & $80.87$ & $-2.740$ & $[-6.975{,} -0.043]$ & Lower & $-2.906$ & $0.166$ \\
CS & $(500,400)$ & $69.44$ & $68.32$ & $-1.118$ & $[-3.318{,} 0.439]$ & Includes 0 & $-1.354$ & $0.236$ \\
CS & $(500,100)$ & $50.17$ & $46.16$ & $-4.010$ & $[-8.706{,} 0.298]$ & Includes 0 & $-4.298$ & $0.287$ \\
CS & $(500,500)$ & $242.7$ & $242.7$ & $0.000$ & $[0.000{,} 0.000]$ & Includes 0 & $0.000$ & $0.000$ \\
WR & $(200,100)$ & $2.464$ & $2.451$ & $-0.013$ & $[-0.025{,} -0.001]$ & Lower & $-0.015$ & $0.002$ \\
WR & $(500,400)$ & $1.947$ & $1.968$ & $0.021$ & $[0.004{,} 0.040]$ & Higher & $0.023$ & $-0.002$ \\
WR & $(500,100)$ & $2.553$ & $2.543$ & $-0.010$ & $[-0.019{,} -0.001]$ & Lower & $-0.011$ & $0.001$ \\
WR & $(500,500)$ & $2.130$ & $2.130$ & $0.000$ & $[0.000{,} 0.000]$ & Includes 0 & $0.000$ & $0.000$ \\
AS & $(200,100)$ & $14.59$ & $15.01$ & $0.428$ & $[-0.268{,} 1.427]$ & Includes 0 & $0.262$ & $0.166$ \\
AS & $(500,400)$ & $40.58$ & $42.43$ & $1.846$ & $[0.091{,} 4.027]$ & Higher & $1.702$ & $0.144$ \\
AS & $(500,100)$ & $23.81$ & $23.95$ & $0.138$ & $[-0.445{,} 0.637]$ & Includes 0 & $0.046$ & $0.092$ \\
AS & $(500,500)$ & $36.58$ & $36.58$ & $0.000$ & $[0.000{,} 0.000]$ & Includes 0 & $0.000$ & $0.000$ \\
\bottomrule
\end{tabular}
\end{adjustbox}
\par\smallskip{\footnotesize Lower/Higher means that the $\Delta$MSE interval lies wholly below/above zero. Variance differences are point estimates without separate intervals.}
\end{table}

\begin{table}[H]
\centering\small
\caption{Target comparison: fixed TD trace, $K=10$ minus \texttt{MINT}. All four tasks use five training seeds per policy pair.}
\label{tab:dmc-retention-main-fixed10}
\setlength{\tabcolsep}{4pt}
\begin{adjustbox}{max width=\linewidth}
\begin{tabular}{@{}lcrrrclrr@{}}
\toprule
Task & $(\pi,\beta)$ & MINT MSE & Ref. MSE & $\Delta$MSE & $95\%$ interval & MSE direction & $\Delta$bias$^2$ & $\Delta$var \\
\midrule
FH & $(200,100)$ & $1{,}184$ & $1{,}186$ & $1.407$ & $[-2.898{,} 5.636]$ & Includes 0 & $2.186$ & $-0.779$ \\
FH & $(500,400)$ & $245.0$ & $241.5$ & $-3.431$ & $[-6.460{,} -0.402]$ & Lower & $-3.221$ & $-0.210$ \\
FH & $(500,100)$ & $302.0$ & $300.3$ & $-1.718$ & $[-7.507{,} 2.053]$ & Includes 0 & $-1.637$ & $-0.081$ \\
FH & $(500,500)$ & $509.0$ & $452.8$ & $-56.20$ & $[-97.78{,} -25.70]$ & Lower & $-57.21$ & $1.006$ \\
CS & $(200,100)$ & $83.61$ & $82.56$ & $-1.042$ & $[-2.967{,} 0.461]$ & Includes 0 & $-0.994$ & $-0.048$ \\
CS & $(500,400)$ & $69.44$ & $62.41$ & $-7.035$ & $[-14.82{,} -0.267]$ & Lower & $-7.012$ & $-0.023$ \\
CS & $(500,100)$ & $50.17$ & $48.58$ & $-1.588$ & $[-3.247{,} -0.219]$ & Lower & $-1.497$ & $-0.092$ \\
CS & $(500,500)$ & $242.7$ & $232.5$ & $-10.24$ & $[-20.57{,} -0.478]$ & Lower & $-10.32$ & $0.084$ \\
WR & $(200,100)$ & $2.464$ & $2.414$ & $-0.050$ & $[-0.143{,} 0.003]$ & Includes 0 & $-0.047$ & $-0.003$ \\
WR & $(500,400)$ & $1.947$ & $1.738$ & $-0.209$ & $[-0.299{,} -0.121]$ & Lower & $-0.216$ & $0.006$ \\
WR & $(500,100)$ & $2.553$ & $2.549$ & $-0.005$ & $[-0.020{,} 0.006]$ & Includes 0 & $-0.004$ & $-0.001$ \\
WR & $(500,500)$ & $2.130$ & $1.882$ & $-0.248$ & $[-0.359{,} -0.135]$ & Lower & $-0.281$ & $0.033$ \\
AS & $(200,100)$ & $14.59$ & $14.20$ & $-0.382$ & $[-0.563{,} -0.213]$ & Lower & $-0.361$ & $-0.021$ \\
AS & $(500,400)$ & $40.58$ & $38.07$ & $-2.506$ & $[-4.557{,} -0.454]$ & Lower & $-2.468$ & $-0.037$ \\
AS & $(500,100)$ & $23.81$ & $22.24$ & $-1.572$ & $[-2.341{,} -0.720]$ & Lower & $-1.527$ & $-0.045$ \\
AS & $(500,500)$ & $36.58$ & $33.04$ & $-3.546$ & $[-4.783{,} -1.644]$ & Lower & $-3.551$ & $0.005$ \\
\bottomrule
\end{tabular}
\end{adjustbox}
\par\smallskip{\footnotesize Lower/Higher means that the $\Delta$MSE interval lies wholly below/above zero. Variance differences are point estimates without separate intervals.}
\end{table}

\begin{table}[H]
\centering\small
\caption{Target comparison: fixed TD trace, $K=1$ minus \texttt{MINT}. All four tasks use five training seeds per policy pair.}
\label{tab:dmc-retention-main-fixed1}
\setlength{\tabcolsep}{4pt}
\begin{adjustbox}{max width=\linewidth}
\begin{tabular}{@{}lcrrrclrr@{}}
\toprule
Task & $(\pi,\beta)$ & MINT MSE & Ref. MSE & $\Delta$MSE & $95\%$ interval & MSE direction & $\Delta$bias$^2$ & $\Delta$var \\
\midrule
FH & $(200,100)$ & $1{,}184$ & $1{,}081$ & $-102.8$ & $[-144.2{,} -59.96]$ & Lower & $-97.85$ & $-4.982$ \\
FH & $(500,400)$ & $245.0$ & $239.5$ & $-5.478$ & $[-16.16{,} 6.929]$ & Includes 0 & $-4.352$ & $-1.126$ \\
FH & $(500,100)$ & $302.0$ & $296.0$ & $-6.051$ & $[-18.90{,} 12.62]$ & Includes 0 & $-4.952$ & $-1.098$ \\
FH & $(500,500)$ & $509.0$ & $509.0$ & $0.000$ & $[0.000{,} 0.000]$ & Includes 0 & $0.000$ & $0.000$ \\
CS & $(200,100)$ & $83.61$ & $79.50$ & $-4.105$ & $[-13.50{,} 0.996]$ & Includes 0 & $-4.016$ & $-0.090$ \\
CS & $(500,400)$ & $69.44$ & $71.65$ & $2.213$ & $[-7.661{,} 12.98]$ & Includes 0 & $2.314$ & $-0.102$ \\
CS & $(500,100)$ & $50.17$ & $43.69$ & $-6.479$ & $[-15.78{,} 2.545]$ & Includes 0 & $-6.345$ & $-0.134$ \\
CS & $(500,500)$ & $242.7$ & $242.7$ & $0.000$ & $[0.000{,} 0.000]$ & Includes 0 & $0.000$ & $0.000$ \\
WR & $(200,100)$ & $2.464$ & $2.715$ & $0.252$ & $[-0.111{,} 0.611]$ & Includes 0 & $0.354$ & $-0.103$ \\
WR & $(500,400)$ & $1.947$ & $2.195$ & $0.248$ & $[0.136{,} 0.349]$ & Higher & $0.293$ & $-0.045$ \\
WR & $(500,100)$ & $2.553$ & $2.164$ & $-0.389$ & $[-0.975{,} 0.055]$ & Includes 0 & $-0.184$ & $-0.206$ \\
WR & $(500,500)$ & $2.130$ & $2.130$ & $0.000$ & $[0.000{,} 0.000]$ & Includes 0 & $0.000$ & $0.000$ \\
AS & $(200,100)$ & $14.59$ & $15.83$ & $1.244$ & $[-0.187{,} 3.549]$ & Includes 0 & $1.269$ & $-0.025$ \\
AS & $(500,400)$ & $40.58$ & $44.95$ & $4.371$ & $[1.309{,} 7.970]$ & Higher & $4.412$ & $-0.041$ \\
AS & $(500,100)$ & $23.81$ & $25.22$ & $1.408$ & $[0.793{,} 2.081]$ & Higher & $1.458$ & $-0.050$ \\
AS & $(500,500)$ & $36.58$ & $36.58$ & $0.000$ & $[0.000{,} 0.000]$ & Includes 0 & $0.000$ & $0.000$ \\
\bottomrule
\end{tabular}
\end{adjustbox}
\par\smallskip{\footnotesize Lower/Higher means that the $\Delta$MSE interval lies wholly below/above zero. Variance differences are point estimates without separate intervals.}
\end{table}

\begin{table}[H]
\centering\small
\caption{MSE interval directions over the 16 task--pair conditions. Entries count lower / includes zero / higher for reference minus \texttt{MINT}, using the within-window variance estimator defined above.}
\label{tab:dmc-retention-main-counts}
\setlength{\tabcolsep}{4pt}
\begin{adjustbox}{max width=\linewidth}
\begin{tabular}{@{}lc@{}}
\toprule
Reference & Lower / Includes zero / Higher \\
\midrule
Fixed TD trace, $K=1$ & $1 / 12 / 3$ \\
Fixed TD trace, $K=2$ & $4 / 9 / 3$ \\
Fixed TD trace, $K=3$ & $5 / 9 / 2$ \\
Fixed TD trace, $K=4$ & $5 / 10 / 1$ \\
Fixed TD trace, $K=5$ & $9 / 7 / 0$ \\
Fixed TD trace, $K=6$ & $12 / 4 / 0$ \\
Fixed TD trace, $K=7$ & $12 / 4 / 0$ \\
Fixed TD trace, $K=8$ & $11 / 5 / 0$ \\
Fixed TD trace, $K=9$ & $13 / 3 / 0$ \\
Fixed TD trace, $K=10$ & $11 / 5 / 0$ \\
Shuffled-$K$ & $6 / 8 / 2$ \\
Pure $10$-step & $9 / 6 / 1$ \\
Retrace & $5 / 8 / 3$ \\
\bottomrule
\end{tabular}
\end{adjustbox}

\end{table}

\paragraph{Interpretation and limits.}
The main result is higher reward retention on selected sparse-task pairs together with lower variance than a shuffled control at the same length distribution. This is a finite-checkpoint target result and does not identify the cause of the separately reported control-performance gains. Absolute target MSE is dominated by squared bias in these cells; neither it nor target variance alone measures the usefulness of an update for policy improvement. The measurements do not separate action-independent value offsets from action-dependent errors. No online shuffled-$K$ learning comparison is provided. At $\pi=\beta$, \texttt{MINT} and fixed-$1$ targets agree to within $1.7\times10^{-5}$ in the reported check; tiny stored-versus-recomputed log-probability differences can produce occasional non-one-step draws.
\FloatBarrier

\subsubsection{SAC Temperature and Retained Length}\label{app:dmc11-temperature}
All three DMC training methods use automatic entropy tuning. The median within-run Spearman correlations over 55 \texttt{MINT} runs are $-0.098$ for $\alpha$ versus $K$ and $-0.045$ for $\alpha$ versus logged keep probability; first-difference medians are $-0.028$ and $-0.031$. Opposite task-specific signs do not support a universal temperature-to-horizon map. These autocorrelated series are descriptive rather than independent-observation tests. Temperature participates in learning, but continuation depends on current and stored behavior policies; the association does not identify a causal mechanism.

\begin{figure}[!htbp]
\centering
\includegraphics[width=\textwidth]{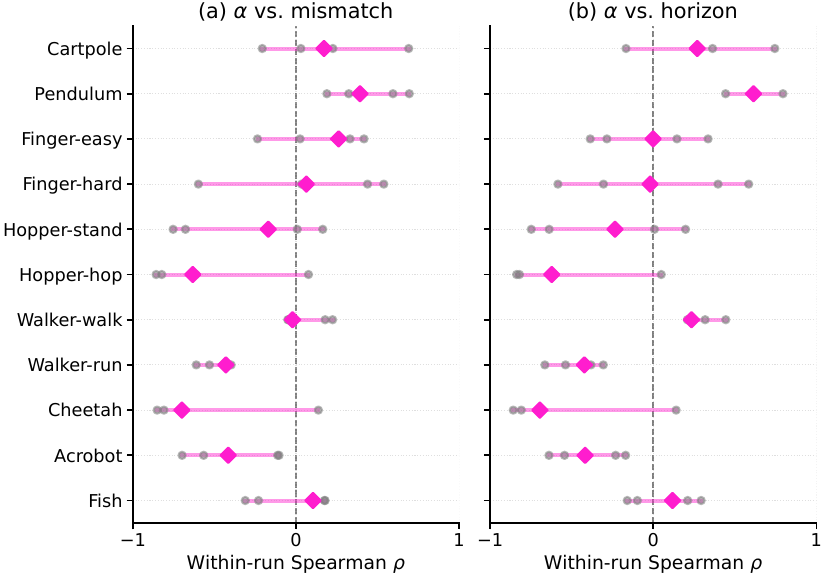}
\caption{\textbf{DMC temperature associations.} Each of 55 runs uses 50 training records. Gray points are seeds; pink diamonds are task medians and segments are seed min--max ranges, not confidence intervals. Weak overall median correlations do not imply independence.}
\label{fig:dmc11-temperature}
\end{figure}

\FloatBarrier
\subsubsection{All-Task and All-Stage Bias--Variance Results}\label{app:dmc11-full-planes}
\paragraph{Reading the training stages.}
At 100K, the only behavior checkpoint equals the target policy, so the stopping rule is one-step; this is a reference case rather than evidence that adaptation improves an off-policy target. Later stages include older behavior checkpoints. Comparing stages changes policies, critics, and behavior mixtures together, so movement in these planes is not a controlled time trend of one fixed operator. At 500K, \texttt{Finger-Hard}, \texttt{Cheetah}, and \texttt{Acrobot} retain visible cases where \texttt{MINT} does not improve both coordinates relative to the cap-ten TD trace. The planes complement the control returns rather than proving that a target-level coordinate change causes the training-score difference.
\begin{figure}[!htbp]
\centering
\includegraphics[width=\textwidth]{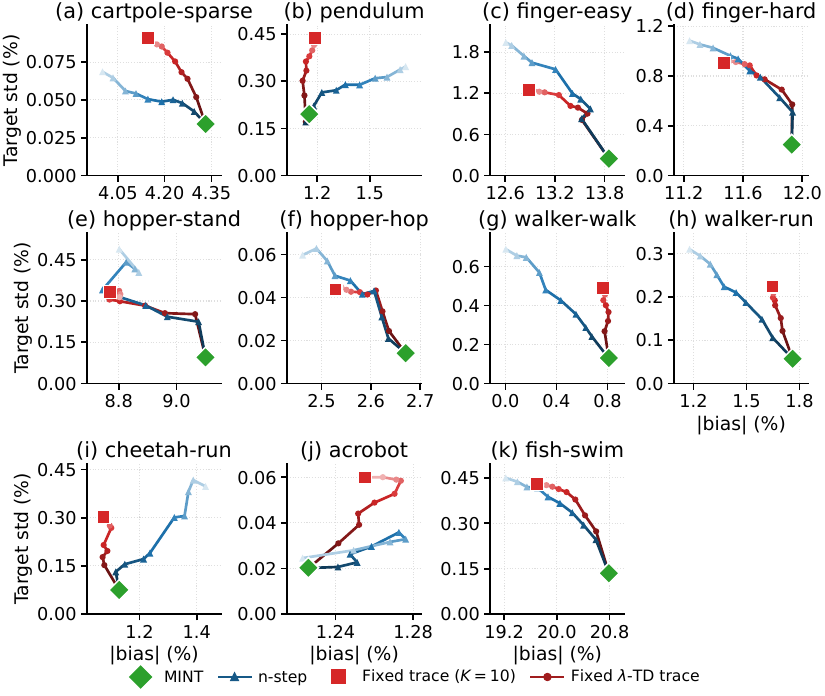}
\caption{\textbf{All-task DMC bias--variance plane at 100K.} Eleven task panels; 51/55 task--seed cells retained (\cref{tab:dmc11-bv-seeds}). Red: fixed-$K$ $\lambda$-weighted TD trace, square: $K=10$; blue: pure $n$-step; green diamond: \texttt{MINT}. Dark-to-light points mark length 1--10. Horizontal: absolute aggregated signed bias; vertical: target standard deviation. Both are percentages relative to the normalization value $100$, not a soft-value upper bound. Panel scales differ and no confidence intervals are shown. The available behavior and target checkpoints coincide. \texttt{MINT} is one-step by construction; overlapping marks can hide its diamond. The other fixed-length targets remain useful reference curves with learned critics.}
\label{fig:dmc11-bv-100k}
\end{figure}

\begin{figure}[!htbp]
\centering
\includegraphics[width=\textwidth]{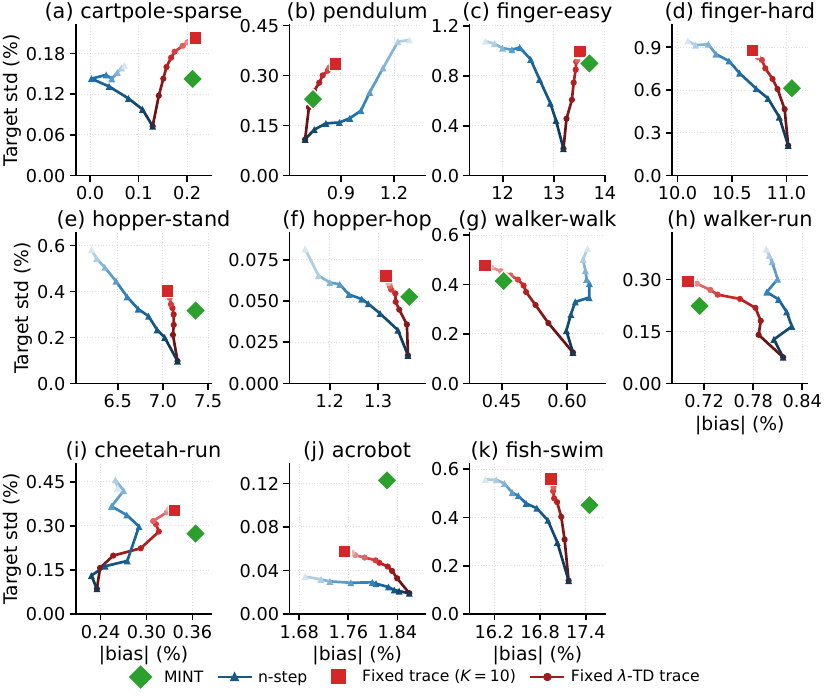}
\caption{\textbf{All-task DMC bias--variance plane at 200K.} Eleven task panels; 50/55 task--seed cells retained (\cref{tab:dmc11-bv-seeds}). Red: fixed-$K$ $\lambda$-weighted TD trace, square: $K=10$; blue: pure $n$-step; green diamond: \texttt{MINT}. Dark-to-light points mark length 1--10. Horizontal: absolute aggregated signed bias; vertical: target standard deviation. Both are percentages relative to the normalization value $100$, not a soft-value upper bound. Panel scales differ and no confidence intervals are shown. The first off-policy stage includes current and older behavior checkpoints. \texttt{Finger}/\texttt{Hopper} and \texttt{Walker} panels show different trade-offs; compare both coordinates within a panel rather than reading a single universal frontier.}
\label{fig:dmc11-bv-200k}
\end{figure}

\begin{figure}[!htbp]
\centering
\includegraphics[width=\textwidth]{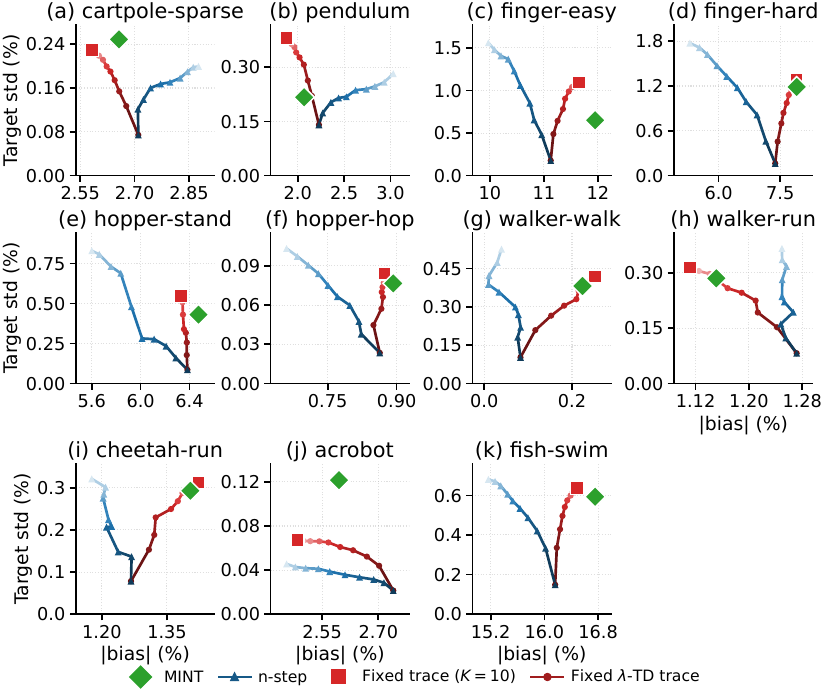}
\caption{\textbf{All-task DMC bias--variance plane at 300K.} Eleven task panels; 50/55 task--seed cells retained (\cref{tab:dmc11-bv-seeds}). Red: fixed-$K$ $\lambda$-weighted TD trace, square: $K=10$; blue: pure $n$-step; green diamond: \texttt{MINT}. Dark-to-light points mark length 1--10. Horizontal: absolute aggregated signed bias; vertical: target standard deviation. Both are percentages relative to the normalization value $100$, not a soft-value upper bound. Panel scales differ and no confidence intervals are shown. This figure includes all eleven tasks at 300K, including the \texttt{Acrobot} counterexample. \texttt{Finger-Easy}, \texttt{Walker-Walk}, \texttt{Fish}, and \texttt{Acrobot} show spread savings, modest changes, and a negative case under the same stage-specific aggregation.}
\label{fig:dmc11-bv-300k}
\end{figure}

\begin{figure}[!htbp]
\centering
\includegraphics[width=\textwidth]{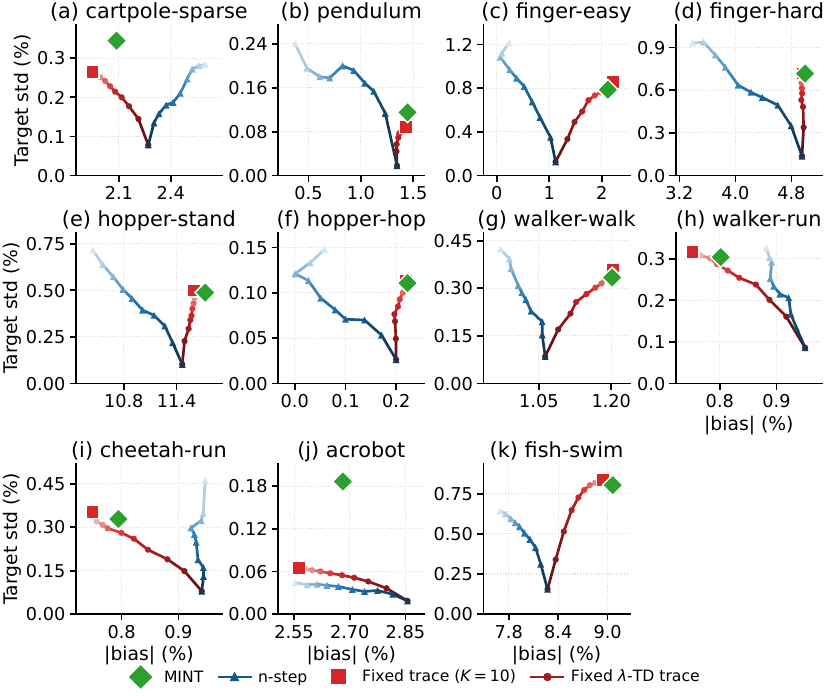}
\caption{\textbf{All-task DMC bias--variance plane at 500K.} Eleven task panels; 45/55 task--seed cells retained (\cref{tab:dmc11-bv-seeds}). Red: fixed-$K$ $\lambda$-weighted TD trace, square: $K=10$; blue: pure $n$-step; green diamond: \texttt{MINT}. Dark-to-light points mark length 1--10. Horizontal: absolute aggregated signed bias; vertical: target standard deviation. Both are percentages relative to the normalization value $100$, not a soft-value upper bound. Panel scales differ and no confidence intervals are shown. These panels use the 500K checkpoints. \texttt{Pendulum} retains one seed and \texttt{Finger-Hard} two; these panels do not characterize all five trained runs. Negative cases remain visible, so lower spread or bias is not asserted uniformly.}
\label{fig:dmc11-bv-500k}
\end{figure}

\FloatBarrier
\subsubsection{Additional CartPole Balance and Hopper Transition Diagnostics}
\label{app:dmc_sparse}\label{app:dmc-additional}

The \texttt{CartPole-Balance-Sparse} $\lambda$ sweep and \texttt{Hopper} transition-distance diagnostic use separate protocols from \cref{tab:dmc-protocols}. \Cref{fig:two_vertical_wrap} documents parameter sensitivity and normalized next-state distances; the latter is not a policy-TV estimate. One-step matching and empirical backup gains are examined separately in \cref{app:surrogate-tv,app:kappa-measured}.

\begin{figure}[!htbp]
    \centering
    \begin{subfigure}{0.90\textwidth}
        \centering
        \includegraphics[width=\linewidth]{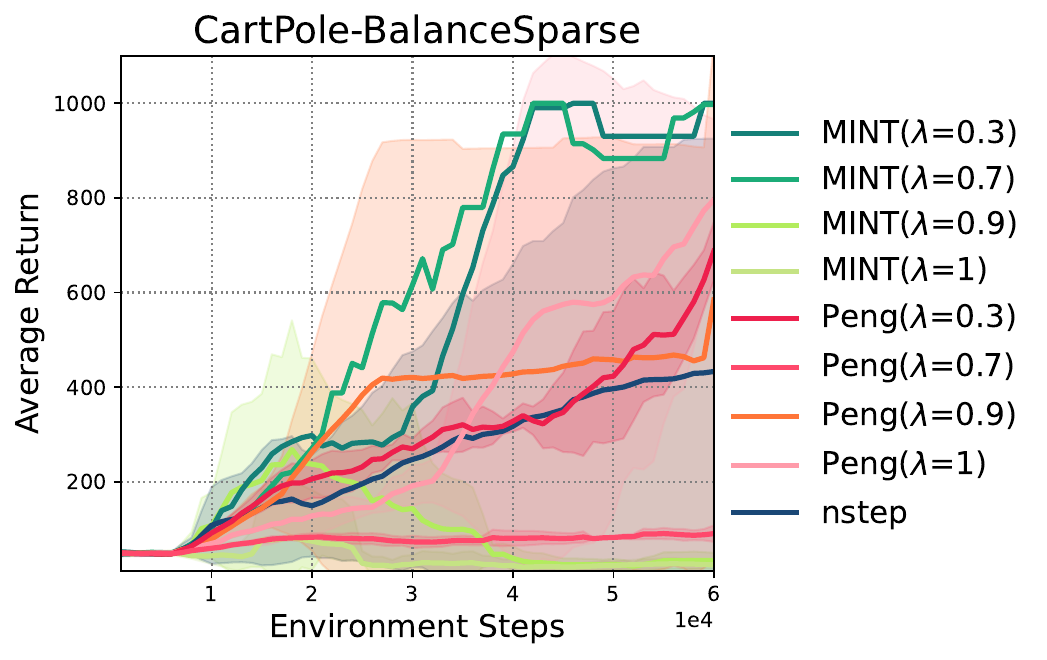}
        \caption{}
        \label{fig:cartpole_bal}
    \end{subfigure}
    \par\medskip
    \begin{subfigure}{0.75\textwidth}
        \centering
        \includegraphics[width=\linewidth]{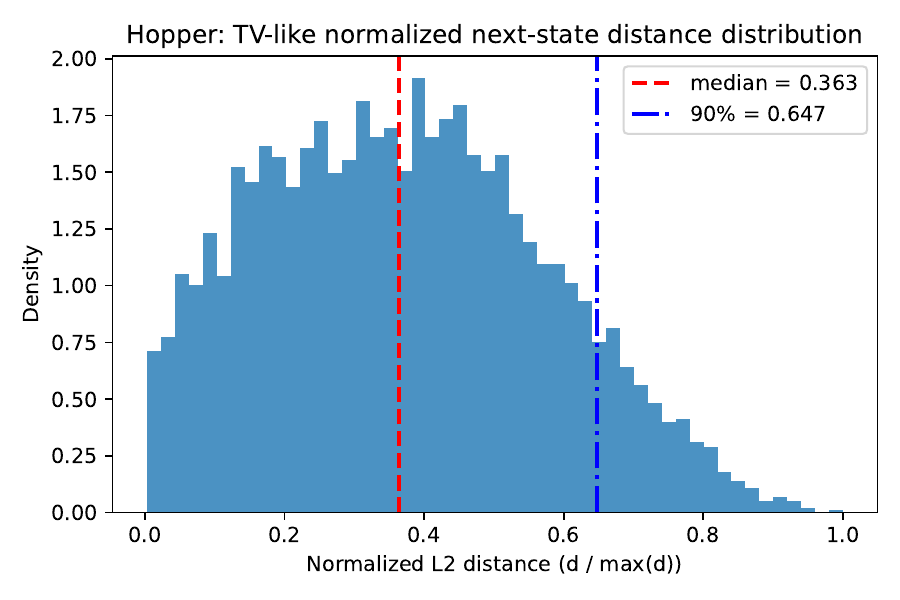}
        \caption{}
        \label{fig:mujoco_tv}
    \end{subfigure}
    \caption{\revr{\textbf{(a)} \texttt{CartPole-Balance-Sparse}: the original $\lambda\in\{0.3,0.7,0.9,1\}$ comparison. \texttt{MINT} succeeds at $\lambda=0.3$ and $0.7$; the poor-performing $0.9$ and $1$ settings are also retained. \textbf{(b)} Normalized next-state distances on \texttt{Hopper} (median ${\approx}0.36$, 90th percentile ${\approx}0.64$), used as descriptive dynamics diagnostics; they do not certify the surrogate contraction condition of \cref{thm:surrogate-contraction}.}}
    \label{fig:two_vertical_wrap}
\end{figure}

\FloatBarrier
\newpage
\subsubsection{Supplementary Five-Task DMC Evaluation}\label{app:dmc-five-task-evaluation}\label{sec:exp-ablation-trunc}
This supplementary evaluation uses the distinct conditions in \cref{tab:dmc-protocols}. Its original SAC-multistep label does not by itself identify the current pure ten-step baseline.

\paragraph{Setup.}
The tasks are \texttt{CartPole-Swingup-Sparse}, \texttt{Hopper-Stand}, \texttt{Finger-Turn-Easy}, \texttt{Acrobot-Swingup-Sparse}, and \texttt{Ball-in-Cup-Catch}. \Cref{tab:dmc-ablation} reports three-seed, 1M-step conditions, averaging the final 100K steps. A separate five-seed \texttt{CartPole-Swingup-Sparse} re-run at $L=10$ gives $617\pm113$; both its cap and seed set differ from the table's $L=5$ condition. The five-seed curves in \cref{fig:dmc-learning-curves} retain their original curve protocol.

\paragraph{Task-specific outcomes.}
\texttt{MINT} improves on the original multistep baseline on \texttt{CartPole} and \texttt{Hopper-Stand}, while \texttt{Finger-Turn-Easy} favors that baseline. \texttt{Acrobot-Swingup-Sparse} remains unsolved and \texttt{Ball-in-Cup-Catch} saturates for both reported methods. These comparisons do not isolate a bias--variance mechanism or imply that backup adaptation solves reward discovery.

\begin{table}[!htbp]
\centering
\caption{\textbf{DMControl ablation.} Mean evaluation return at 1M environment steps (mean$_{10}$ $\pm$ std, 3 seeds). Bold marks the best per row. SAC-\texttt{MINT} improves over uncorrected multi-step on CartPole and Hopper; both learn on Finger, saturate on Ball-in-Cup, and fail on the tested Acrobot setting.}
\label{tab:dmc-ablation}
\small
\begin{tabular}{lccc}
\toprule
\textbf{Task} & \textbf{SAC-\texttt{MINT}} & \textbf{SAC-multistep} & \textbf{Vanilla SAC} \\
\midrule
\multicolumn{4}{l}{\emph{Discriminative}} \\
CartPole-Swingup-Sparse & $\mathbf{719 \pm 46}$ & $24$           & $9 \pm 16$ \\
Hopper-Stand            & $\mathbf{891 \pm 24}$ & $57 \pm 90$    & --- \\
Finger-Turn-Easy        & $686 \pm 26$          & $\mathbf{767 \pm 62}$ & --- \\
\midrule
\multicolumn{4}{l}{\emph{Scope (non-discriminative)}} \\
Ball-in-Cup-Catch       & $976 \pm 2$           & ---            & $976 \pm 1$ \\
Acrobot-Swingup-Sparse  & $\le 1$               & ---            & --- \\
\bottomrule
\end{tabular}
\end{table}

\begin{figure}[!htbp]
\centering
\includegraphics[width=\textwidth]{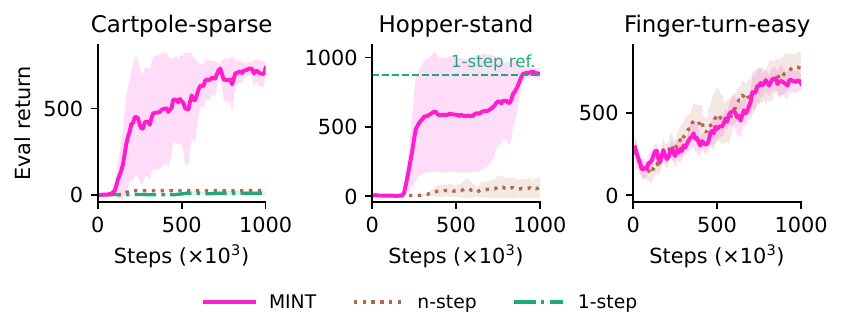}
\caption{\textbf{Supplementary DMC learning curves.} Five seeds, $95\%$ bootstrap intervals, 1M steps. The \texttt{Hopper-Stand} one-step reference is its final return. These curves retain their original settings, separately from the three-seed table and the \texttt{CartPole} re-run (\cref{tab:dmc-protocols,tab:dmc-ablation}).}
\label{fig:dmc-learning-curves}
\end{figure}

\FloatBarrier
\subsubsection{Fixed and Automatic Temperature on CartPole-Swingup-Sparse}\label{app:dmc-alpha-comparison}
The \texttt{CartPole-Swingup-Sparse} comparison uses cap $L=10$ (\cref{fig:alpha-trunc-diag}). Fixed $\alpha=0.2$ is reported with a diffuse policy before reward, $\hat K\approx2$, and failed learning. Under automatic tuning, reward is observed, $\alpha$ later settles at $0.06$--$0.10$, and $\hat K$ approaches $5.7$ with successful learning. Entropy alone does not determine the target--behavior discrepancy or stopping law. The supplementary evaluation also reports this regime on \texttt{Hopper-Stand} and early $\alpha$ collapse before reward on sparse \texttt{Acrobot}; these observations do not isolate the cause of that failure.

\begin{figure}[!htbp]
\centering
\includegraphics[width=0.85\textwidth]{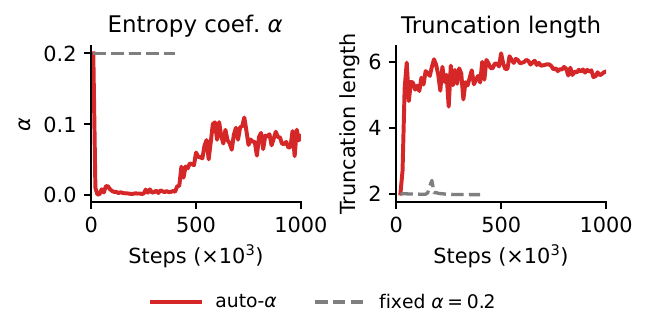}
\caption{\textbf{Fixed/auto-temperature comparison on \texttt{CartPole-Swingup-Sparse}.} Cap $L=10$; entropy coefficient (left) and realized length (right). The original plotted ranges are preserved. This protocol comparison is distinct from the within-run correlations in \cref{app:dmc11-temperature}.}
\label{fig:alpha-trunc-diag}
\end{figure}
\FloatBarrier

\subsection{Atari~100k: Primary Results and Distributional Limits}
\label{app:atari100k}\label{sec:exp-atari}

\paragraph{Scope and protocol.}
The primary result uses the DrQ($\varepsilon$)-based \texttt{MINT} configuration with $\lambda=0.9$, cap length $10$, and five seeds on each of 26 games (130 runs). Only $\lambda=0.7$ is retained as an additional ablation (seeds 5--9, versus primary seeds 0--4). Published DrQ($\varepsilon$) 100-run references are used. The full-suite aggregate is shown in the main text (\cref{fig:atari-v3-aggregate}); the numerical table below supports that comparison. Primary \texttt{MINT} uses hard target updates every 100 gradient steps and target clipping to $[-100,100]$. These changes accompany truncation and $\lambda$ weighting, so their effects are not isolated. The Atari target removes intermediate $V-Q$ corrections and changes target updates and clipping; it is a modified truncated $\lambda$-return, so the original TD-error operator's fixed-point guarantee is not asserted for it.

\paragraph{TD-error targets and target-update sensitivity.}
Direct application of the correction-preserving TD-error target learned poorly on \texttt{Breakout}, \texttt{Qbert}, and \texttt{Seaquest}. Frequent target copying was accompanied by large critic values; increasing the copy period to $100$ gradient steps did not resolve every tested setting. This motivated the stabilized variant specified above. We suspect that frequent target updates in this DrQ backbone amplify feedback through the TD($\lambda$) target, with residual sensitivity even at period $100$. This is a hypothesis about the backbone--target interaction: the diagnostics do not distinguish overfitting from other bootstrapping instability, and changing the copy period does not uniformly improve the unmodified DrQ backbone. An isolated test would vary the period while holding the return, clipping, and update budget fixed, and compare training-replay errors with held-out episode and Monte-Carlo errors.

\subsubsection{Full-suite aggregate and distributional performance}\label{app:atari-performance-evidence}
\Cref{tab:atari-context,tab:atari-context-ci} provide broader benchmark context \citep{agarwal2021deep,schwarzer2023bbf}. \texttt{MINT} exceeds DER's IQM point reference, but is below the SPR, EfficientZero, and BBF point references in full-suite IQM. Its DrQ-relative strengths must therefore be read as task-dependent benefits rather than a claim of leading Atari~100k performance. Runs, algorithms, and evaluation details differ; these references are not matched target ablations. The primary DrQ comparison elsewhere uses the 100-run table. The 10-run figure-derived references below are kept separate and approximate; their interval endpoints are not recomputed.

\begin{table}[!htbp]\centering\small
\caption{\textbf{Atari 26-game aggregate point estimates.} \texttt{MINT}: primary, five seeds/game. Literature rows use published point estimates \citep{agarwal2021deep,schwarzer2023bbf}; the BBF row aggregates 50 runs/game from the authors' released scores. Bottom rows are approximate estimates recovered from the rliable paper's 10-run figure. The 10-run and 100-run DrQ references are reported separately. Higher is better except optimality gap.}
\label{tab:atari-context}\label{tab:atari-v3-aggregate}
\begin{tabular}{lrrrrr}\toprule
Method & Runs/game & Median & IQM & Mean & Gap\\\midrule
\texttt{MINT} ($\lambda=0.9$) & 5 & $0.152$ & $0.251$ & $0.512$ & $0.627$\\
DER & 100 & $0.189$ & $0.183$ & $0.350$ & $0.698$\\
DrQ($\varepsilon$) & 100 & $0.313$ & $0.280$ & $0.465$ & $0.631$\\
SPR & 100 & $0.396$ & $0.337$ & $0.616$ & $0.577$\\
EfficientZero & 3 & $1.116$ & $1.020$ & $1.945$ & $0.371$\\
BBF (author scores) & 50 & $0.883$ & $1.034$ & $2.232$ & $0.339$\\
\midrule
DER (figure, approx.) & 10 & $0.210$ & $0.184$ & $0.348$ & $0.697$\\
DrQ (figure, approx.) & 10 & $0.331$ & $0.302$ & $0.505$ & $0.613$\\
SPR (figure, approx.) & 10 & $0.415$ & $0.375$ & $0.701$ & $0.554$\\
\bottomrule\end{tabular}\end{table}
\begin{table}[!htbp]\centering\small
\caption{\textbf{Full-suite estimates and uncertainty.} Point estimate [95\% stratified-bootstrap interval] from the archived aggregate results. Entries use numerical interval endpoints, which may differ slightly from those in the figures. BBF is published context, not a matched comparison.}
\label{tab:atari-context-ci}
\begin{tabular}{lcc}\toprule
Metric & \texttt{MINT} (5 runs/game) & BBF (50 runs/game)\\\midrule
Median & $0.152\;[0.120,0.180]$ & $0.883\;[0.825,0.942]$\\
IQM & $0.251\;[0.225,0.279]$ & $1.034\;[0.999,1.070]$\\
Mean & $0.512\;[0.470,0.554]$ & $2.232\;[2.183,2.281]$\\
Optimality Gap & $0.627\;[0.609,0.646]$ & $0.339\;[0.331,0.347]$\\
\bottomrule\end{tabular}\end{table}

\paragraph{Run-level coverage and comparison scope.}
\Cref{fig:atari-run-profile} shows the pooled game/run distribution. At HNS $0.1$, the reported fractions are \texttt{MINT} $0.585$, approximate 10-run DrQ $0.692$, and BBF $0.900$; at HNS $1.0$ they are $0.192$, $0.135$, and $0.445$. These are run fractions, whereas \cref{fig:atari-v3-profile} counts game means. The lower-threshold deficit and stronger upper tail against DrQ are consistent across these two views, without distributional dominance. The reported probability of improvement against BBF is $0.0544\,[0.0395,0.0692]$ (26 games); no DrQ probability is inferred without baseline run data. Neither the low BBF probability nor a positive DrQ group mean is a causal test of truncation.

\begin{figure}[!htbp]
\centering
\includegraphics[width=\textwidth]{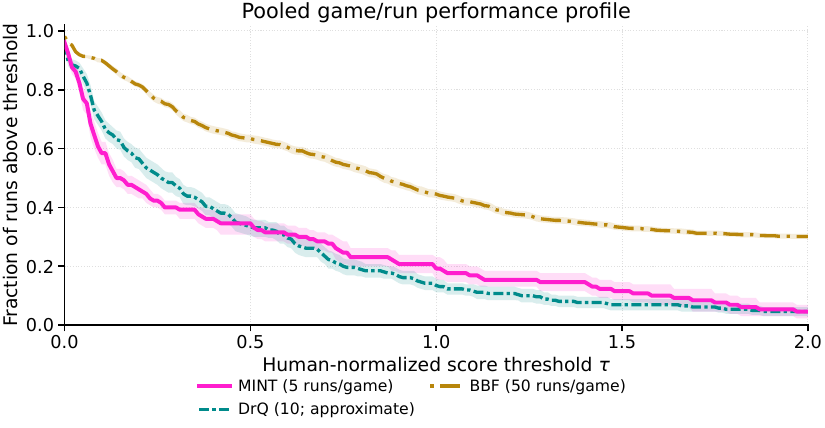}
\caption{\textbf{Pooled-run Atari performance profile.} Fraction of game/run scores above HNS threshold, with pointwise 95\% intervals. MINT: primary, five runs/game; BBF: 50 author runs/game. DrQ is the approximate 10-run profile recovered from the rliable paper, not the 100-run game-mean reference used in the group analysis. Lines connect the 201 evaluated thresholds without smoothing or additional interpolation; bands report the archived pointwise intervals. This differs from the game-mean profile in \cref{fig:atari-v3-profile}.}
\label{fig:atari-run-profile}
\end{figure}
\paragraph{Evaluation-time performance.}
\Cref{tab:atari-checkpoints} supplements the IQM curve with all four aggregate metrics. The nominal 10K--90K checkpoints use five evaluation episodes, whereas the final checkpoint uses 100. These checkpoints are evaluations of \texttt{MINT}, not a matched DrQ learning curve or a wall-clock comparison. They should not be read as a test of uniform improvement between adjacent checkpoints.
\begin{table}[!htbp]\centering\small
\caption{\textbf{Primary \texttt{MINT} evaluation checkpoints.} Point estimate [95\% interval]; 26 games and five training seeds. The larger final evaluation budget differs from the intermediate budget.}
\label{tab:atari-checkpoints}
\begin{tabular}{rcccc}\toprule
Nominal steps & Median & IQM & Mean & Gap\\\midrule
10K & \shortstack{$0.008$\\$[0.001,0.011]$} & \shortstack{$0.002$\\$[-0.001,0.006]$} & \shortstack{$0.020$\\$[-0.020,0.061]$} & \shortstack{$0.989$\\$[0.954,1.025]$}\\
50K & \shortstack{$0.068$\\$[0.063,0.087]$} & \shortstack{$0.108$\\$[0.096,0.123]$} & \shortstack{$0.271$\\$[0.246,0.296]$} & \shortstack{$0.770$\\$[0.753,0.788]$}\\
100K & \shortstack{$0.152$\\$[0.120,0.179]$} & \shortstack{$0.251$\\$[0.224,0.279]$} & \shortstack{$0.512$\\$[0.469,0.554]$} & \shortstack{$0.627$\\$[0.608,0.645]$}\\
\bottomrule\end{tabular}\end{table}

\FloatBarrier
\subsubsection{Task-group performance}\label{app:atari-groups}
\paragraph{Group benefits and concentration.}
We apply the rough exploration taxonomy of \citet[Appendix C.2]{bellemare2016unifying} to the 26 games: easy/human-optimal (11), easy/score-exploit (6), hard/dense (7), and hard/sparse (2). Group mean HNS is \texttt{MINT}/DrQ $0.689/0.463$, $0.741/0.875$, $0.077/0.151$, and $0.372/0.343$, respectively. These are comparisons against published game means, without baseline-run uncertainty. In that taxonomy, easy/hard concerns whether local exploration can find a high-scoring policy; the hard class is then divided by reward density. Reward density and exploration difficulty are different axes, and these historical labels do not certify coverage in a 100K run. Human-optimal is a category name, not a statement that this agent reaches human score; score-exploit does not mean that \texttt{MINT} learned an exploit. The sparse group contains only two games and \texttt{PrivateEye} remains near random.

\begin{figure}[!htbp]
\centering
\includegraphics[width=\textwidth]{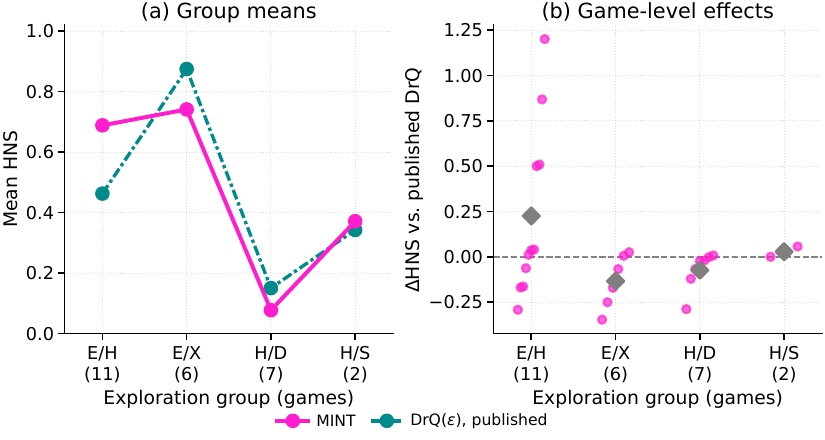}
\caption{\textbf{Atari task groups under Bellemare et al.\ \citep{bellemare2016unifying}.} E/H: easy/human-optimal; E/X: easy/score-exploit; H/D: hard/dense; H/S: hard/sparse (11/6/7/2 games). (a) Equal-game mean HNS. (b) Game-mean differences and group means against published DrQ($\varepsilon$). No baseline-run confidence intervals or subgroup pooled-run IQMs are inferred.}
\label{fig:atari-v3-groups}
\end{figure}

\paragraph{Group performance beyond the mean.}
\Cref{tab:atari-group-metrics,fig:atari-group-metrics} add median and pooled-run IQM to the mean-based view. Easy/human-optimal has \texttt{MINT}/DrQ mean $0.689/0.463$ and median of game means $0.669/0.429$, with $7/11$ game means higher. Thus the positive point-reference comparison is visible in the group's median as well as its mean. The \texttt{MINT} pooled-run IQM is $0.521$, but DrQ's group IQM cannot be recovered from its game means. The remaining seven games after removing the four largest gains still have a negative average difference: group central improvement and concentrated contribution are distinct observations.

Easy/score-exploit is weaker in both mean ($0.741/0.875$) and median ($0.398/0.423$), with $2/6$ higher game means. Hard/dense is lower in both ($0.077/0.151$ mean; $0.075/0.124$ median), with only $1/7$ higher. Hard/sparse has mean and median $0.372/0.343$, but consists of just \texttt{Freeway} and \texttt{PrivateEye}; seed uncertainty does not measure sensitivity to choosing these two tasks. These are descriptive group effects against published references, not evidence of significance or a learned exploration mechanism.

\begin{table}[!htbp]\centering\small
\caption{\textbf{Atari group performance.} Mean HNS and median across game means for both methods; IQM pools \texttt{MINT}'s game/seed scores only. DrQ group IQM is unavailable. Wins count \texttt{MINT} game means above the published DrQ reference, not a probability of improvement. DrQ values use rounded published returns.}
\label{tab:atari-group-metrics}
\setlength{\tabcolsep}{4pt}
\begin{tabular}{lrrrrrr}\toprule
& Wins & \multicolumn{2}{c}{Mean} & \multicolumn{2}{c}{Median} & IQM\\
Group & /games & \texttt{MINT} & DrQ & \texttt{MINT} & DrQ & \texttt{MINT}\\\midrule
E/H & 7/11 & $0.689$ & $0.463$ & $0.669$ & $0.429$ & $0.521$\\
E/X & 2/6 & $0.741$ & $0.875$ & $0.398$ & $0.423$ & $0.400$\\
H/D & 1/7 & $0.077$ & $0.151$ & $0.075$ & $0.124$ & $0.069$\\
H/S & 2/2 & $0.372$ & $0.343$ & $0.372$ & $0.343$ & $0.364$\\
\bottomrule\end{tabular}\end{table}

\begin{figure}[!htbp]
\centering
\includegraphics[width=\textwidth]{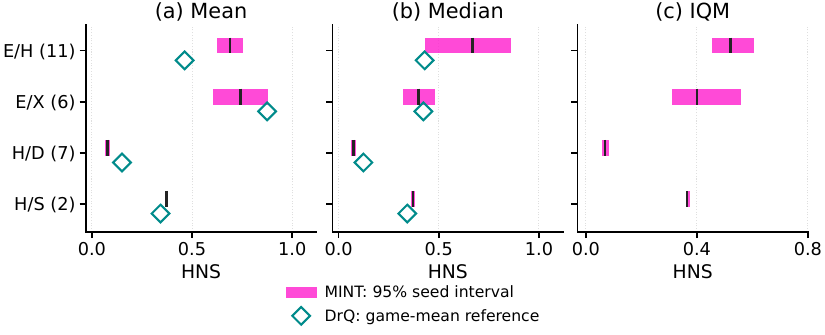}
\caption{\textbf{Group mean, median, and IQM with seed uncertainty.} E/H: easy/human-optimal; E/X: easy/score-exploit; H/D: hard/dense; H/S: hard/sparse. Pink rectangles are \texttt{MINT}'s 95\% percentile seed-bootstrap intervals (50,000 resamples, seeds independently resampled within each fixed game; no task resampling). Black vertical ticks are estimates; open diamonds are DrQ point references from published game means. Median is over game means, while IQM pools game/run scores. DrQ group IQM, baseline intervals, and difference intervals are unavailable. The sparse group's tight seed interval does not establish robustness over task selection.}
\label{fig:atari-group-metrics}
\end{figure}

\begin{table}[!htbp]\centering\small
\caption{\textbf{Where the Atari mean difference comes from.} Equal-game arithmetic decomposition against rounded published game means, without a confidence interval for the difference.}
\label{tab:atari-v3-contributions}
\begin{tabular}{lrr}\toprule
Group & Mean difference & Contribution to suite mean\\\midrule
Easy/human & $+0.225587$ & $+0.095440$\\
Easy/exploit & $-0.133670$ & $-0.030847$\\
Hard/dense & $-0.073158$ & $-0.019696$\\
Hard/sparse & $+0.028792$ & $+0.002215$\\\midrule
Total & & $+0.047112$\\\bottomrule
\end{tabular}\end{table}

The four largest-gain games are \texttt{DemonAttack}, \texttt{Boxing}, \texttt{Assault}, and \texttt{CrazyClimber}. Together they contribute $+0.118419$ to the full-suite mean difference; the other 22 contribute $-0.071307$. The remaining seven easy/human games have mean difference $-0.085$. Eighteen of the twenty \texttt{MINT} runs from those four games lie in the upper tail excluded by the full-suite IQM (32 lowest and 32 highest scores removed, 66 retained). Thus large game-specific gains raise mean HNS while leaving lower median and IQM point estimates. IQM describes central performance rather than failing to reflect an improvement. This does not establish a uniform distributional advantage or a general exploration improvement.

\begin{figure}[!htbp]
\centering
\includegraphics[width=\textwidth]{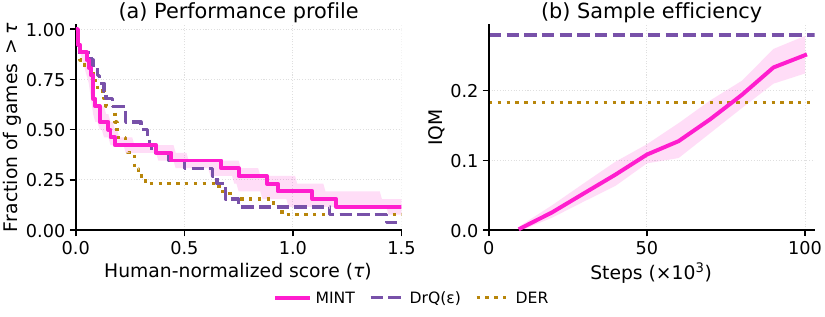}
\caption{\textbf{Atari game-mean profile and IQM progression.} Left: fraction of 26 game means above HNS threshold $\tau$; \texttt{MINT} averages five seeds/game and baselines use published game means. This is not a pooled-run profile. Right: \texttt{MINT} IQM by nominal training step (five evaluation episodes at 10K--90K, 100 at the endpoint); DrQ/DER horizontal lines are final-IQM references, not baseline learning curves. Bands show pointwise bootstrap intervals for \texttt{MINT}; final-reference comparisons do not establish a learning-speed advantage.}
\label{fig:atari-v3-profile}
\end{figure}

At HNS $0.1$, \texttt{MINT} exceeds the threshold in $16/26$ games versus $20/26$ for DrQ; at HNS $1.0$, the ordering reverses to $5/26$ versus $3/26$. This crossing profile is consistent with stronger upper-tail outcomes and weaker low-threshold coverage.

\FloatBarrier
\subsubsection{Interpreting group effects and realized horizons}\label{app:atari-backup-discussion}
\paragraph{Backup design under shared exploration.}
The variant retains the DrQ backbone's $\varepsilon$-greedy schedule and adds no directed-exploration objective. This makes backup design under standard exploration the relevant empirical question. Identical exploration schedules do not ensure identical state-visitation or replay distributions: changes in value estimates alter greedy actions, state visitation, and replay contents. Consequently, the present online comparison does not isolate learning from identical experience. Nor can a common exploration limitation explain the relative losses by itself, since the published DrQ reference scores higher on many of the same games.

\paragraph{Task structure, hypotheses, and counterexamples.}
Seven of eleven easy/human-optimal game means improve, but four games account for the large positive contribution and the other seven have a negative average difference. Six of seven hard/dense games have lower means; \texttt{MsPacman}'s small positive difference is the exception. Thus neither easy exploration nor dense reward alone determines benefit. We hypothesize that agreement-driven truncation can help when useful reward experiences are repeatedly available, by changing how observed rewards and bootstrapped values enter the target. Scores alone do not measure reward frequency, the fraction of rewards retained before stopping, or bootstrap accuracy, so this remains a mechanism hypothesis.

\texttt{Pong} is an easy/human-optimal counterexample: its mean HNS is lower by $0.169$ against the published game reference. \texttt{Qbert} is lower by $0.067$. The sparse group's small positive mean difference combines \texttt{Freeway}'s $+0.057$ with \texttt{PrivateEye}'s near-zero score; it does not show that exploration bottlenecks were solved. Game-level differences use published DrQ means, without baseline-run uncertainty. As emphasized by \citet{badia2020agent57}, finding rewards and assigning credit over long delays are separate challenges. Their Skiing/Solaris examples are absent from this 26-game suite; our group results do not establish an advantage on those long-delay tasks.

\paragraph{Reward retention versus bootstrapping.}
For a realized length $K$, shared state-value evaluations, and before clipping, the correction-free target is a truncated mixture of $n$-step returns. Changing $K$ and $\lambda$ changes both which observed rewards contribute and where value estimates are used. Shorter traces can reduce reliance on distant replay segments while increasing reliance on nearby bootstrap estimates; neither effect is guaranteed to help. Agreement with a greedy action does not certify that reward has been reached or that the bootstrap value is accurate. Target-update delay and clipping also change, preventing attribution of score differences to stopping alone. Direct tests would hold replay, target updates, clipping, and value evaluations fixed while comparing fixed-cap $\lambda$ mixtures and trajectory-independent matched-length or shuffled-length controls, together with reward-inclusion and bootstrap-error measurements. These measurements have not been made here.

\paragraph{Agreement and realized length.}
Mean $K$ falls from $6.20$ to $5.05$ while GreedyMatch rises from $14.1\%$ to $28.1\%$; final five-seed game means have rank correlation $-0.932$. Both short and long lengths include unsuccessful games. \textcolor{black}{GreedyMatch is the fraction of matching behavior and current greedy (argmax) actions, rather than the survival-conditioned agreement probability of the theory.} These are descriptive relationships, not an optimal-horizon certificate.

\begin{figure}[!htbp]
\centering
\includegraphics[width=0.90\textwidth]{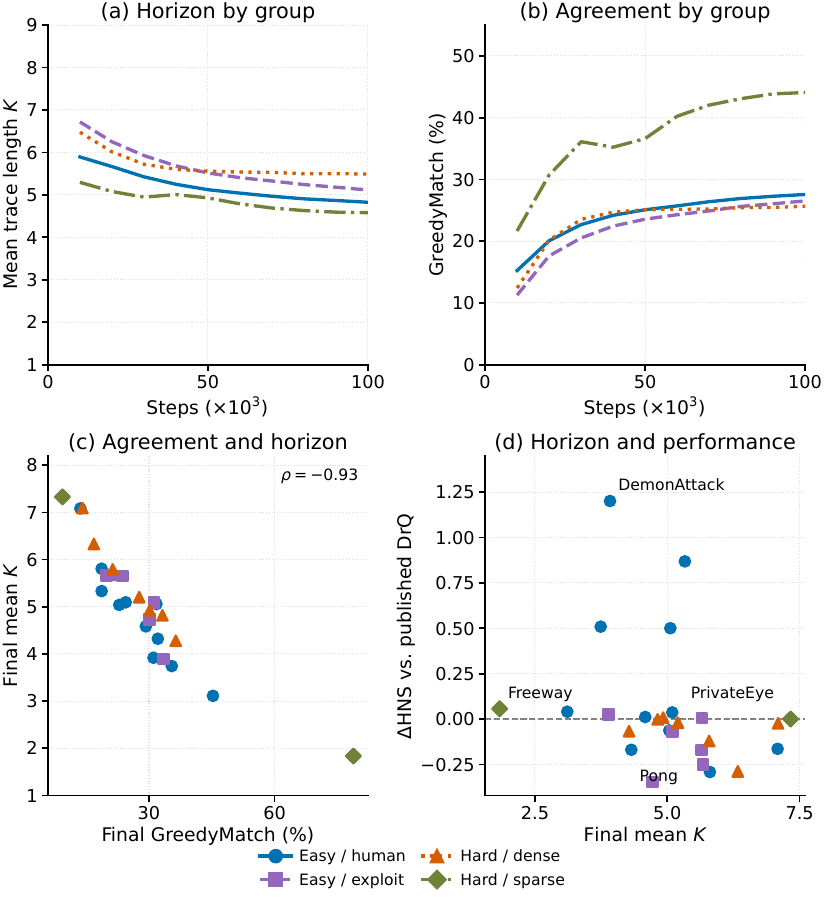}
\caption{\textbf{Primary Atari agreement and horizon diagnostics.} Group curves are equal-game averages of five-seed iteration means. Final scatter points are five-seed game means; correlations are descriptive across 26 games. Steps are nominal iteration counts times 10K; no smoothing is applied. Published DrQ game means are point references.}
\label{fig:atari-v3-horizons}
\end{figure}

The first-to-final nominal checkpoints are 10K and 100K interactions. Final means range from \texttt{Freeway}'s $1.83$ to \texttt{PrivateEye}'s $7.33$. \texttt{Assault} ($3.74$) and \texttt{DemonAttack} ($3.92$) improve with short traces, while \texttt{Pong} ($4.32$) and \texttt{Qbert} ($4.28$) lose; \texttt{BankHeist} ($6.33$) and \texttt{PrivateEye} retain long traces without a corresponding gain. Thus trace length alone does not predict benefit. The logged mean also does not determine the discounted trace mass $\mathbb E[\sum_{t=0}^{K-1}(\gamma\lambda)^t]$, which depends on the length distribution. It is neither a mixing-time estimate nor evidence of optimal credit assignment.

\FloatBarrier
\subsubsection{Additional \texorpdfstring{$\lambda=0.7$}{lambda=0.7} Ablation}\label{app:atari-lambda-ablation}
The additional variant has mean HNS $0.5098$ and IQM $0.2115$, versus primary $0.5118$ and $0.2509$. Its final mean $K$ is longer ($5.287$ versus $5.052$), while GreedyMatch is lower ($24.37\%$ versus $28.08\%$). Eleven game means rise and fifteen fall; probability of improvement is $0.4508\,[0.3846,0.5162]$. For fixed policies and a fixed trajectory, the fifth-power TD-weight ratio is $(0.7/0.9)^5\approx0.285$; this identity is not an explanation of the online score difference. The TD($\lambda$) target structure, target-update feedback, and learned policies interact, and the seed sets also differ. This comparison therefore does not isolate either the effect of realized horizon or the proposed target-update mechanism (\cref{app:atari100k}).

\begin{figure}[!htbp]
\centering
\includegraphics[width=\textwidth]{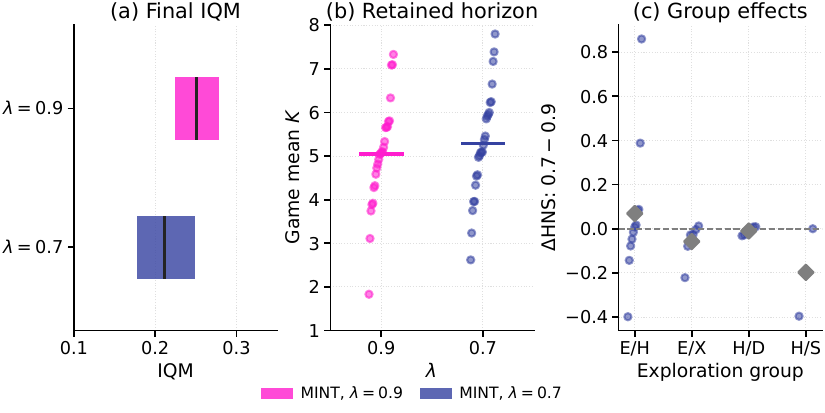}
\caption{\textbf{Additional Atari $\lambda$ ablation.} $\lambda=0.9$ (seeds 0--4) versus $0.7$ (seeds 5--9). IQM uses filled $95\%$ bootstrap confidence-interval bars and black estimate ticks. Other panels retain horizon and game-group HNS differences. This is supplementary to the primary $0.9$ result.}
\label{fig:atari-v3-lambda}
\end{figure}

\begin{table}[!htbp]\centering\small
\caption{\textbf{Primary Atari performance by game and task group.} \texttt{MINT} raw return: mean $\pm$ standard deviation over five seeds; DrQ return: published point reference. The final columns are human-normalized game means and their difference. Standard deviations are not confidence intervals; no uncertainty for the difference is inferred. Baseline protocols are not matched ablations. Values are rounded, including \texttt{PrivateEye}'s near-zero HNS.}
\label{tab:atari-games}
\setlength{\tabcolsep}{3pt}
\begin{tabular}{lrrrrr}\toprule
Game & \texttt{MINT} return & DrQ return & \texttt{MINT} HNS & DrQ HNS & $\Delta$HNS\\\midrule
\multicolumn{6}{l}{\textit{E/H: easy/human-optimal}}\\
\texttt{Assault} & $844.1\pm255.1$ & $579.6$ & $1.1965$ & $0.6875$ & $+0.5090$\\
\texttt{Asterix} & $859.3\pm119.1$ & $763.6$ & $0.0783$ & $0.0668$ & $+0.0115$\\
\texttt{BattleZone} & $4466.0\pm1460.7$ & $10165.3$ & $0.0605$ & $0.2241$ & $-0.1636$\\
\texttt{Boxing} & $19.4\pm2.8$ & $9.0$ & $1.6102$ & $0.7417$ & $+0.8685$\\
\texttt{Breakout} & $21.0\pm9.1$ & $19.8$ & $0.6691$ & $0.6285$ & $+0.0406$\\
\texttt{ChopperCommand} & $1086.8\pm110.1$ & $844.6$ & $0.0419$ & $0.0051$ & $+0.0368$\\
\texttt{CrazyClimber} & $34074.8\pm8583.3$ & $21539.0$ & $0.9300$ & $0.4295$ & $+0.5005$\\
\texttt{DemonAttack} & $3505.9\pm622.2$ & $1321.5$ & $1.8439$ & $0.6429$ & $+1.2009$\\
\texttt{Gopher} & $487.2\pm178.6$ & $621.6$ & $0.1065$ & $0.1689$ & $-0.0624$\\
\texttt{Jamesbond} & $269.3\pm104.8$ & $349.1$ & $0.8776$ & $1.1691$ & $-0.2915$\\
\texttt{Pong} & $-15.1\pm4.3$ & $-9.1$ & $0.1597$ & $0.3286$ & $-0.1690$\\
\addlinespace
\multicolumn{6}{l}{\textit{E/X: easy/score-exploit}}\\
\texttt{Kangaroo} & $579.2\pm116.4$ & $1088.4$ & $0.1767$ & $0.3474$ & $-0.1707$\\
\texttt{Krull} & $4135.0\pm791.6$ & $4402.1$ & $2.3766$ & $2.6268$ & $-0.2502$\\
\texttt{KungFuMaster} & $9955.4\pm1747.2$ & $11467.4$ & $0.4314$ & $0.4987$ & $-0.0673$\\
\texttt{RoadRunner} & $8501.8\pm5680.5$ & $11211.4$ & $1.0838$ & $1.4297$ & $-0.3459$\\
\texttt{Seaquest} & $612.3\pm165.9$ & $352.3$ & $0.0130$ & $0.0068$ & $+0.0062$\\
\texttt{UpNDown} & $4612.7\pm2105.7$ & $4324.5$ & $0.3655$ & $0.3397$ & $+0.0258$\\
\addlinespace
\multicolumn{6}{l}{\textit{H/D: hard/dense-reward}}\\
\texttt{Alien} & $729.4\pm96.0$ & $865.2$ & $0.0727$ & $0.0924$ & $-0.0197$\\
\texttt{Amidar} & $135.0\pm42.5$ & $137.8$ & $0.0754$ & $0.0770$ & $-0.0016$\\
\texttt{BankHeist} & $20.0\pm7.6$ & $232.9$ & $0.0078$ & $0.2960$ & $-0.2882$\\
\texttt{Frostbite} & $499.1\pm403.6$ & $1014.2$ & $0.1016$ & $0.2223$ & $-0.1206$\\
\texttt{Hero} & $3476.6\pm912.3$ & $4167.9$ & $0.0822$ & $0.1054$ & $-0.0232$\\
\texttt{MsPacman} & $1270.7\pm241.2$ & $1218.1$ & $0.1450$ & $0.1371$ & $+0.0079$\\
\texttt{Qbert} & $923.8\pm513.0$ & $1810.7$ & $0.0572$ & $0.1239$ & $-0.0667$\\
\addlinespace
\multicolumn{6}{l}{\textit{H/S: hard/sparse-reward}}\\
\texttt{Freeway} & $22.0\pm0.6$ & $20.3$ & $0.7430$ & $0.6858$ & $+0.0572$\\
\texttt{PrivateEye} & $28.2\pm43.9$ & $3.5$ & $0.0000$ & $-0.0003$ & $+0.0004$\\
\addlinespace
\bottomrule\end{tabular}\end{table}

\addtocontents{apc}{\protect\setcounter{tocdepth}{2}}
\clearpage
\begingroup\revcolor
\section{Cap, Signal, and Contraction in the Deep-RL Regime}\label{app:rwrl}

This appendix examines how the cap and disagreement signal affect the effective horizon (\cref{app:cap-signal}), compares the action-level surrogate with a one-step $\epsilon$-tolerant matching diagnostic (\cref{app:surrogate-tv}), and probes empirical backup gains under stochastic dynamics (\cref{app:kappa-measured}). The separate MuJoCo $\lambda$ sweep is reported in \cref{app:lambda-sweep}.

\paragraph{Common setup.}
We use RWRL \texttt{walker\_walk} \citep{dulacarnold2021challenges}, the Real-World RL Challenge variant of the DeepMind Control task, with a SAC backbone. RWRL provides a benchmark-defined actuator-noise channel: Gaussian noise of scale $\sigma\in\{0.0,0.1,0.3\}$ is added to the executed action inside the environment, so that $\mathcal{P}(s'\mid s,a)$ is non-degenerate at a specified level; we use this published mechanism rather than a perturbation of our own. \textcolor{black}{For the fixed-$\alpha$ 200K and auto-$\alpha$ 780K training arms in \cref{app:cap-signal}, the backbone follows the main table ($\lambda=0.7$, $\gamma=0.99$, \cref{tab:sac_hyperparams}), with caps $K_{\max}\in\{5,10,50\}$. The fixed-checkpoint stopping diagnostic uses a separate auto-$\alpha$ 1M protocol with $\lambda=0.9$ (\cref{app:rwrl-stopping-hist}).} The backup-gain probes use the separate protocol stated in \cref{app:kappa-measured}. The retained-horizon studies have two arms. The \emph{fixed-$\alpha$} arm uses $\alpha=0.05$, 3 seeds, and 200k environment steps. With $\alpha=0.2$, the main MuJoCo setting, the observed return on this DMControl-derived task is approximately $27$ after 80k steps. Its per-step reward is roughly $3.5\times$ smaller than \texttt{Hopper}'s, so $\alpha$ is rescaled to the reward scale. The eleven-task DMC comparison instead uses automatic entropy tuning (\cref{app:dmc11}). The \emph{auto-$\alpha$} arm learns $\alpha$ and runs 5 seeds per $(\sigma,K_{\max})$ cell (15 trials per cap) for 780k steps. These runs are a $K_{\mathrm{eff}}$ diagnostic on \texttt{MINT} itself---the only baseline included is uncorrected $n$-step at the same cap---not a re-run of the main comparison. All statistics are read at a common checkpoint for every trial.

\paragraph{\revr{What is and is not measured.}}
\revr{The diagnostics measure retained horizons and one-step matching, with the following scope. (i) We do not measure the multi-step state-coupling meeting time $T_{\beta,\pi}$ of \cref{def:meeting-time} on continuous control, nor a correlation between it and $\hat K$. \textcolor{black}{The measured state-level object is a calibrated one-step $\epsilon$-tolerant matching diagnostic, denoted $\widehat m_\epsilon(s)$ and compared with action-level TV in \cref{app:surrogate-tv}; it is not identified with exact kernel TV $d(s)$;} the retained-horizon and early-stopping evidence is summarized in the main text (\cref{tab:rwrl-fixed-alpha}). \textcolor{black}{(ii) We do not train an uncapped ($K_{\max}=\infty$) arm; the largest training cap is $50$. We separately measure fixed-checkpoint surrogate stopping without this cap, with right censoring at a verified rollout horizon of $1000$ (\cref{app:rwrl-stopping-hist}).} Neither substitute establishes the multi-step coupling guarantee for the deployed estimator; \textcolor{black}{they provide evidence about the selected horizon and an empirical association between action-level TV and the matching diagnostic.}}

\subsection{Cap-Length Sensitivity and Horizon Diagnostics}\label{app:cap-signal}\label{app:mujoco-cap}

The MuJoCo \texttt{Hopper} sweep in \cref{fig:hopper-cap} tests $K_{\max}\in\{3,5,10,20\}$ over 1M environment steps. \texttt{MINT} has limited return sensitivity across these caps, while Peng's $Q(\lambda)$ and uncorrected $n$-step vary more. The RWRL studies below examine both early stopping and return sensitivity under distinct entropy settings; the controlled toy-chain experiment separately compares capped and uncapped sampled backups (\cref{app:ablations}).

\Cref{tab:rwrl-fixed-alpha} in the main text reports the fixed-$\alpha$ retained horizons and early-stopping fractions. Here we give the corresponding normalized horizons and mean keep probabilities (\cref{tab:rwrl-fixed-alpha-diagnostics}), followed by the separate auto-$\alpha$ study.

\begin{table}[!htbp]
\centering\small
\caption{Additional diagnostics for the fixed-$\alpha$ RWRL comparison in \cref{tab:rwrl-fixed-alpha}: $\alpha=0.05$, 200k steps, three seeds, noise levels pooled.}
\label{tab:rwrl-fixed-alpha-diagnostics}
\begin{tabular}{rcc}
\toprule
$K_{\max}$ & $K_{\mathrm{eff}}/K_{\max}$ & $\E_\beta[\hat p]$ \\
\midrule
5 & $0.79$ & $0.85$ \\
50 & $0.10$ & $0.70$ \\
\bottomrule
\end{tabular}
\end{table}

\paragraph{Auto-$\alpha$ diagnostic: the stopping distribution needs room.}
The auto-$\alpha$ arm exhibits cap tracking (\cref{tab:rwrl-auto-alpha}). For $K_{\max}=50$, $K_{\mathrm{eff}}$ rises during warm-up, peaks near 100k steps, and declines to $25\%$ below its peak by 780k steps, with three quarters of traces stopping early. For $K_{\max}\in\{5,10\}$ it settles at ${\approx}97\%$ of the cap. The learned $\alpha$ converges to $0.002$--$0.014$ in every arm (last column), accompanied by a concentrated target policy, reported log-ratios near $-38$, and mean keep probabilities near $0.98$. Such small ratios make the clipped agreement weight negligible and continuation likely. At the same cap of $50$, the separate fixed-$\alpha$ arm has a lower mean keep probability ($0.70$ versus $0.98$) and a shorter effective horizon (approximately $5$ versus $21.9$). These observations associate entropy settings with surrogate survival under the tested protocols; they do not isolate temperature as the cause or measure a state-coupling meeting time. We pool over $\sigma$ because its effect on $K_{\mathrm{eff}}$ was smaller than the run-to-run spread, so we do not claim a trend with the noise level. A repeat of the 45-trial grid on different hardware with the same seeds reproduced every cell to within $0.1$ in $K_{\mathrm{eff}}$ and $0.6$ percentage points in the early-stopping fraction.

\begin{table}[!htbp]
\centering\small
\caption{Auto-$\alpha$ arm: $K_{\mathrm{eff}}$ and $\Pr[\hat K<K_{\max}]$ at 20k $\to$ 100k $\to$ 780k steps (15 trials per cap, $\sigma$ pooled), and the learned $\alpha$ at the end of training (5-seed mean).}
\label{tab:rwrl-auto-alpha}
\begin{tabular}{rccc}
\toprule
$K_{\max}$ & $K_{\mathrm{eff}}$ & $\Pr[\hat K<K_{\max}]$ & learned $\alpha$ \\
\midrule
5  & $4.34\to4.86\to4.85$ & $23.9\%\to5.5\%\to5.7\%$ & $0.0023$ \\
10 & $8.45\to9.49\to9.69$ & $27.2\%\to9.3\%\to5.9\%$ & $0.0082$ \\
50 & $23.98\to\mathbf{29.25}\to\mathbf{21.86}$ & $76.7\%\to61.8\%\to\mathbf{75.9\%}$ & $0.0058$--$0.0138$ \\
\bottomrule
\end{tabular}
\end{table}

\paragraph{Return sensitivity to cap length.}
Sweeping $K_{\max}$ quantifies return reductions at smaller caps (\cref{tab:rwrl-cap-cost}). Reducing $K_{\max}$ from 50 to 10 lowers $K_{\mathrm{eff}}$ by $56\%$ but final return by only $7\%$; reducing it from 10 to 5 halves $K_{\mathrm{eff}}$ again and lowers return by $60\%$. These results show that the cap can remain consequential for the implemented surrogate. The fixed-$\alpha$ arm shows the same direction at its two caps ($603\pm110$ at $K_{\max}=50$ versus $25\pm3$ at $K_{\max}=5$; 200k steps, 3 seeds), so this sensitivity also occurs without automatic entropy tuning. These training comparisons change the learned policy and its exploration/replay distribution together with the cap; they do not isolate a target-only effect of cap length. Fixed-temperature learning is particularly limited at cap $5$, while cap $50$ attains a substantial intermediate return at the shorter 200K budget. The fixed and automatic-temperature arms also differ in duration and seed count. Together with the \texttt{Hopper} sweep, they support using a cap large enough to accommodate the stopping regime, rather than selecting one exact horizon for every task. The common cap of $5$ in the main MuJoCo comparison remains a matched budget.

\revr{\Cref{tab:rwrl-sigma-return} breaks the $K_{\max}\in\{10,50\}$ returns down by noise level: the agent learns a strong policy at every $\sigma$ and performance is stable up to $\sigma=0.3$, with the gain from $K_{\max}=10$ to $50$ marginal, so the cap effect is not driven by any single noise level.}

\begin{table}[!htbp]
\centering\small
\caption{\revr{Eval return on RWRL \texttt{walker\_walk} by transition-noise level (auto-$\alpha$ arm, 780k steps, mean $\pm$ s.d.\ over 5 seeds; max return 1000).}}
\label{tab:rwrl-sigma-return}
{\color{black}
\begin{tabular}{lcc}
\toprule
$\sigma$ & $K_{\max}=10$ & $K_{\max}=50$ \\
\midrule
$0.0$ & $938.3\pm39.7$ & $963.2\pm6.0$ \\
$0.1$ & $878.1\pm104.7$ & $967.6\pm6.7$ \\
$0.3$ & $865.7\pm79.3$ & $963.7\pm3.5$ \\
pooled ($n=15$) & $894.0\pm80.4$ & $964.8\pm5.5$ \\
\bottomrule
\end{tabular}}
\end{table}

\begin{table}[!htbp]
\centering\small
\caption{Return as a function of the cap (auto-$\alpha$ arm, 780k steps, 15 trials per cell; the only experiment with a $K_{\max}=10$ condition).}
\label{tab:rwrl-cap-cost}
\begin{tabular}{rcccc}
\toprule
$K_{\max}$ & $K_{\mathrm{eff}}$ & $\Delta K_{\mathrm{eff}}$ & return & $\Delta$ return \\
\midrule
50 & $21.86$ & --- & $965$ & --- \\
10 & $9.69$ & $-56\%$ & $894$ & $\mathbf{-7\%}$ \\
5  & $4.85$ & $-50\%$ & $358$ & $\mathbf{-60\%}$ \\
\bottomrule
\end{tabular}
\end{table}

\begingroup\color{black}
\FloatBarrier
\subsubsection{Stopping Distributions across Behavior-Policy Ages}\label{app:rwrl-stopping-hist}\label{app:rwrl-age}
This fixed-checkpoint diagnostic uses RWRL \texttt{walker\_walk}, Gaussian actuator noise $\sigma\in\{0.1,0.3\}$, and no observation noise. Policies were trained with automatic entropy tuning, automatic target entropy, $\lambda=0.9$, and cap/window length $50$. We fix $\pi$ at 1M and separately evaluate $\beta$ at 900K, 500K, and 100K: behavior ages 100K, 500K, and 900K. The target checkpoint's $\alpha$ ranges from $0.004$ to $0.009$ across seeds. Each condition uses seeds $0$--$2$, $40$ states/seed sampled every $25$ steps along a behavior rollout, and $256$ indicator draws/state ($30{,}720$ draws). These draws share realized behavior trajectories; the independent training repeats are the three seeds. Different $\beta$ conditions also change visited states, so this is not an intervention on age alone at identical states.

At each realized action we sample the indicator in \cref{eq:practical-disagreement}, retain the anchor, and record $\hat T=\inf\{i\ge1:\hat A_i=0\}$ without imposing the training cap. The plotted $T$ is this action-level stopping time, not a state-coupling meeting time. Simulator state and episode clocks are restored for each measurement. Every age-sweep rollout records length $1000$, time-limit truncation, and no true terminal; draws still surviving at this horizon are right-censored. Only measurements with complete episode-clock restoration are included.

\begin{figure}[H]
\centering
\includegraphics[width=\textwidth]{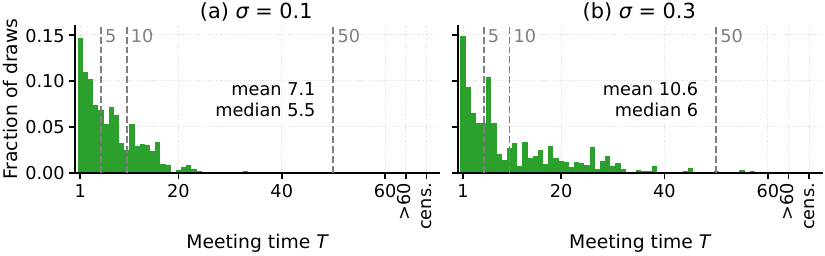}
\captionsetup{font={color=black},labelfont={color=black}}
\caption{\textbf{Recent-behavior stopping distribution.} Auto-$\alpha$, $\lambda=0.9$; $\pi$ at 1M, $\beta$ at 900K, training cap $50$. Panels pool three seeds, $40$ states/seed, and $256$ draws/state. Dashed caps $5/10/50$ are applied to the same stopping draws. Means are $7.06/10.56$, medians $5.5/6$, and maxima $56/75$ at $\sigma=0.1/0.3$; neither condition has censored draws. Horizontal ``meeting time'' $T$ denotes the surrogate stopping time $\hat T$.}
\label{fig:rwrl-stopping-hist}
\end{figure}

For recent $\beta$, exactly $2/30{,}720$ and $179/30{,}720$ draws exceed $50$ ($0.007\%/0.583\%$); another $0/55$ stop exactly at $50$. Thus $\Pr(\hat T\ge50)$ is $0.007\%/0.762\%$, distinct from strict cap cutting, $\Pr(\hat T>50)$. Cap $50$ therefore cuts fewer than $0.6\%$ of draws in both recent-policy conditions.

\paragraph{Older behavior policies and cap intervention.}
\Cref{fig:rwrl-age-survival,tab:rwrl-behavior-age} show longer tails at older $\beta$. All caps below are applied after measurement to fixed policies and shared draws; these are not separately trained cap arms. Writing $K_L=\min(L,\hat T)$, the integer-time identity
\[
 \mathbb E[K_{L+1}]-\mathbb E[K_L]=\Pr(\hat T>L)
\]
connects the survival curve directly to the marginal effect of increasing the cap.

\begin{figure}[H]
\centering
\includegraphics[width=\textwidth]{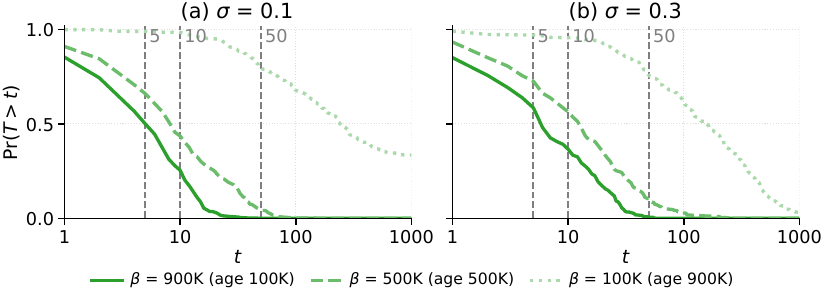}
\captionsetup{font={color=black},labelfont={color=black}}
\caption{\textbf{Surrogate stopping survival across behavior ages.} Fixed $\pi$ at 1M; solid/dashed/dotted curves use $\beta$ at 900K/500K/100K (ages 100K/500K/900K). Each noise--age condition pools three seeds and $30{,}720$ indicator draws. Dashed vertical lines mark caps $5/10/50$; the horizontal axis is logarithmic. Recent and intermediate conditions have no censoring. At age 900K, $33.5\%/3.0\%$ of draws remain censored at the observation horizon $1000$ for $\sigma=0.1/0.3$; their unobserved tails are not extrapolated.}
\label{fig:rwrl-age-survival}
\end{figure}

\begin{table}[H]
\centering\small\color{black}
\captionsetup{font={small,color=black},labelfont={color=black}}
\caption{\textbf{Behavior age and cap-$50$ intervention.} Pooled draw summaries, fixed $\pi$ at 1M. Values marked $\ge$ are lower bounds obtained by recording censored draws at $1000$. For $L\le100$, $\mathbb E[K_L]$ and $\Pr(\hat T>L)$ are fully observed empirical summaries despite tail censoring. Percentages are descriptive, without an independent-draw confidence claim.}
\label{tab:rwrl-behavior-age}
\setlength{\tabcolsep}{4pt}
\begin{tabular}{@{}ccrrrrr@{}}
\toprule
$\sigma$ & $\beta$ / age (K) & Mean $\hat T$ & Median & $\mathbb E[K_{50}]$ & \shortstack{Cut at 50\\(\%)} & \shortstack{Censored\\(\%)} \\
\midrule
$0.1$ & $900/100$ & $7.06$ & $5.5$ & $7.06$ & $0.007$ & $0$ \\
$0.1$ & $500/500$ & $14.95$ & $9$ & $14.21$ & $4.47$ & $0$ \\
$0.1$ & $100/900$ & $\ge447.2$ & $242$ & $45.96$ & $79.73$ & $33.5$ \\
\midrule
$0.3$ & $900/100$ & $10.56$ & $6$ & $10.52$ & $0.583$ & $0$ \\
$0.3$ & $500/500$ & $22.38$ & $13$ & $18.10$ & $10.38$ & $0$ \\
$0.3$ & $100/900$ & $\ge243.9$ & $159$ & $44.61$ & $75.70$ & $3.0$ \\
\bottomrule
\end{tabular}
\end{table}

Increasing the cap from $50$ to $100$ changes mean length by approximately $0/0.04$ at the reported precision in the recent conditions, $0.71/2.66$ at age 500K, and $37.10/34.42$ at age 900K. The oldest conditions' mean $\hat T$ remains only lower-bounded, while their cap-$50/100$ summaries are observed. Seed variation is substantial: at $\sigma=0.1$, age 900K, all seed-0 draws are censored, and the per-seed horizon-truncated means are $1000.0/167.1/174.6$. At age 500K, per-seed means are $20.9/8.9/15.0$ and $12.3/42.7/12.2$ for $\sigma=0.1/0.3$.

The same trials' 1M training logs report mean $\hat K=19.6$--$23.4$ and cap-reach rates $19$--$27\%$. Replay mixes collection-time behavior ages of approximately $0$--$990$K; none of the single-checkpoint conditions reconstructs that mixture, and the replay was not saved to verify mixture weights. The age sweep supports behavior-dependent cap margin, without proving that age alone explains the training gap or that a stopping distribution improves return or critic accuracy. It is separate from the fixed-$\alpha$ 200K comparison in \cref{tab:rwrl-fixed-alpha} and the $\lambda=0.7$, 780K training sweep above.
\endgroup

\begingroup\color{black}
\subsection{The Action-Level Surrogate and a One-Step Matching Diagnostic}\label{app:surrogate-tv}

\Cref{prop:dpi-stochastic} gives the exact one-step relation $\E_\beta[\hat p_i\mid s]=D_s\ge d(s)$, where $d(s)=\mathrm{TV}(\mathcal P^\beta(\cdot\mid s),\mathcal P^\pi(\cdot\mid s))$. RWRL's action-noise channel factorizes as $a\mapsto\tilde a=\mathrm{clip}(a+\varepsilon)\mapsto s'=F(s,\tilde a)$, with the same Markov maps for both policies, so the relation applies to the exact resulting distributions. Closeness of next-state coordinates alone does not establish TV overlap.

On the $K_{\max}=50$ auto-$\alpha$ arm (seed 0), action-level $D_s$ is evaluated by numerical quadrature of the two Gaussian policies. The simulator-based comparison uses a calibrated $\epsilon$-tolerant matching statistic $\widehat m_\epsilon(s)$, with a $\beta=\pi$ calibration control, at 250 replay states across four checkpoints. We report an $\epsilon$-tolerant matching diagnostic, rather than an estimate of exact kernel TV. Its observed ordering relative to action-level TV is an empirical diagnostic comparison, not a verification of DPI.

The measured ordering $\E_\beta[\hat p]\ge\widehat m_\epsilon(s)$ holds at all 250 retained states. The reported Spearman correlations are $0.95$ across the arm/checkpoint cells and $0.86$ across states (\cref{tab:surrogate-tv}), with level gaps $0.23$--$0.40$. The diagnostic excludes states whose calibration floor exceeds $2\%$, so this selection may affect the observed correlations and gaps. These are directional measurements on one task and one seed; they neither identify $d(s)$ nor order multi-step stopping times or certify the $\chi_\lambda$ contraction condition.
\endgroup

\begin{table}[!htbp]
\centering\small
\caption{\textcolor{black}{Action-level TV versus the $\epsilon$-tolerant matching diagnostic on RWRL \texttt{walker\_walk} ($K_{\max}=50$ arm, seed 0) at four checkpoints. $\E_\beta[\hat p]\ge\widehat m_\epsilon(s)$ at all 250 retained states; reported Spearman correlations are $0.95$ across cells and $0.86$ across states. The matching statistic is not an estimate of exact kernel TV.}}
\label{tab:surrogate-tv}
\begin{tabular}{lcc}
\toprule
checkpoint & $\E_\beta[\hat p]$ & \textcolor{black}{matching diagnostic $\widehat m_\epsilon(s)$} \\
\midrule
29\%  & $0.981$ & $0.755$ \\
57\%  & $0.874$ & $0.473$ \\
71\%  & $0.909$ & $0.655$ \\
100\% & $0.735$ & $0.365$ \\
\bottomrule
\end{tabular}
\end{table}

\subsection{\revr{Empirical Gain of the Deployed Operator}}\label{app:kappa-measured}

\revr{\textcolor{black}{The $\epsilon$-tolerant matching diagnostic does not certify the sufficient condition} of \cref{thm:surrogate-contraction}. We therefore probe the implemented backup empirically. Subtracting backups applied to two value functions cancels the reward and entropy terms and leaves a linear map on $\Delta=Q-Q\prime$; its measured gain over the visited distribution is}
\[
\hat\kappa \;=\; \frac{\lVert \hat{\mathcal{T}}Q-\hat{\mathcal{T}}Q'\rVert}{\lVert Q-Q'\rVert},
\]
\revr{which we estimate by fitted power iteration over the visited distribution. The probes use the action-level surrogate on windows of length $L=10$, with retained length $\hat K_{\mathrm{probe}}:=\min(\hat T_{\beta,\pi},K_{\max},L)$. Thus the nominal caps $5$ and $50$ give effective probe caps of $5$ and $10$, respectively. It is a fitted, visited-distribution diagnostic composed with critic projection. It is neither the mixed-norm state-level bound $\kappa^\star$ of \cref{cor:practical-expected} nor the global surrogate upper bound $\kappa_{\chi,L}$ of \cref{thm:surrogate-contraction}.}

\paragraph{Setup.}
The backup-gain probes use a separate auto-$\alpha$ protocol on RWRL \texttt{walker\_walk} with injected transition noise ($\sigma\in\{0.1,0.3\}$), 100k training steps, 3 seeds, and 60 probes across training checkpoints. We use this arm because $\hat p_i$ saturates there and many traces reach the effective probe cap. The policy pairs were trained at $(\lambda,\gamma)=(0.9,0.999)$. During the coefficient comparisons, these policy pairs and the sampled trajectories are held fixed while the backup coefficients are changed to the main-experiment values or the training values. The transition operator is evaluated on shared samples, so no density model of the kernel is needed.

\begin{table}[!htbp]
\centering\small
\caption{\revr{Empirical backup gain $\hat\kappa$ on RWRL \texttt{walker\_walk} with transition noise. \textbf{Top:} operating point (pooled over cap, noise, seed). \textbf{Middle:} deployed coefficients by nominal cap. \textbf{Bottom:} comparison at $\lambda=0$ with the one-step Bellman reference $\gamma$. Probes use windows of length $L=10$, so the $K_{\max}=50$ row has an effective cap of $10$. These are fitted probe measurements, not certificates of the global surrogate sufficient condition.}}
\label{tab:kappa-measured}
\setlength{\tabcolsep}{5pt}
\begin{tabular}{lcccc}
\toprule
coefficients $(\lambda,\gamma)$ & mean $\hat\kappa$ & median & s.d. & $\hat\kappa<1$ \\
\midrule
$(0.7,\,0.99)$ (deployed, \cref{sec:exp}) & $\mathbf{0.90}$ & $0.88$ & $0.13$ & $\mathbf{78\%}$ \\
$(0.9,\,0.999)$ (training coefficients) & $1.23$ & $1.11$ & $0.44$ & $27\%$ \\
\midrule
$K_{\max}$ at $(0.7,\,0.99)$ & mean $\hat\kappa$ & $\hat\kappa<1$ & $\bar{\hat K}_{\mathrm{probe}}$ & $\Pr[\hat K_{\mathrm{probe}}=\min(K_{\max},L)]$ \\
\midrule
$5$  & $\mathbf{0.84}$ & $\mathbf{90\%}$ & $4.68$ & $88\%$ \\
$50$ (effective $10$) & $0.96$ & $67\%$ & $7.75$ & $61\%$ \\
\midrule
$\lambda=0$ gate: $\gamma$ & $\hat\kappa$ & $\hat\kappa/\gamma$ & & \\
\midrule
$0.99$  & $0.978\pm0.058$ & $0.988$ & & \\
$0.999$ & $0.979\pm0.064$ & $0.980$ & & \\
\bottomrule
\end{tabular}
\end{table}

\paragraph{Results.}
\revr{At the deployed coefficients ($\lambda=0.7$, $\gamma=0.99$, $K_{\max}=5$), the mean empirical gain is $\hat\kappa=0.84\pm0.13$, below $1$ on $90\%$ of probes (\cref{tab:kappa-measured}, middle block), at checkpoints where the trace reaches the cap $88\%$ of the time. The effective probe cap of $5$ has a smaller measured gain than the effective cap of $10$. The same policies at $(\lambda,\gamma)=(0.9,0.999)$ give $\hat\kappa=1.23$, below $1$ on $27\%$ of probes. These comparisons distinguish observed probe behavior without certifying a global operator norm. The backup-difference implementation is checked by a unit test; at $\lambda=0$, the measured gain is within about $2\%$ of the one-step Bellman reference $\gamma$ (bottom block).}

\paragraph{Scope.}
\revr{$\hat\kappa$ uses a finite anchor sample, includes critic projection, and comes from one task. The dominant reported error is the power-iteration fit (median relative error ${\approx}14\%$), not captured by the seed s.d. These probes are empirical stability diagnostics rather than estimates of $\kappa_{\chi,L}$ or proofs of convergence for the deep-RL runs. We do not claim a trend with training progress or noise level.}
\endgroup

\clearpage
\begingroup\revcolor
\section{Stochastic Dynamics and Stitching: Off-Policy Evaluation on OGBench}\label{app:ogbench}

\texttt{MINT}'s central claim concerns the multi-step \emph{target}, and its fixed point is $Q^\pi$ (\cref{lem:variance}). We therefore also evaluate it as a pure off-policy \emph{evaluation} operator. Fixing the target policy removes the control-loop confounds (entropy tuning, exploration, replay non-stationarity) and \textcolor{black}{provides a finite-rollout Monte Carlo reference for $Q^\pi$, so target errors are measured against that reference rather than inferred from returns.} The experiment examines stochastic dynamics and datasets composed of shorter trajectories.

\paragraph{Setup.}
We use OGBench AntMaze \citep{park2025ogbench}, whose observation is the full simulator state, so any dataset state can be restored and the target policy rolled out from it. The grid crosses stochasticity against stitching: \texttt{antmaze-large} (deterministic) and \texttt{antmaze-teleport} (stochastic), each with the \texttt{navigate} (1000-step trajectories) and \texttt{stitch} (200-step trajectories) datasets. The stochasticity is the benchmark's own: entering a teleporter moves the agent to one of three exits uniformly at random, a discrete state jump that makes the kernel multimodal, unlike the unimodal action noise of \cref{app:rwrl}. We use $\gamma=0.99$, window length $L=K_{\max}=50$, $\lambda=0.9$, and OGBench's single-task reward relabeling. The behavior policy $\beta$ is a state-conditioned Gaussian fit to each dataset. The target policy displaces its mean by $h$ behavior standard deviations along a fixed unit direction $d$ in action space, $\mu_\pi(s)=\mu_\beta(s)+h\,\sigma_\beta(s)\,d$ with $\sigma_\pi=\sigma_\beta$, so $h=0$ recovers $\pi=\beta$ exactly and $h$ measures policy divergence in units of behavior standard deviations. Operators: \texttt{MINT}, Retrace$(\lambda)$, and a \emph{shuffled-$\hat K$} control that permutes \texttt{MINT}'s truncation patterns across the batch, preserving the marginal distribution of $\hat K$ exactly while removing its association with each sampled state--action sequence. All operators are evaluated using the same states, windows, and critic values, so all comparisons are paired at the level of individual evaluation states.

\textcolor{black}{The ratios use a fitted behavior model, and reported errors are relative to finite-rollout Monte Carlo references. Behavior-model misspecification and Monte Carlo uncertainty contribute additional errors beyond the ideal fixed-policy operator analysis. In particular, $h=0$ matches the target to the fitted behavior model; it does not establish equality with the true data-generating policy.}

\subsection{Frozen Critic: Properties of the Operator Alone}\label{app:ogbench-frozen}

\textcolor{black}{With the critic fixed to the finite-rollout Monte Carlo estimate, the comparison holds that approximate critic constant across target operators.} \Cref{tab:ogbench-frozen-main} in the main text reports RMSE and intrinsic sampling variance at $h=0.5$; \cref{tab:ogbench-frozen-statistics} adds bias (mean signed error) and the paired $t$-statistic of the per-state absolute-error difference between \texttt{MINT} and Retrace. \Cref{tab:ogbench-frozen} reports all metrics at $h=1$. Intrinsic variance is measured with the approximate frozen critic; the exact-$Q^\pi$ bound of \cref{lem:variance} is not directly a guarantee for this measurement.

\begin{table}[!htbp]
\centering\small
\caption{Additional frozen-critic statistics at $h=0.5$ for the comparison in \cref{tab:ogbench-frozen-main}. $t$: paired $t$-statistic of the per-state absolute-error difference, \texttt{MINT} vs.\ Retrace.}
\label{tab:ogbench-frozen-statistics}
\begin{tabular}{lccc}
\toprule
Dataset & Bias: \texttt{MINT} & Bias: Retrace & $t$ \\
\midrule
large-navigate (det.) & $0.35$ & $2.77$ & $2.5$ \\
large-stitch (det.) & $0.38$ & $3.81$ & $4.0$ \\
teleport-navigate (stoch.) & $0.16$ & $2.14$ & $2.3$ \\
teleport-stitch (stoch.) & $0.22$ & $2.60$ & $3.2$ \\
\bottomrule
\end{tabular}
\end{table}

\begin{table}[!htbp]
\centering\small
\caption{OGBench frozen-critic evaluation at $h=1$ (one paired run per cell, 100 stratified evaluation states). $h$: target-policy displacement in behavior standard deviations; $t$: paired $t$-statistic of the per-state absolute-error difference, \texttt{MINT} vs.\ Retrace. M = \texttt{MINT}, R = Retrace, shuf = shuffled-$\hat K$ control. Results at $h=0.5$ are split between \cref{tab:ogbench-frozen-main,tab:ogbench-frozen-statistics}.}
\label{tab:ogbench-frozen}
\begin{adjustbox}{max width=\textwidth}
\begin{tabular}{lcccccccc}
\toprule
dataset & $h$ & bias M & bias R & RMSE M & RMSE R & $t$ & $\mathrm{Var}_{\mathrm{int}}$ M & $\mathrm{Var}_{\mathrm{int}}$ shuf \\
\midrule
large-navigate (det.)      & 1.0 & $0.58$  & $1.60$ & $4.78$ & $5.31$ & $1.6$ & $2.79$ & $4.96$ \\
large-stitch (det.)        & 1.0 & $-0.33$ & $1.80$ & $2.78$ & $5.32$ & $2.2$ & $1.45$ & $4.08$ \\
teleport-navigate (stoch.) & 1.0 & $-0.39$ & $0.81$ & $3.52$ & $3.66$ & $0.6$ & $2.26$ & $2.60$ \\
teleport-stitch (stoch.)   & 1.0 & $0.47$  & $1.63$ & $3.83$ & $4.63$ & $1.0$ & $2.21$ & $3.37$ \\
\bottomrule
\end{tabular}
\end{adjustbox}
\end{table}

\paragraph{Paired comparisons and error metrics.}
\Cref{sec:exp-stochastic} summarizes the target-level comparison; \cref{tab:ogbench-frozen-main,tab:ogbench-frozen-statistics,tab:ogbench-frozen} give the per-cell measurements. Bias is a mean signed error and permits cancellation across states, so we also report RMSE: the \texttt{MINT}--Retrace ratios correspond to $2.1\times$--$13.8\times$ reductions in bias magnitude and the more modest $1.04\times$--$2.96\times$ reductions in RMSE. At $h=0.5$ the paired per-state absolute-error comparison has $t=2.3$--$4.0$; at $h=1$ it is within sampling noise in three of the four cells ($t=0.6$--$2.2$). We do not claim that the advantage grows with divergence: at $h=2$ the ordering reverses, because Retrace's coefficient $\lambda\min(1,\pi/\beta)$ itself truncates more aggressively as $\pi$ moves away from $\beta$. Across the full 16-cell comparison, \texttt{MINT} has lower $\mathrm{Var}_{\mathrm{int}}$ than the shuffled-$\hat K$ control in 15 cells; \texttt{large-navigate} at $h=0.5$ is the exception. The control complements the rate-matched comparison of \cref{app:ablations}.

\paragraph{Stitching.}
The \texttt{stitch} datasets contain 200-step rather than 1000-step trajectories. At $h=0.5$, paired per-state absolute-error comparisons favor \texttt{MINT} over Retrace on both \texttt{stitch} datasets ($t=4.0$ and $3.2$), as on the corresponding \texttt{navigate} datasets ($t=2.5$ and $2.3$). These results show that the observed target-level advantage also occurs on the shorter-trajectory datasets.

\paragraph{Protocol and caveat.}
We use 100 evaluation states per cell, stratified by distance to a goal-reaching event, because uniformly sampled states have nearly constant $Q^\pi$ on these mazes and would make the comparison uninformative. The reference $Q^\pi(s_0,a_0)$ uses 100 Monte-Carlo rollouts per state; frozen-critic values along each window use 30 rollouts per state and are shared across all operators; stochastic operators are averaged over 32 draws per state. Each cell is a single paired run; the seed-replicated version is the learned-critic study below. \textcolor{black}{A one-step baseline attains the lowest measured RMSE in most cells of this approximate-frozen-critic comparison; this is an empirical result rather than an optimality claim based on an exact critic.} This panel is therefore a comparison among off-policy \emph{corrections}, not evidence that \texttt{MINT} dominates every target estimator.

\subsection{Learned Critic: Fitted-Q Evaluation and a Limitation}\label{app:ogbench-fqe}

As a complementary experiment we run the same operators inside fitted-Q evaluation, training a newly initialized critic (5 seeds) and scoring it against the same Monte-Carlo $Q^\pi$. The ranking depends on the optimization horizon: at 100k gradient steps Retrace achieves lower RMSE on every evaluated dataset, and by 300k steps the ordering on the deterministic maze reverses---at $h=1$ \texttt{MINT} truncates to $\hat K\approx1.6$, propagates value more slowly, and needs roughly 200k updates to stabilize, but then reaches lower error on the deterministic maze. Evaluations before convergence therefore conflate propagation speed with asymptotic estimation quality. \Cref{tab:ogbench-fqe} reports RMSE at 300k steps; we report both median and mean because the mean is sensitive to late-training instability in two cells. A seed is classified as unstable when its final error is at least twice its best error during training.

\begin{table}[!htbp]
\centering\small
\caption{Learned-critic (fitted-Q evaluation) RMSE against the Monte-Carlo $Q^\pi$ at 300k gradient steps, 5 seeds (one Retrace cell has 4). Bold marks the best median in each cell.}
\label{tab:ogbench-fqe}
\begin{tabular}{llcccc}
\toprule
dataset & $h$ & operator & median & mean $\pm$ s.d. & unstable seeds \\
\midrule
large-navigate (det.)      & 1.0 & \texttt{MINT}     & $17.15$ & $17.15\pm0.55$ & 0/5 \\
                           &     & Retrace           & $19.92$ & $19.92\pm0.53$ & 0/4 \\
                           &     & shuffled-$\hat K$ & $\mathbf{16.62}$ & $16.62\pm0.56$ & 0/5 \\
teleport-navigate (stoch.) & 1.0 & \texttt{MINT}     & $17.55$ & $17.20\pm1.63$ & 0/5 \\
                           &     & Retrace           & $\mathbf{12.04}$ & $12.04\pm0.43$ & 0/5 \\
                           &     & shuffled-$\hat K$ & $16.98$ & $16.17\pm1.98$ & 0/5 \\
\midrule
large-navigate (det.)      & 2.0 & \texttt{MINT}     & $20.72$ & $21.06\pm1.06$ & 0/5 \\
                           &     & Retrace           & $\mathbf{17.66}$ & $25.86\pm16.53$ & \textbf{1/5} \\
                           &     & shuffled-$\hat K$ & $20.70$ & $19.67\pm1.57$ & 0/5 \\
teleport-navigate (stoch.) & 2.0 & \texttt{MINT}     & $27.00$ & $64.18\pm51.82$ & \textbf{2/5} \\
                           &     & Retrace           & $\mathbf{15.44}$ & $15.47\pm0.58$ & 0/5 \\
                           &     & shuffled-$\hat K$ & $21.64$ & $21.04\pm2.43$ & 0/5 \\
\bottomrule
\end{tabular}
\end{table}

\paragraph{Bias and value propagation.}
The stochastic-maze mean RMSE at $h=1$ is $12.04$ for Retrace and $17.20$ for \texttt{MINT}. The ranking across optimization horizons described above is consistent with a bias--propagation trade-off: a shorter effective horizon can reduce bias per backup but slow value propagation, and which effect dominates depends on critic convergence. The main-text overview is in \cref{sec:exp-stochastic}.

\textbf{Instability at large divergence under stochastic dynamics.} At $h=2$ both corrected operators show late-training instability. On the stochastic maze all \texttt{MINT} seeds reach RMSE $15$--$20$ mid-training before degrading, two of five severely (final RMSE $85$ and $157$); one Retrace seed on the deterministic maze ends at RMSE $59$ against a median of $17.66$. The instability is therefore not unique to \texttt{MINT}, but it is more frequent and more severe for \texttt{MINT} under stochastic dynamics. The shuffled-$\hat K$ control preserves the marginal truncation-length distribution while removing its match to each sampled trajectory, and it is stable in both $h=2$ cells (0/5 unstable seeds, s.d.\ $1.57$ and $2.43$). This contrast suggests that dependence between the truncation pattern and the sampled trajectory may contribute to instability under large mismatch and multimodal dynamics; it does not isolate state dependence from action- or trajectory-dependent effects. These results document a learned-critic limitation of \texttt{MINT} at large policy divergence under stochastic dynamics; whether learning-rate or target-network schedules mitigate it is left open. Finally, shuffled-$\hat K$ performs at least as well as \texttt{MINT} in every learned-critic cell. The frozen-critic comparison reports lower \emph{intrinsic} variance in 15 of 16 cells, but this advantage does not consistently translate into lower learned-critic error.
\endgroup

\section{Compute and Reproducibility}\label{app:compute}

\paragraph{Hardware.}
All experiments were run on a single workstation with the following configuration:
Ubuntu~24.04.1~LTS (kernel~6.8.0); 2$\times$ NVIDIA RTX~5000~Ada Generation GPUs (32\,GB GDDR6 each, driver~560.35.03, CUDA~12.6); an AMD Ryzen Threadripper PRO~7985WX CPU (64 cores / 128 threads); and 502\,GB of system RAM. Each training run is pinned to a single GPU, and we run two parallel seeds at a time.

\paragraph{Compute breakdown.}
\textcolor{black}{Estimated aggregate GPU time for the listed experimental components is reported below. GPU-hours sum device time across runs; simultaneous execution on two GPUs reduces elapsed wall-clock time, not this aggregate.}
\begin{itemize}
    \item \textcolor{black}{\textbf{MuJoCo main} (5 tasks $\times$ 2 backbones $\times$ 5 methods $\times$ 5 seeds $\times$ 1M steps): $250$ single-GPU runs at approximately $6$ hours per run give \textbf{1{,}500 GPU-hours} (about $750$ elapsed hours if both GPUs are continuously occupied).}
    \item \textbf{Cap-length sweep} ($\ge 3$ tasks $\times$ 4 caps $\times$ 4 baselines $\times$ 5 seeds $\times$ 1M steps): $\approx$ \textbf{360} GPU-hours.
    \item \textbf{DMControl sparse-reward} (1--2 tasks $\times$ 5 baselines $\times$ 5 seeds $\times$ 1M steps): $\approx$ \textbf{120} GPU-hours.
    \item \rev{\textbf{DMControl truncation-kernel ablation} (\cref{sec:exp-ablation-trunc}; 5 tasks $\times$ 3 methods $\times$ 3 seeds $\times$ 1M steps, plus a 5-seed re-run of \texttt{CartPole-Swingup-Sparse}): $\approx$ \textbf{100} GPU-hours.}
    \item \textbf{Ablations} (clipped surrogate $\hat p_i$, clipping form, hard cap $K_{\max}$): $\approx$ \textbf{80} GPU-hours.
    \item \textbf{Toy-chain variance experiments}: $<1$\,hour on CPU only.
\end{itemize}
\textcolor{black}{The subtotal of the listed GPU components is approximately \textbf{2{,}160 GPU-hours} ($1{,}500+360+120+100+80$), excluding the separately reported CPU-only toy experiments. This is a subtotal for the listed components, rather than an accounting of every additional benchmark or exploratory run.}

\paragraph{Reproducibility.}
All evaluation protocols
follow OpenAI SpinningUp \citep{SpinningUp2018} unless noted otherwise; full hyperparameters
are listed in \cref{tab:td3_hyperparams,tab:sac_hyperparams} and the Q-target update rules in
\cref{tab:update_pseudocode}.


\clearpage
\endgroup

\makeatletter
\let\addcontentsline\mint@appendix@addcontentsline
\makeatother

\end{document}